\documentclass{bmvc2k}

\usepackage{graphicx}
\usepackage{microtype}
\usepackage{subfigure}
\usepackage{booktabs} 
\usepackage{amssymb}
\usepackage{amsmath}
\usepackage{algorithm}
\usepackage{algorithmic}

\usepackage{amsmath,amsfonts,bm}

\def\eqref#1{equation~\ref{#1}}

\def\1{\bm{1}}

\DeclareMathAlphabet{\mathsfit}{\encodingdefault}{\sfdefault}{m}{sl}
\SetMathAlphabet{\mathsfit}{bold}{\encodingdefault}{\sfdefault}{bx}{n}

\DeclareMathOperator*{\argmax}{arg\,max}
\DeclareMathOperator*{\argmin}{arg\,min}

\usepackage{threeparttable}
\usepackage{array}
\newcolumntype{H}{>{\setbox0=\hbox\bgroup}c<{\egroup}@{}}
\usepackage{xr}
\usepackage{multirow}
\usepackage{enumitem}

\usepackage{tikz}
\usetikzlibrary{automata, positioning, arrows}
\def\firstcircle{(0,0) circle (1.8cm)}
\def\secondcircle{(0:2cm) circle (1.8cm)}
\definecolor{xablue}{HTML}{00ABF5}
\definecolor{xayellow}{HTML}{FACD00}
\definecolor{xapink}{HTML}{FF6D8A}
\definecolor{xagreen}{HTML}{00DDCB}
\definecolor{xagray}{HTML}{545454}
\colorlet{circle edge}{black}
\colorlet{circle area}{blue!20}

\title{Noisy Differentiable Architecture Search}

\addauthor{Xiangxiang Chu}{chuxiangxiang@xiaomi.com}{1}
\addauthor{Bo Zhang}{zhangbo11@xiaomi.com}{1}

\addinstitution{
Xiaomi AI Lab\\
Beijing, China
}

\runninghead{Chu et al.}{NoisyDARTS}

\begin{document}

\maketitle

\begin{abstract}
\emph{Simplicity is the ultimate sophistication}. Differentiable Architecture Search (DARTS) has now become one of the mainstream paradigms of neural architecture search. However, it largely suffers from the well-known performance collapse issue due to the aggregation of skip connections. It is thought to have overly benefited from the residual structure which accelerates the information flow. To weaken this impact, we propose to inject unbiased random noise to impede the flow. We name this novel approach NoisyDARTS. In effect,  a network optimizer should perceive this difficulty at each training step and refrain from overshooting, especially on skip connections. In the long run, since we add no bias to the gradient in terms of expectation, it is still likely to converge to the right solution area. We also prove that the injected noise plays a role in smoothing the loss landscape, which makes the optimization easier.  Our method features extreme simplicity and acts as a new strong baseline. 
We perform extensive experiments across various search spaces, datasets, and tasks, where we robustly achieve state-of-the-art results. Our code is available\footnote{\url{https://github.com/xiaomi-automl/NoisyDARTS.git}}.

\end{abstract}

\section{Introduction} \label{sec:intro}
Differentiable architecture search \cite{liu2018darts}  suffers from 
a well-known \emph{performance collapse} issue noted by \cite{chu2019fair,chen2019progressive}. Namely, while the over-parameterized model is well optimized, its inferred model tends to have an excessive number of skip connections, which dramatically degrades the searching performance. Quite an amount of previous research has focused on addressing this issue  \cite{chen2019progressive,zela2020understanding,liang2019darts,chu2019fair,li2019sgas}. Among them, Fair DARTS \cite{chu2019fair} concludes that it is due to an unfair advantage in an exclusively competitive environment. Under this perspective, early-stopping methods like \cite{zela2020understanding,liang2019darts} or greedy pruning \cite{li2019sgas} can be regarded as means to prevent such unfairness from overpowering. However, the one-shot network is generally not well converged if halted too early, which gives low confidence to derive the final model.

More precisely, most of the existing approaches \cite{chen2019progressive,liang2019darts,zela2020understanding}  addressing the fatal collapse can be categorized within the following framework:  
first, characterize the outcome when the collapse occurs (e.g larger Hessian eigenvalue as in   RobustDARTS \cite{zela2020understanding} or too many skip connections in a cell \cite{chen2019progressive,liang2019darts}), and then carefully design various criteria to avoid stepping into it. There are two main drawbacks of these methods. One is that the search results heavily rely on the validity of  human-designed criteria, otherwise, inaccurate criteria may reject good models (see Section~\ref{sec:hessian}).  The other is that these criteria only force the searching process to stay away from a bad solution. However, the goal of neural architecture search is not just to avoid bad solutions but to robustly find much better ones. 


 Our contributions can be summarized in the following,

\setlist{nosep,after=\vspace{\baselineskip}}
\begin{itemize}[noitemsep,after=]
	\item  Other than designing various criteria, we demonstrate a simple but effective approach to address the performance collapse issue in DARTS. Specifically, we inject various types of independent noise into the candidate operations to make good ones robustly win.  This approach also has an effect of smoothing loss landscape.

	\item We prove that the required characteristics of the injected noise should be unbiased and of moderate variance. Furthermore, it is the unbiasedness that matters, not a specific noise type. Surprisingly, our well-performing models are found with rather high Hessian eigenvalues, disproving the need for the \textbf{single-point Hessian norm} as an indicator of the collapse \cite{zela2020understanding}, since \emph{it can't describe the overall curvatures of its wider neighborhood}.
	\item Extensive experiments performed across various search spaces (including the more difficult ones proposed in \cite{zela2020understanding} and datasets (\textbf{15} benchmarks in total) show that our method can address the collapse effectively.  Moreover, we robustly achieve state-of-the-art results with 3$\times$ fewer search costs than RobustDARTS.

\end{itemize}

	


\section{Related work}

\textbf{Differentiable architecture search\quad }DARTS \cite{liu2018darts} has widely disseminated the paradigm of solving architecture search with gradient descent \cite{cai2018proxylessnas,nayman2019xnas,xie2018snas}. It constructs an over-parameterized supernet incorporating all the choice operations. Each discrete choice is assigned with a continuous architectural weight $\alpha$ to denote its relative importance, and the outputs of all the paralleling choices are summed up using a softmax function $\sigma$. Through iterative optimization of supernet parameters and architectural ones, competitive operations are supposed to stand out with the highest $\sigma(\alpha)$ to be chosen to derive the final model. Though being efficient, it is known unstable to reproduce \cite{Yu2020Evaluating}.

\textbf{Endeavors to improve the performance collapse in DARTS\quad}
Several previous works have focused on addressing the collapse. For instance, P-DARTS \cite{chen2019progressive} point out that DARTS gradually leans towards skip connection operations since they ease the training. However, while being parameter-free, they are essentially weak to learn visual features which lead to degenerate performance. To resolve this, they proposed to drop out paths through skip connections with a decay rate. Still, the number of skip connections in normal cells varies, for which they impose a hard-coded constraint, limiting this number to be $M$. Later DARTS+ \cite{liang2019darts} simply early stops when there are exactly two skip connections in a cell. RobustDARTS \cite{zela2020understanding} discovers degenerate models (where skip connections are usually dominant) correlate with increasingly large Hessian eigenvalues, for which they utilize an early stopping strategy while monitoring these values.  






\section{Noisy DARTS}
\subsection{Motivation}
We are motivated by two distinct and orthogonal aspects of DARTS: how to make the optimization easier and how to remove the unfair competition from candidate operations. 

\paragraph{Smooth loss landscape helps stochastic gradient optimization (SGD) to find the solution path at early optimization stages.} 
SGD can escape local minima to some extent \cite{kleinberg2018alternative} but still have difficulty navigating chaotic loss landscapes \cite{li2018visualizing}. Combining it with a smoother loss function can relieve the pain for optimization which leads to better solutions \cite{Gulcehre2016Mollifying,li2018visualizing}.  
Previously, RobustDARTS \cite{zela2020understanding}  empirically finds that the collapse is highly related to the sharp curvature of the loss w.r.t $\alpha$, for which they use Hessian eigenvalues as an indicator of the collapse. However, this indirect indicator at a local minimum fails to characterize its relatively larger neighborhood, which we discuss in detail in Section~\ref{sec:hessian}. Therefore, we are driven to contrive a more direct and effective way to smooth the landscape.

\paragraph{Adding noise is a promising way to smooth the loss landscape.}
Random noises are used to boost  adversarial  generalization by \cite{liu2018towards,He_2019_CVPR,lecuyer2019certified}.
A recent study by \cite{wu2020revisiting} points out that a flatter adversarial loss landscape is closely related to better generalization. This leads us to first reformulate DARTS from the probabilistic distribution's perspective as follows,
\begin{equation}\label{eq:max_darts}
\begin{aligned}
\alpha^* &  =\argmin_{\alpha} \mathcal{L}_{v}(w,\alpha, z)  =\argmax_{\alpha}  \Bbb E_{x,y \sim P_{val};z\sim P(z)}  \log P(y|x, w^*, \alpha, z)  \\
\text{s.t.} \quad  w^* &= \argmax_{w}\Bbb E_{x,y \sim P_{train}; z\sim P(z)}  \log P(y|x, w, \alpha, z) \\
\end{aligned}
\end{equation}
where $z \sim \delta(z)$.
The random variable $z$ is subject to the Dirac distribution and added to the intermediate features. For a multiplicative version, we can simply set $z \sim \delta(z-1)$. We follow \cite{liu2018darts} for the rest notations. 
To incorporate noise and smooth DARTS (Equation \ref{eq:max_darts}), we propose a direct approach by setting, 
\begin{equation}\label{eq:max_noisydarts}
\begin{aligned}
z \sim  N(\mu, \sigma).\
\end{aligned}
\end{equation}


We choose additive Gaussian noise for simplicity. Experiments on uniform noise are also provided in Section~\ref{sec:more_ablation} (supplementary).  The remaining problem is where to inject the noise and how to calibrate $\mu$ and $\sigma$. 

\textbf{Unfairness of skip connections from fast convergence.\quad} Apart from the above-mentioned perspective, we notice from prior work that \emph{skip connections} are the primary subject to consider \cite{chen2019progressive,liang2019darts,chu2019fair}. While being summed with other operations, a skip connection builds up a \emph{residual structure} as in \cite{he2016deep}.  A similar form is also proposed in highway networks \cite{srivastava2015training}. Such a residual structure is generally helpful for training deep networks, as well as the supernet of DARTS. However, as skip connections excessively benefit from this advantage \cite{chen2019progressive,chu2019fair}, it leads us to overestimate its relative importance, while others are under-evaluated.
Therefore, it is appropriate to disturb the gradient flow (by injecting noise as a natural choice) right after the intermediate outputs of various candidate operations. In this way, we can regularize the gradient flow from different candidate operations and let them compete in a fair environment.  We term this approach \textbf{NFA}, short for ``Noise For All".
Considering the unfair advantage is mainly from the skip connection, we can also choose to inject noises only after this operation. We call this approach \textbf{OFS}, short for ``Only For Skip". This option is even simpler than NFA. We use OFS as the default implementation.

\subsection{Requirements for the injected noise}\label{theory}

A basic and reasonable requirement is that, applying Equation~\ref{eq:max_noisydarts} should make a close approximation to Equation~\ref{eq:max_darts}.  
Since each iteration is based on backward propagation, we relax this requirement to \textbf{having an unbiased gradient in terms of its expectation at each iteration.} Here we induce the requirement based on OFS for simplicity.

\textbf{Design of $\mu$ }\quad
We let $\tilde{x}$ be the noise injected into the skip operation, and $\alpha_{s}$ be the corresponding architectural weight. The loss of a skip connection operation can be written as,
\begin{equation}
\begin{split}
	{\cal L} &= g(y), \quad y = f(\alpha_{s}) \cdot (x + \tilde{x}) \\
\end{split}
\end{equation}
where $g(y)$ is the validation loss function and $f(\alpha_{s})$ gives the softmax output for $\alpha_{s}$. 
Approximately, when the noise is much smaller than the output features, we have
\begin{equation}\label{eq:approximate}
y^\star  \approx f(\alpha_{s}) \cdot x \quad when \quad \tilde{x} \ll x.
\end{equation}
In the noisy scenario, the gradient of the  parameters via the skip connection operation becomes,
\begin{equation}\label{eq:noise-exp}
\begin{split}
\frac{\partial{\cal L}}{\partial{\alpha_{s}}} &= 
\frac{\partial{\cal L}}{\partial y} \frac{\partial y}{\partial \alpha_{s}} = \frac{\partial{\cal L}}{\partial y}  \frac{\partial f(\alpha_{s})} {\partial{\alpha_{s}}} 
 \left(x + \tilde{x}  \right).
\end{split}
\end{equation}
As random noise $\tilde{x}$ brings uncertainty to the gradient update, skip connections have to overcome this difficulty in order to win over other operations. Their unfair advantage is then much weakened. However, not all types of noise are equally effective in this regard. 
Formally,  the expectation of its gradient can be written as,
\begin{equation}{\label{eq:gradient}}
\begin{aligned}
{\Bbb E} _{\tilde{x}} \left[\nabla_{\alpha_{s}} \right] &=
{\Bbb E} _{\tilde{x}} \left[\frac{\partial \cal L}{\partial{y}} \frac{\partial f(\alpha_{s})}{\partial{\alpha_{s}}} \left( x + \tilde{x} \right) \right]  \approx \frac{\partial \cal L}{\partial{y^\star}}  \frac{\partial f(\alpha_{s})}{\partial{\alpha_{s}}} 
\left(x  + {\Bbb E} _{\tilde{x}} \left[\tilde{x} \right] \right).
\end{aligned}
\end{equation}
Supposing that $\frac{\partial \cal L}{\partial{y}} $ is smooth, we can use $\frac{\partial \cal L}{\partial{y^{\star}}} $ to approximate the its small neighbor hood. Based on 
 the premise stated in Equation~\ref{eq:approximate}, we take $\frac{\partial \cal L}{\partial{y^\star}} $ out of the expectation in Equation~\ref{eq:gradient} to make an approximation. As there is still an extra ${\Bbb E}\left[\tilde{x} \right]$ in the gradient of skip connection, to keep the gradient unbiased, ${\Bbb E}\left[\tilde{x} \right]$ should be 0. It's intuitive to see the unbiased injected noise can play a role of encouraging the exploration of other operations. 

\textbf{Design of $\sigma$}\quad
The variance $\sigma^2$ controls the magnitude of the noise, which also represents the strength to step out of local minima. Intuitively, the noise should neither be too big (overtaking) nor too small (ineffective).
For simplicity, we start with Gaussian noise and other options are supposed to work as well. Notably, applying Equation~\ref{eq:max_noisydarts} when $\sigma$=0 falls back to Equation~\ref{eq:max_darts}.


\subsection{Stepping out of the performance collapse by noise}

Based on the above analysis, we propose NoisyDARTS to step out of the performance collapse. In practice, we inject Gaussian noise $\tilde{x} \sim {\cal N}(\mu, \sigma)$ into skip connections to weaken the unfair advantage. 
Formally, the edge $e_{i,j}$ from node $i$ to $j$ in each cell operates on $i$-th input feature $x_i$ and its output is denoted as $o_{i, j}(x_{i})$. The intermediate node $j$ gathers all inputs from the incoming edges:$x_{j} = \sum_{i<j}o_{i, j}(x_i)$.
Let ${\cal O}=\{o_{i,j}^{0}, o_{i, j}^{1}, \cdots, o_{i, j}^{M-1}\}$ be the set of $M$ candidate operations on edge $e_{i, j}$ and specially  let $o_{i,j}^{0}$ be the skip connection $o_{i,j}^{skip}$. NoisyDARTS injects the additive noise $\tilde{x}$ into skip operation $o_{i,j}^{skip}$ to get a mixed output,
\begin{equation}
\overline{o}_{i,j}(x) = \sum_{k=1}^{M-1} f(\alpha_{o^{k}}) o^{k}(x) + f(\alpha_{o^{skip}})o^{skip}( x + \tilde{x} ).
\end{equation}


The architecture search problem remains the same as the original DARTS, which is to alternately learn $\alpha^{*}$ and network weights $w^{*}$ that minimize the validation loss ${\cal L}_{val}(\alpha^{*}, w^{*})$. 
To summarize, NoisyDARTS (OFS) is shown in Algorithm~\ref{alg:noisy-darts-ofs} (supplementary). The NFA version is in Algorithm~\ref{alg:noisy-darts-nfa} (supplementary). 



\textbf{The role of noise}. The role of the injected noise is threefold. Firstly, it breaks the unfair advantage so that the final chosen skip connections indeed have substantial contribution for the standalone model. Secondly, it encourages more exploration to escape bad local minima, whose role is akin to the noise in SGLD \cite{zhang2017hitting}. Lastly,  it smooths the loss landscape w.r.t $\alpha_{s}$ (NFA is similar).  If we denote validation loss as $\mathcal{L}_{v}$, this role can be explained due to the fact that our approach implicitly controls the loss landscape. Supposing that the injected noise $z$ is small and   $z\sim N(\bold{0},\sigma^2\bold{I})$,  the expectation of the loss over $z$ can be approximated by

\begin{small}
\begin{equation}{\label{eq:hessian_reg}}
\vspace*{-0.5\baselineskip}
\setlength\abovedisplayskip{3pt}
\setlength\belowdisplayskip{3pt}
\begin{aligned}
{\Bbb E} _{z} \left[\mathcal{L}_{v}(w,\alpha_{s},z) \right] &\approx  {\Bbb E} _{z} [\mathcal{L}_{v}(w,\alpha_{s},\bold{0}) + \nabla_{z}\mathcal{L}_{v}(w,\alpha_{s},\bold{0})z  
\ + \frac{1}{2}  z^T \nabla_{z}^2 \mathcal{L}_{v}(w,\alpha_{s},\bold{0}) z ] \\
&= \mathcal{L}_{v}(w,\alpha_{s},\bold{0}){\Bbb E} _{z} \boldsymbol{1}  + \nabla_{z=\bold{0}}\mathcal{L}_{v}(w,\alpha_{s},\bold{0})\Bbb E _{z}z  
\ + \Bbb E _{z} \frac{1}{2}  z^T \nabla_{z}^2 \mathcal{L}_{v}(w,\alpha_{s},\bold{0}) z ] \\ 
&= \mathcal{L}_{v}(w,\alpha_{s},\bold{0} ) + \frac{\sigma^2}{2}Tr\{\nabla_{z}^2 \mathcal{L}_{v}(w,\alpha_{s},\bold{0})  \} \\&\approx \mathcal{L}_{v}(w,\alpha_{s},\bold{0} ) + \frac{\beta \sigma^2\alpha_{s}^2 }{2}Tr\{\nabla_{\alpha_{s}}^2 \mathcal{L}_{v}(w,\alpha_{s},\bold{0})  \} \\
\text{where \quad \quad} z&\sim N(\bold{0},\sigma^2\bold{I}), \beta=\Bbb E{\frac{1}{o_{skip}(x)^To_{skip}(x)}},  \text{ $\bold{I}$ is unit matrix.}
%
\end{aligned}
\end{equation}
\end{small}

Its role can be better understood via the visualization in Figure~\ref{fig:landscape-comparison}, where DARTS obtains a sharp landscape with oval contours and ours has round ones.

\section{Experiments}\label{sec:exp}

\subsection{Search spaces and 15 benchmarks}{\label{sec:ss}
To verify the validity of our method, we adopt several search spaces: the DARTS search space from \cite{liu2018darts}, MobileNetV2's search space as in \cite{cai2018proxylessnas}, four harder spaces (from $S_1$ to $S_4$) from \cite{zela2020understanding}. We use NAS-Bench-201 \cite{dong2020nasbench} to benchmark our methods. 

\textbf{DARTS's search space (Benchmark 1)} It consists of a stack of duplicate normal cells and reduction cells, which are represented by a DAG of 4 intermediate nodes. Between every two nodes there are several candidate operations (max pooling, average pooling, skip connection, separable convolution 3$\times$3 and 5$\times$5, dilation convolution 3$\times$3 and 5$\times$5). 

\textbf{MobileNetV2's search space (Benchmark 2)}  It is the same as that in ProxylessNAS \cite{cai2018proxylessnas}. We search proxylessly on ImageNet in this space. It uses the standard MobileNetV2's backbone architecture \cite{sandler2018mobilenetv2}, which comprises 19 layers and each contains 7 choices: inverted bottleneck blocks denoted as Ex\_Ky (expansion rate $x \in \{3,6\}$, kernel size $y \in \{3,5,7\}$) and a skip connection. The stem, the first bottleneck block and the tail is kept unchanged, see Figure~\ref{fig:imagenet-architecture} (supplementary) for reference.

$\mathbf{S}_1$-$\mathbf{S}_4$ (\textbf{Benchmark 3-14}) These are reduced search spaces introduced by RobustDARTS \cite{zela2020understanding}. $S_1$ is a preoptimized search space with two operation per edge, see \cite{zela2020understanding} for the detail. For each edge in the DAG, $S_2$ has only  \{3 $\times$ 3 SepConv, SkipConnect\}, $S_3$ has \{3 $\times$ 3 SepConv, SkipConnect, Zero (None)\}, and $S_4$ has \{3 $\times$ 3 SepConv, Noise\}. We search on three datasets for each search space, which makes 12 benchmarks.

\textbf{NAS-Bench-201 (Benchmark 15)} NAS-Bench-201 \cite{dong2020nasbench} is a cell based search space with known evaluations of each candidate architecture, where DARTS severely suffers  from the performance collapse. It includes 15625 sub architectures in total. Specifically, it has 4 intermediate nodes and 5 candidate operations (none, skip connection, 1$\times$1 convolution, 3$\times$3 convolution and 3$\times$3 average pooling).

\subsection{Datasets}
We use a set of standard image classification datasets CIFAR-10, CIFAR-100 \cite{krizhevsky2009learning}, SVHN \cite{netzer2011reading} and ImageNet \cite{deng2009imagenet} for both searching and training. We also search for GCNs on ModelNet \cite{wu20153d} as in \cite{li2019sgas} (see Section \ref{sec:gcn} in the supplementary).
 
\subsection{Searching Results}


\begin{table}
	\centering
	\setlength{\tabcolsep}{0.5pt}%
		\small
		\begin{threeparttable}
			\begin{tabular}{*{4}{|l}H*{1}{H}|} 			
				\hline
				Models  & Params & $\times+$ & Top-1 Acc &  Cost&Method  \\
				&\scriptsize{(M)}&\scriptsize{(M)}&($\%$)&\scriptsize{(G$\cdot$d)}& \\
				\hline
				P-DARTS \cite{chen2019progressive} & 3.4 & 532$^\dagger$ & 97.49 & 0.3 & GD \\
				PC-DARTS \cite{xu2020pcdarts} & 3.6 & 558$^\dagger$ & 97.43 & 0.1 & GD \\ 
				GDAS \cite{dong2019searching} & 3.4 & 519$^\dagger$ & 97.07 &0.3&GD \\
				\textbf{NoisyDARTS-a}& 3.3 & 534 & \textbf{97.63 }&0.4& GD \\
				NoisyDARTS-b& 3.1 & 511 & 97.53 &0.4 & GD \\
				\hline
				DARTS$^\star$\cite{liu2018darts} & 3.3 & 528$^\dagger$ & 97.00$\pm$0.14 & 0.4 & GD \\ 
				SNAS$^\star$\cite{xie2018snas} & 2.8 & 422$^\dagger$ & 97.15$\pm$0.02 & 1.5&GD\\
				PR-DARTS$^\star$ \cite{zhou2020NAS} &3.4&-&97.19$\pm$0.08 &0.2 &\\
				\hline


				P-DARTS \cite{chen2019progressive}$^\ddagger$  & 3.3$\pm$0.21 & 540$\pm$34 & 97.19$\pm$0.14 & 0.3& GD \\

				PC-DARTS \cite{xu2020pcdarts}$^\ddagger$ & 3.7$\pm$0.57 & 592$\pm$90 & 97.11$\pm$0.22 & 0.1& GD \\ 
				RDARTS \cite{zela2020understanding} &-& -& 97.05$\pm$0.21&1.6& GD \\
				DARTS- \cite{chu2020darts} & 3.5$\pm$0.13 & 583$\pm$22 & 97.41$\pm$0.08 & 0.4 & \\
				\textbf{NoisyDARTS} & 3.1$\pm$0.22 & 502$\pm$38 &97.35$\pm$0.23 &0.4 &\\

				\hline
				MixNet-M$^\ast$ \cite{tan2020mixconv} & 4.9& 359 & 97.90 & $\approx$3k &\\
				SCARLET-A \cite{chu2019scarletnas} & 5.4 & 364 & 98.05 & & \\
				EfficientNet B0$^\ast$ \cite{tan2019efficientnet} &5.2 &387 & 98.10 & $\approx$3k &\\
				\textbf{NoisyDARTS-A-t}$^\ast$ & 4.3 & 447 & \textbf{98.28} &12& TF \\
				\hline
			\end{tabular}
			\begin{tablenotes}
			\tiny
			\item[$^\dagger$]  Computed from the authors' code  $^\ddagger$ Re-run their code with 4 independent searches.
			\item[$^\star$] Averaged on the single \textbf{best} model trained for several times 
			\item[$^\ast$] Transferring ImageNet-pretrained models to CIFAR-10

			\end{tablenotes}
			\end{threeparttable}
		\begin{threeparttable}
		\begin{tabular}{*{4}{|l}*{1}{H}H|} 		
				\hline
				Models & $\times+$   &Params & Top-1 & Top-5  &  Cost \\
				& \scriptsize{(M)} & \scriptsize{(M)} & \scriptsize{(\%)} & \scriptsize{(\%)} & \scriptsize{(G $\cdot$ d)}\\
				\hline
				MobileNetV2 \cite{sandler2018mobilenetv2}   & 585 & 6.9 & 74.7 & 92.2 & - \\
				\hline
				NASNet-A \cite{zoph2017learning}  & 564 & 5.3 &74.0 & 91.6 & 1800 \\
				AmoebaNet-A \cite{real2019regularized} & 555  & 5.1 & 74.5 &92.0 & $\approx$3k\\

				MdeNAS\cite{zheng2019multinomial} & - & 6.1 & 74.5 & 92.1  & 0.16\\
				P-DARTS \cite{chen2019progressive}$^{\dagger\dagger}$ & 577 & 5.1 & 74.9$^{*}$ & 92.3$^{*}$ & 0.3 \\  
				PC-DARTS  \cite{xu2019pc} & 597 & 5.3 & 75.8 & 92.7 & 3.8\\ 
				DARTS \cite{liu2018darts} & 574 & 4.7 & 73.3 & 91.3 & 0.5\\
				GDAS \cite{dong2019searching}  &581&5.3&74.0&91.5 & 0.2\\
				\hline

				MnasNet-92 \cite{tan2018mnasnet}  & 388 & 3.9 & 74.79 & 92.1 &$\approx$4k \\ 	
				Proxyless-R \cite{cai2018proxylessnas} & 320$^\dagger$ & 4.0  & 74.6 & 92.2 & 8.3\\  
				NoisyDARTS-A &446&4.9&\textbf{76.1}& 93.0 & 12 \\
				FairNAS-C $^\ddagger$ \cite{chu2019fairnas} &321 & 4.4 & 74.7 &92.1 & \\

				FairDARTS-B \cite{chu2019fair}& 541 & 4.8 &75.1 & 92.5 & 0.4 \\

				\hline
				MobileNetV3 \cite{howard2019searching} & 219 & 5.4 &75.2 & 92.2 & $\approx$3k\\
                EfficientNet B0 \cite{tan2019efficientnet} &390& 5.3& 77.2& 93.2& $\approx$3k \\
				MixNet-M \cite{tan2020mixconv} &360 & 5.0 & 77.0 & 93.3 &$\approx$3k\\
 				NoisyDARTS-A$^{\diamond}$ &449 & 5.5 & \textbf{77.9} & 94.0 & 12\\
				
				\hline
			\end{tabular}
			\begin{tablenotes}
			\tiny
			\item[$^{\dagger\dagger}$] Searched on CIFAR-100
			\item[$^\diamond$] NoisyDARTS-A with SE and Swish enabled
			\end{tablenotes}
			\end{threeparttable}
	\caption{Results on CIFAR-10 (left) and ImageNet (right).  NoisyDARTS-a and b are the  models searched on CIFAR-10 when $\sigma=0.2$ and $\sigma=0.1$ respectively (Figure~\ref{fig:the-searched-cells-a} and \ref{fig:the-searched-cells-b} in the supplementary).   NoisyDARTS-A (Figure~\ref{fig:imagenet-architecture} in the supplementary) is searched on ImageNet in the MobileNetV2-like search space as in \cite{wu2018fbnet}.}
	\label{tab:comparison-cifar10-imagenet}
	\vskip -0.2in	
\end{table}

\paragraph{Searching on CIFAR-10.}
In the search phase, we use similar hyperparameters and tricks as \cite{liu2018darts}. All experiments are done on a Tesla V100 with PyTorch 1.0 \cite{paszke2019pytorch}. The search phase takes about 0.4 GPU days. We only use the \emph{first-order} approach for optimization since it is more efficient. The best models are selected under the noise with a zero mean and $\sigma=0.2$. An example of the evolution of the architectural weights during the search phase is exhibited in Figure~\ref{fig:alpha-evolution-best} (see Section~\ref{supp:exp} in the supplementary). 
For training a single model, we use the same strategy and data processing tricks as \cite{chen2019progressive,liu2018darts}, and it takes about 16 GPU hours. 
The results are shown in Table~\ref{tab:comparison-cifar10-imagenet}. The best NoisyDARTS model (NoisyDARTS-a) achieves a new state-of-the-art result of 97.63\% with only 534M FLOPS and 3.25M parameters, whose genotypes are shown in Figure~\ref{fig:the-searched-cells-a} (see Section \ref{app:fig-archs} in the supplementary). 

\paragraph{Searching in the reduced search spaces of RobustDARTS.}
We also study the performance of our approach under reduced search spaces, compared with DARTS \cite{liu2018darts}, RDARTS \cite{zela2020understanding} and SDARTS \cite{chen2020stabilizing}. Particularly we use OFS for $S_1$, $S_2$, and $S_3$  (from \cite{zela2020understanding}), where DARTS severely suffers from the collapse owing to an excessive number of skip connections. For $S_4$ where skip operations are not present, we apply NFA. We kept the  same hyper-parameters as \cite{zela2020understanding} for training every single model to make a fair comparison. Since the unfair advantage is intensified in the reduced search spaces, we use stronger Gaussian noise (e.g. $\sigma=0.6, 0.8$). As before, we don't utilize any regularization tricks. The results are given in Table~\ref{table:rdarts-reduced-ss} and Table~\ref{table:rdarts-reduced-ss-more-sigma} (see Section \ref{app:geno} in the supplementary). Each search is repeated only three times to obtain the average. 
\begin{table}[h]
	\setlength{\tabcolsep}{1.5pt}   
	\centering
		\scriptsize
			\addtolength{\tabcolsep}{1pt}    
			\begin{tabular}{|l*{5}{|c}||c|cccH|}
				\hline
				Data & Space & DARTS &DARTS$^{\text{ADA}}$ & DARTS$^{\text{ES}}$  & \multicolumn{1}{c||}{Ours}  & RDARTS$^{\text{L2}}$ & SDARTS$^{\text{RS}}$  &  SDARTS$^{\text{ADV}}$  &  Ours & $\sigma$  \\
				\hline
				\multirow{4}{2.5em}{C10} & $S_1$ & 95.34$\pm$0.71 & 96.97$\pm$0.08 & 96.95$\pm$0.07 & \textbf{97.05$\pm$0.18} &  97.22 & 97.22 & \textbf{97.27} & \textbf{97.27}  & 0.6 \\
				& $S_2$ & 95.58$\pm$0.40 &96.41$\pm$0.31& 96.59$\pm$0.14 & \textbf{96.59$\pm$0.11} &96.69 &  96.67$^{\dagger}$ & 96.59$^{\dagger}$ &  \textbf{96.71}  & 0.6 \\
				& $S_3$ & 95.88$\pm$0.85   &97.01$\pm$0.34&96.29$\pm$1.14& \textbf{97.42$\pm$0.08} & 97.49 & 97.47 & 97.51 & \textbf{97.53} & 1  \\
				& $S_4$ & 93.05$\pm$0.18 & 96.11$\pm$0.67 & 95.83$\pm$0.21 & \textbf{97.22$\pm$0.08} & 96.44 & 97.07 & 97.13 & \textbf{97.29} & 0.8 \\
				\hline
				\multirow{4}{2.5em}{C100} & $S_1$ & 70.07$\pm$0.41 & 75.06$\pm$0.81 & 71.10$\pm$0.81 & \textbf{77.89$\pm$0.88}& 75.75 & 76.49 & 77.67 & \textbf{78.83}  &1.2 \\
				& $S_2$ & 71.25$\pm$0.92  & 73.12$\pm$1.11& 75.32$\pm$1.43&\textbf{78.15$\pm$0.44} & 77.76 & 77.72 & \textbf{79.44} & 78.82 & 0.8 \\
				& $S_3$ & 70.99$\pm$0.24 &75.45$\pm$0.63&73.01$\pm$1.79& \textbf{79.48$\pm$0.59} & 76.01 & 78.91 & 78.92 & \textbf{79.93} & 0.8 \\
				& $S_4$ &  75.23$\pm$1.51 & 76.34$\pm$0.90 & 76.10$\pm$2.01 & \textbf{78.37$\pm$0.42}  & 78.06 & 78.54 &78.75 & \textbf{78.84} & 1 \\
				\hline
				\multirow{4}{2.5em}{SVHN} & $S_1$ & 90.12$\pm$5.50 & 97.41$\pm$0.07 & 97.20$\pm$0.09 & \textbf{97.44$\pm$0.06}& 95.21 & 97.14$^\dagger$ &  \textbf{97.51}$^\dagger$   & \textbf{97.51} & 1.4 \\ 
				& $S_2$ & 96.31$\pm$0.12 & 97.21$\pm$0.22 &97.32$\pm$0.18 & \textbf{97.60$\pm$0.08} & 97.49 & 97.61 & 97.65 & \textbf{97.66} & 1.3 \\
				& $S_3$ & 96.00$\pm$1.01 &97.42$\pm$0.07 &97.22$\pm$0.19 &\textbf{97.58$\pm$0.06}  & 97.52& \textbf{97.64} & 97.60 & 97.63 & 1.3 \\
				& $S_4$ & 97.10$\pm$0.02 & 97.48$\pm$0.06 & 97.45$\pm$0.15 & \textbf{97.59$\pm$0.09} & 97.50 & 97.54 & 97.58 & \textbf{97.67} & 1 \\
				\hline
			\end{tabular}
	\caption{Comparison in the reduced spaces of RobustDARTS \cite{zela2020understanding}. For NoisyDARTS, we use NFA for $S_4$ since there is no skip connection in it, and OFS for all the rest. $^{\dagger}$: 16 initial channels (retrained). Four right columns are the best out of three runs. ADA: adaptive regularization, ES: early-stop, L2: L2 regularization (ADA, ES, L2 are from RDARTS \cite{zela2020understanding}).}
		\label{table:rdarts-reduced-ss}
\end{table}

\paragraph{Searching proxylessly on ImageNet.}
In the search phase, we use $\mu=0$ and $\sigma=0.2$ and we don't optimize the hyper-parameters regarding cost. It takes about 12 GPU days on Tesla V100 machines (more details are included in Section~\ref{sec:imagenet_training} in the supplementary). As for  training searched models, we use similar training tricks as EfficientNet \cite{tan2019efficientnet}.  The evolution of dominating operations during the search is illustrated in Figure~\ref{fig:stacked-plot-dominant-imagent} (supplementary). Compared with DARTS (66.4\%),  the injected noise in NoisyDARTS successfully eliminates the unfair advantage.
Our model NoisyDARTS-A (see Figure~\ref{fig:imagenet-architecture} in the supplementary) obtains the new state of the art results: $76.1\%$ top-1 accuracy on ImageNet with 4.9M number of parameters. After being equipped with more tricks as in EfficientNet, such as squeeze-and-excitation \cite{hu2018squeeze} and AutoAugment \cite{cubuk2018autoaugment}, it obtains  $77.9\%$ top-1 accuracy.

\paragraph{Searching on NAS-Bench-201.}
 We report the results (averaged on 3 runs of searching) on NAS-bench-201 \cite{dong2020nasbench} in Table ~\ref{tab:bench201}. Our method  surpasses SETN \cite{dong2019one} with a clear margin using 3 fewer search cost. This again proves NoisyDARTS to be a robust and powerful method. Learnable and decayed $\sigma$ are used for ablation purposes (see Section~\ref{sec:ablation}). 
 
\begin{table*}[ht]
	\centering
	\scriptsize
	\begin{tabular}{|l|r|*{6}{c|}} 	
		\hline		
		 \multirow{2}{2.5em}{Method} & Cost  & \multicolumn{2}{c|}{CIFAR-10}  & \multicolumn{2}{c|}{CIFAR-100}   & \multicolumn{2}{c|}{ImageNet16-120}   \\
		\cline{3-8}
		& (hrs) & valid & test & valid & test & valid & test \\
		\hline
		
		DARTS \cite{liu2018darts} & 3.2 & 39.77$\pm$0.00 & 54.30$\pm$0.00 & 15.03$\pm$0.00 & 15.61$\pm$0.00 & 16.43$\pm$0.00 & 16.32$\pm$0.00 \\
		RSPS \cite{li2020random} & 2.2 & 80.42$\pm$3.58 & 84.07$\pm$3.61 & 52.12$\pm$5.55&52.31$\pm$5.77&27.22$\pm$3.24&26.28$\pm$3.09 \\
		SETN \cite{dong2019one} & 9.5 & 84.04$\pm$0.28 & 87.64$\pm$0.00 & 58.86$\pm$0.06 & 59.05$\pm$0.24 & 33.06$\pm$0.02 & 32.52$\pm$0.21   \\
		GDAS \cite{dong2019searching} &  8.7 & 89.89$\pm$0.08 & \textbf{\underline{93.61$\pm$0.09}} & \textbf{71.34$\pm$0.04} & \textbf{70.70$\pm$0.30} & 41.59$\pm$1.33 & 41.71$\pm$0.98 \\ 
		SNAS \cite{xie2018snas}$^\star$ & - & \textbf{90.10$\pm$1.04} & 92.77$\pm$0.83 & 69.69$\pm$2.39 & 69.34$\pm$1.98 & \textbf{\underline{42.84$\pm$1.79}} & \textbf{\underline{43.16$\pm$2.64}} \\
		DSNAS \cite{hu2020dsnas}$^\star$ & - & 89.66$\pm$0.29 & 93.08$\pm$0.13 & 30.87$\pm$16.40 &  31.01$\pm$16.38 & 40.61$\pm$0.09 & 41.07$\pm$0.09 \\
		PCDARTS\cite{xu2020pcdarts}$^\star$ & -  & 89.96$\pm$0.15 & 93.41$\pm$0.30 & 67.12$\pm$0.39 & 67.48$\pm$0.89 & 40.83$\pm$0.08 & 41.31$\pm$0.22 \\ 
		NoisyDARTS& 3.2 & \textbf{\underline{90.26$\pm$0.22}} & \textbf{93.49$\pm$0.25} & \textbf{\underline{71.36$\pm$0.21}} & \textbf{\underline{71.55$\pm$0.51}} & \textbf{42.47$\pm$0.00} & \textbf{42.34$\pm$0.06} \\
		\hline
	\end{tabular}
	\caption{Comparison on NAS-Bench-201. Averaged on 3 searches. The best for  is in bold and underlined, while the second best is in bold. $^\star$: reported by \cite{chen2020drnas}}\smallskip
	\label{tab:bench201}
\end{table*}

\paragraph{Searching GCN on ModelNet10.}
We follow the same setting as SGAS \cite{li2019sgas} to search GCN networks on ModelNet10 \cite{wu20153d} and evaluate them on ModelNet40. Our models (see Figure~\ref{fig:gcn_models} in the supplementary) are on par with  SGAS as reported in Table~\ref{table:sota-modelnet40}.

\subsection{Ablation study}\label{sec:ablation}

\paragraph{With vs without noise.}
We compare the searched models with and without noises on two commonly used search spaces in Table~\ref{table:ablation-exps-cifar-imagenet}.  NoisyDARTS  robustly escapes from the performance collapse across different search spaces and datasets. Note that without noise,  the differentiable approach performs severely worse and obtains only $66.4\%$ top-1 accuracy on the ImageNet classification task. In contrast, our simple yet effective method can find a state-of-the-art model with $76.1\%$. 

\begin{table}
	\centering

		\small
		\begin{tabular}{*{3}{|l}|c|c|}
			\hline
			Method & Type & Dataset & Benchmark & Acc (\%)\\
			\hline
			NoisyDARTS & w/ Noise & CIFAR-10 & 1 & 97.35$\pm$0.23 \\
			DARTS & w/o Noise & CIFAR-10 & 1 &96.62$\pm$0.23$^\star$  \\
			NoisyDARTS & w/ Noise & ImageNet & 2&  76.1 \\
			DARTS & w/o Noise &ImageNet & 2  &66.4\\
			\hline
		\end{tabular}
	\caption{NoisyDARTS is robust across CIFAR-10 and ImageNet. $^\star$: Reported by \cite{Yu2020Evaluating}}
	\label{table:ablation-exps-cifar-imagenet}
\end{table}

\paragraph{Noise vs. Dropout}
Dropout \cite{hinton2012improving} can be regarded as a type of special noise, which is originally designed to avoid overfitting. We use a special type of Dropout: DropPath \cite{zoph2017learning} to act as a baseline, which is a drop-in replacement of our noise paradigm. We search on NAS-Bench-201 using different DropPath rates $r_{drop} \in $  \{0.1, 0.2\} and report the results in Table~\ref{table:droppath-bench}. It appears that Dropout produces much worse results.  

\begin{table}[h]
	\setlength{\tabcolsep}{1pt}
	\centering
	\small
		\begin{tabular}{|*{7}{c|}}
		\hline
		$r_{drop}$ &\multicolumn{2}{c|}{CIFAR-10}  & \multicolumn{2}{c|}{CIFAR-100}  & \multicolumn{2}{c|}{ImageNet-16}  \\
		\cline{2-7}
		& val & test & val & test & val & test \\
		\hline	
		0.1 & 63.10$\pm$17.68 &66.02$\pm$18.63 & 38.66$\pm$15.72 & 38.75$\pm$15.72 & 18.59$\pm$11.61 &18.05$\pm$11.47 \\ 
		0.2 & 51.54$\pm$0.00 & 55.15$\pm$0.00 &28.74$\pm$0.00 &28.86$\pm$0.00 &11.60$\pm$0.00 &10.87$\pm$0.00  \\
	   ours &\textbf{90.26$\pm$0.22} & \textbf{93.49$\pm$0.25} & \textbf{71.36$\pm$0.21} & \textbf{71.55$\pm$0.51} & \textbf{42.47$\pm$0.00} & \textbf{42.34$\pm$0.06} \\
		\hline
		\end{tabular}
	\caption{Applying Drop-path in replace of noise on NAS-Bench-201.}
	\label{table:droppath-bench}
\end{table}

\paragraph{Zero-mean (unbiased) noise  vs. biased noise.}
Experiments in Table~\ref{table:ablation-exps-cifar-imagenet} and \ref{table:ablation-exps-cifar} verify the necessity of the unbiased design, otherwise it brings in a deterministic bias.  We can observe that the average performance of the searched models decreases while the bias increases. Eventually, it fails to overcome the collapse problem because larger biases overshoot the gradient and misguide the whole optimization process. 
\section{Discussion about the Single-point Hessian eigenvalue}\label{sec:hessian}

According to \cite{zela2020understanding}, the collapse is likely to occur when the maximal eigenvalue $\lambda_{max}^{\alpha}$ increases rapidly (whose Hessian matrix is calculated only on a snapshot of $\alpha$, i.e. single-point), under which condition some early stopping strategy was involved to avoid the collapse. To verify their claim, we search with DARTS and NoisyDARTS  across 7 seeds and plot the calculated Hessian eigenvalues per epoch in Figure~\ref{fig:hessian_norm_comparison}.

\begin{figure}
	\centering

\centering	\includegraphics[width=0.3\columnwidth,height=3.5cm]{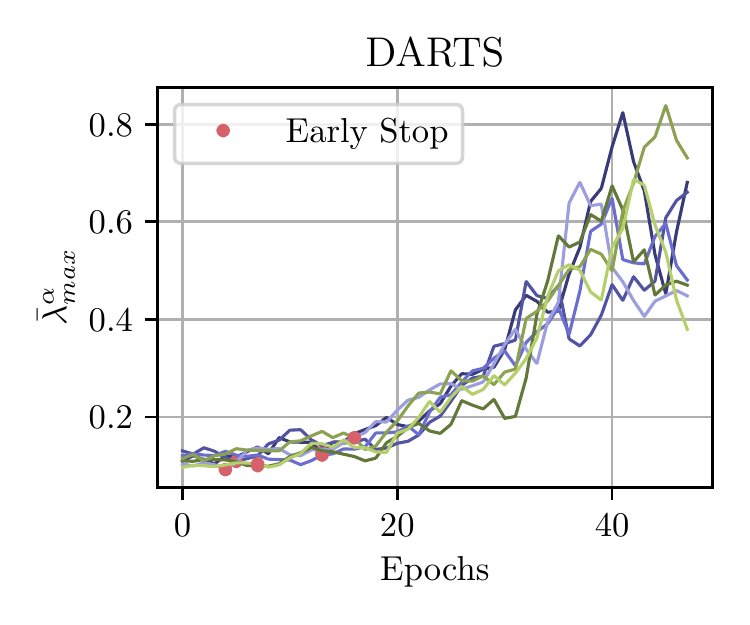}
		\includegraphics[width=0.3\columnwidth,height=3.5cm]{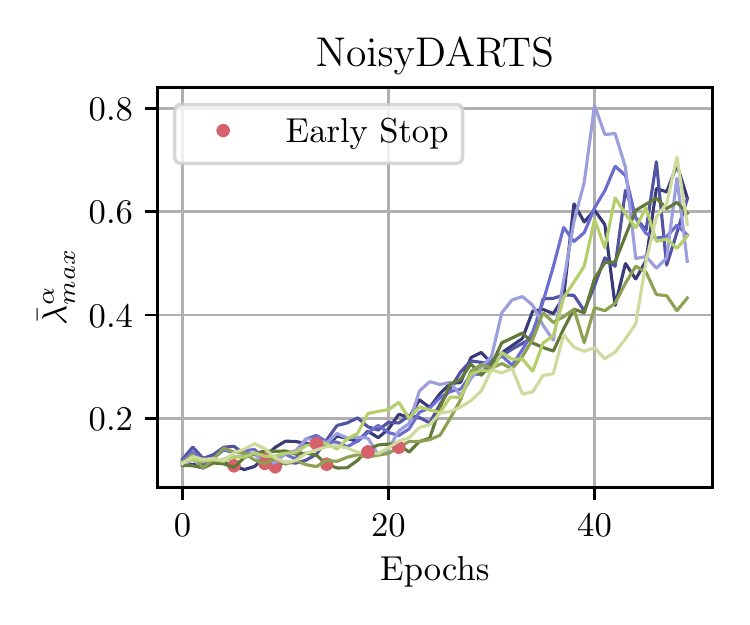}
		\includegraphics[width=0.3\columnwidth,height=3.5cm]{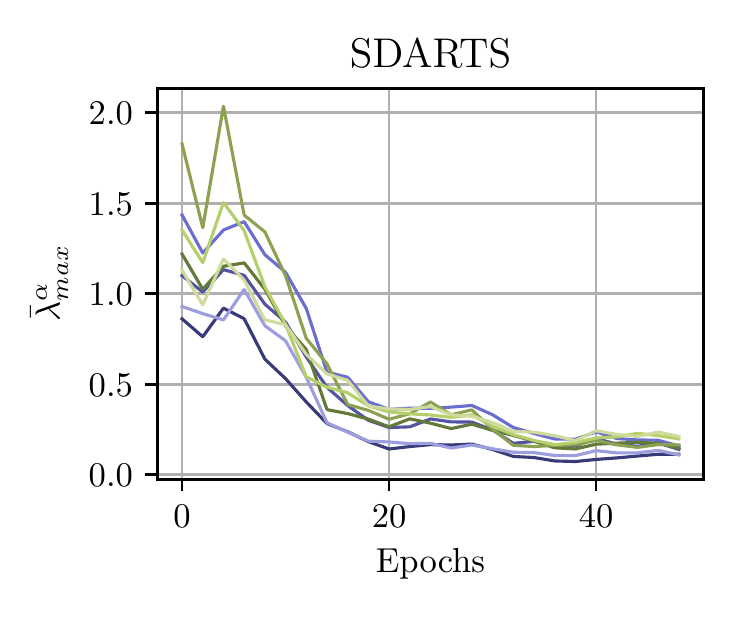}
			\vskip -0.2in

	\caption{Smoothed maximal Hessian eigenvalues $\bar{\lambda}_{max}^{\alpha}$ \cite{zela2020understanding} during the optimization on CIFAR-10.  \cite{zela2020understanding} suggests that the optimization should stop early at the marked points. SDARTS \cite{chen2020stabilizing} regularizes $\bar{\lambda}_{max}^{\alpha}$ while searching. However, without doing so, we don't see the collapse in NoisyDARTS. DARTS has an average accuracy  of $96.9 \%$ while we have $97.35 \%$.} 
	\label{fig:hessian_norm_comparison}
			\vskip -0.2in
\end{figure}

Remarkably, both DARTS and our method show a similar trend. We continue to train the supernet whilst eigenvalues keep increasing, but we still derive mostly good models with an average accuracy of 97.35\%. It's surprising to see that no obvious collapse occurs. Although the Hessian eigenvalue criterion benefits the elimination of bad models \cite{zela2020understanding}, it seems to mistakenly reject good ones. We also find the similar result on CIFAR-100, see Figure~ \ref{fig:hessian_norm_comparison_cifar100}. 

We observe similar results  in the reduced space too (see Section \ref{sec: more_dis_hessian} in the supplementary).  \textbf{We think that a single-point Hessian eigenvalue indicator at a local minimum cannot represent the curvatures of its wider neighborhood. It requires the wider landscape be smoother to avoid the collapse}. It is more clearly shown in Figure~\ref{fig:landscape-comparison}, where NoisyDARTS has a tent-like shape that eases the optimization. 

\textbf{Comparison with SDARTS.}  SDARTS \cite{chen2020stabilizing}, which performs perturbation on architectural weights to implicitly regularize the Hessian norm. However, we inject noise only into the skip connections' output features or to all candidate operations, which suppresses the unfair advantage by disturbing the overly fluent gradient flow. Moreover, our method is efficient  and nearly no extra cost is required. In contrast, SDARTS-ADV needs \textbf{2$\times$} search cost than ours. 

Our method differs from SDARTS \cite{chen2020stabilizing} in Hessian eigenvalue trend, as shown in Figure~\ref{fig:hessian_norm_comparison}.  SDARTS enjoys decreasing hessian eigenvalues while ours can have growing ones. The validation landscape of SDARTS is shown in Figure \ref{fig:landscape-comparison}. SDARTS has a rather carpet-like landscape. It seems that too flat landscape of SDARTS may not correspond to a good model.

\begin{figure}[ht]
	\centering
	\subfigure[DARTS]{
		\includegraphics[width=0.24\textwidth,scale=1]{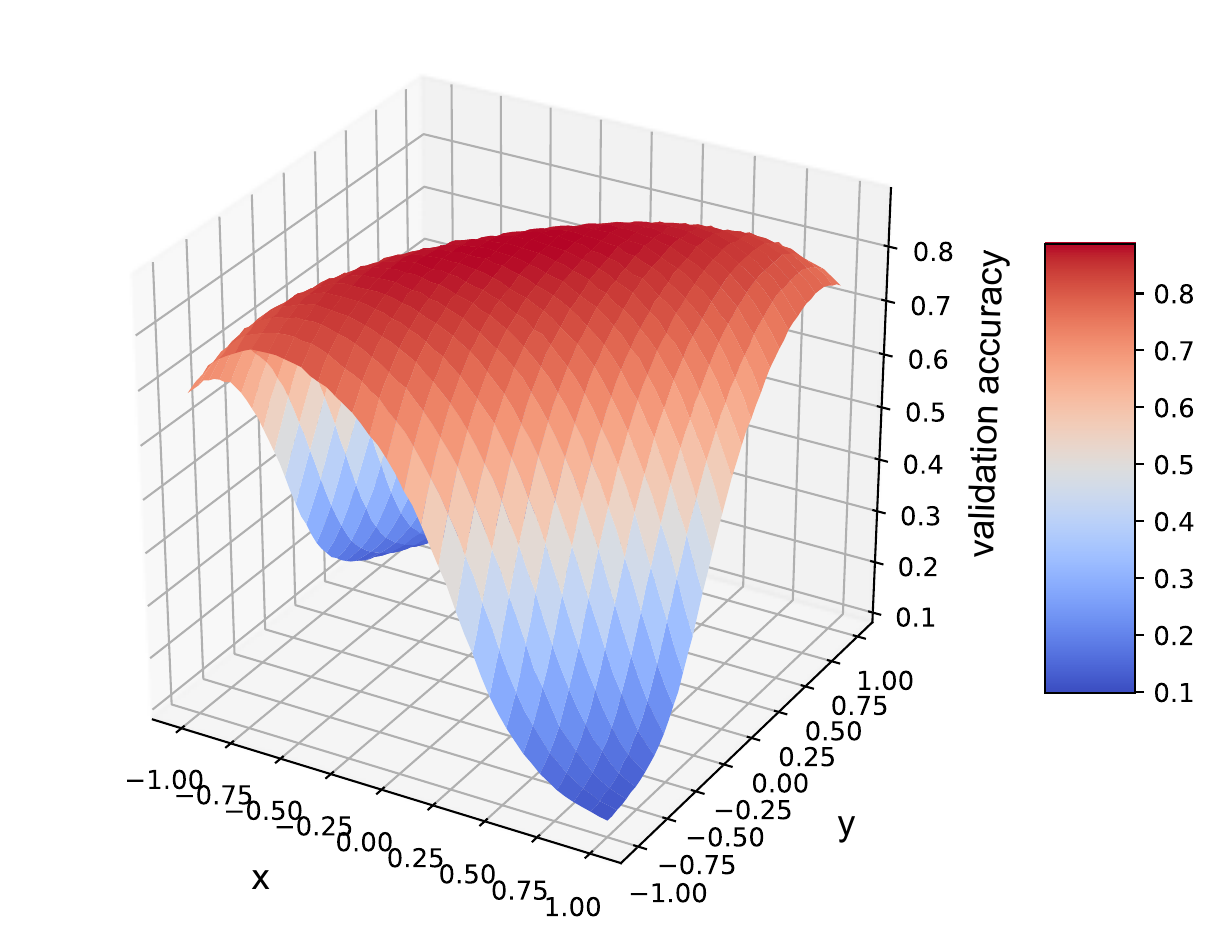}
		\includegraphics[width=0.2\textwidth,scale=1]{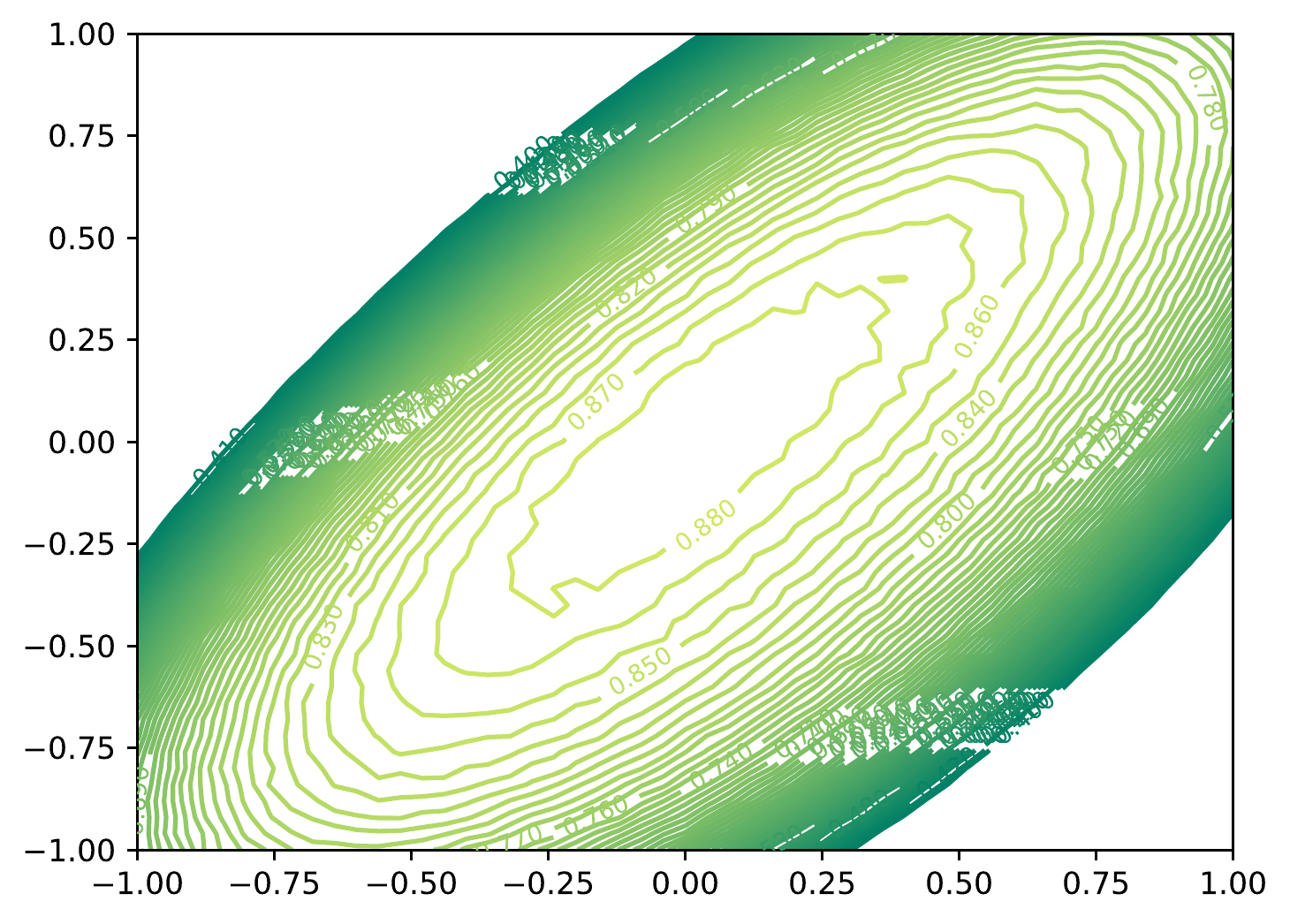}
	}
	\subfigure[SDARTS]{
		\includegraphics[width=0.24\textwidth]{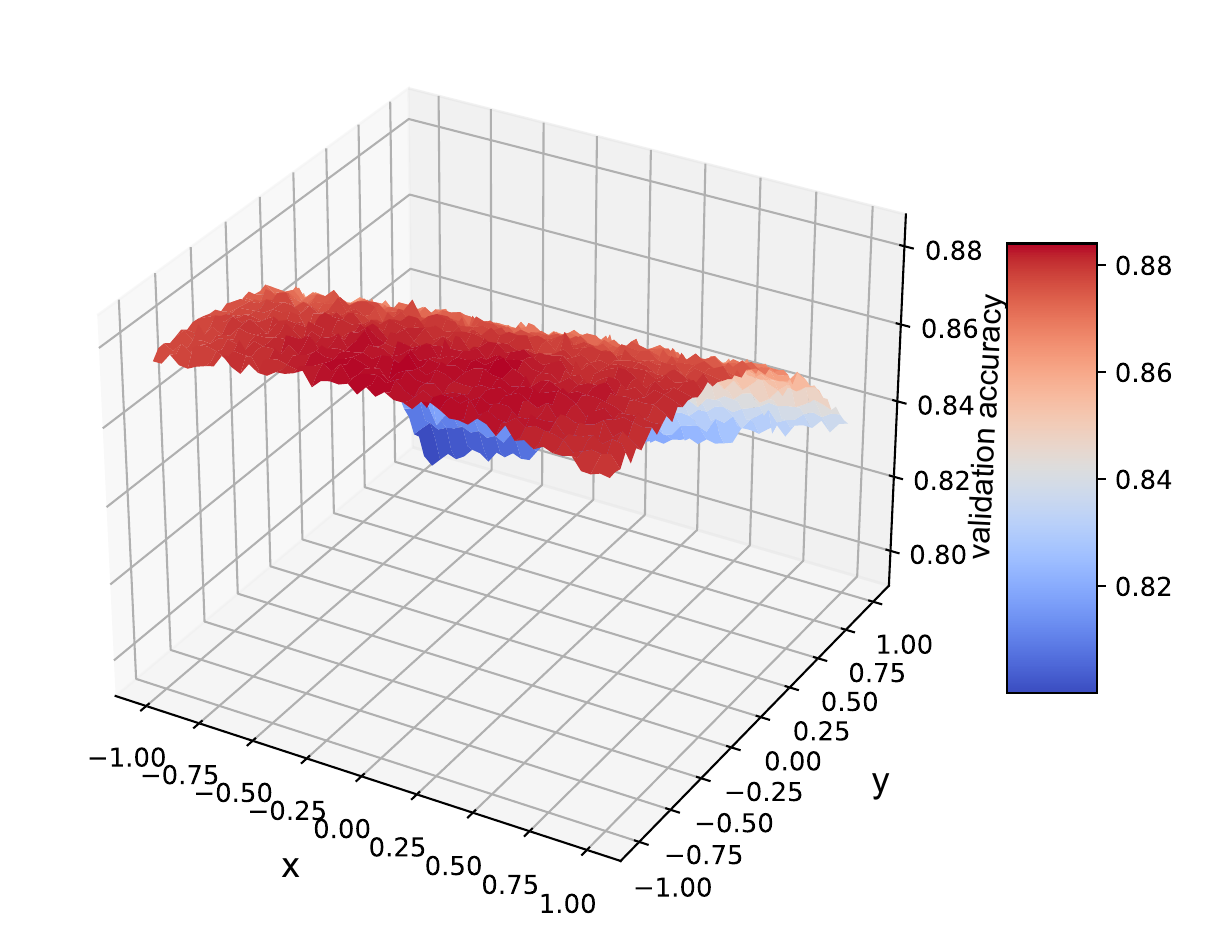}
		\includegraphics[width=0.2\textwidth]{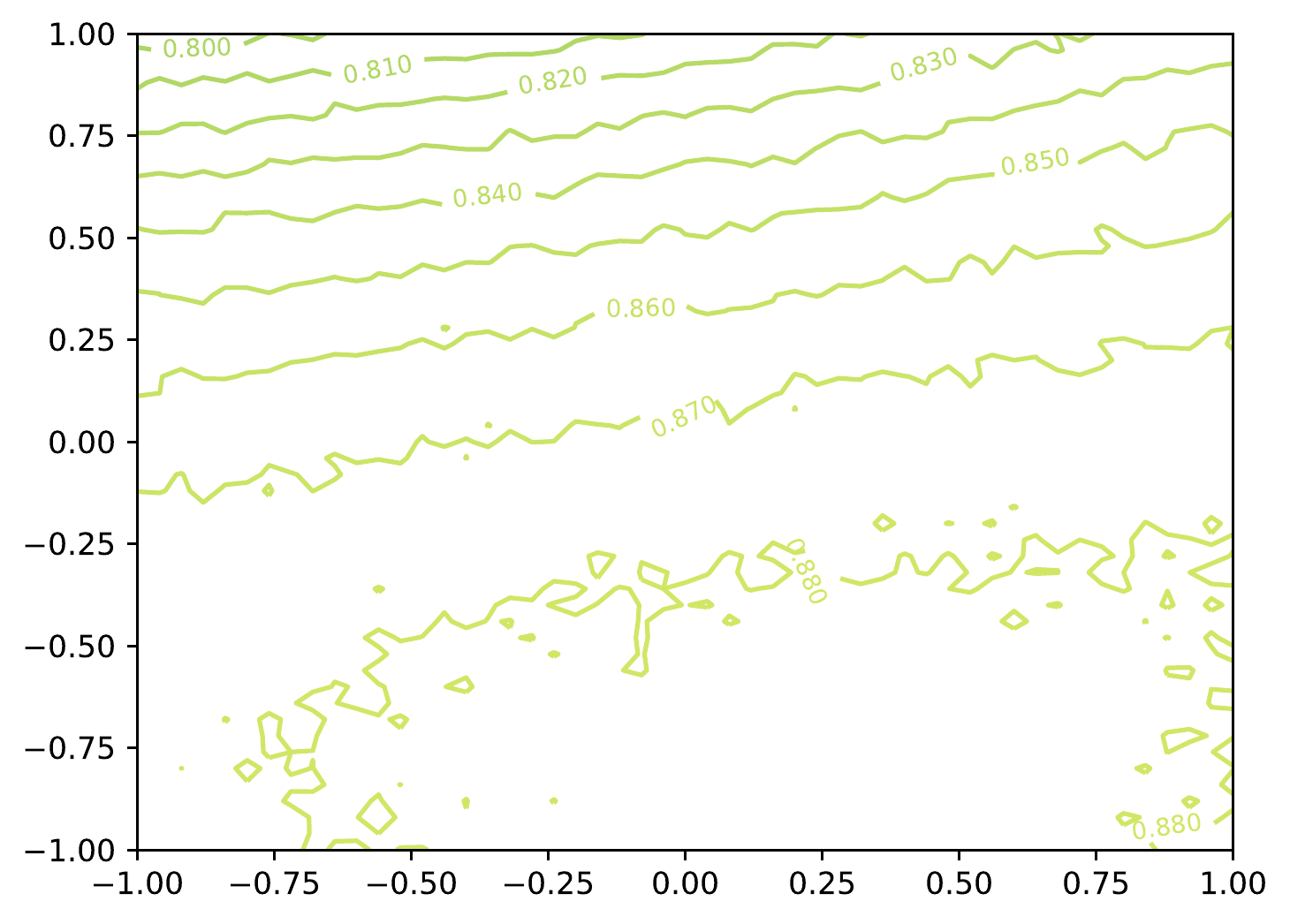}
	}
	\subfigure[NoisyDARTS, NFA]{
		\includegraphics[width=0.24\textwidth]{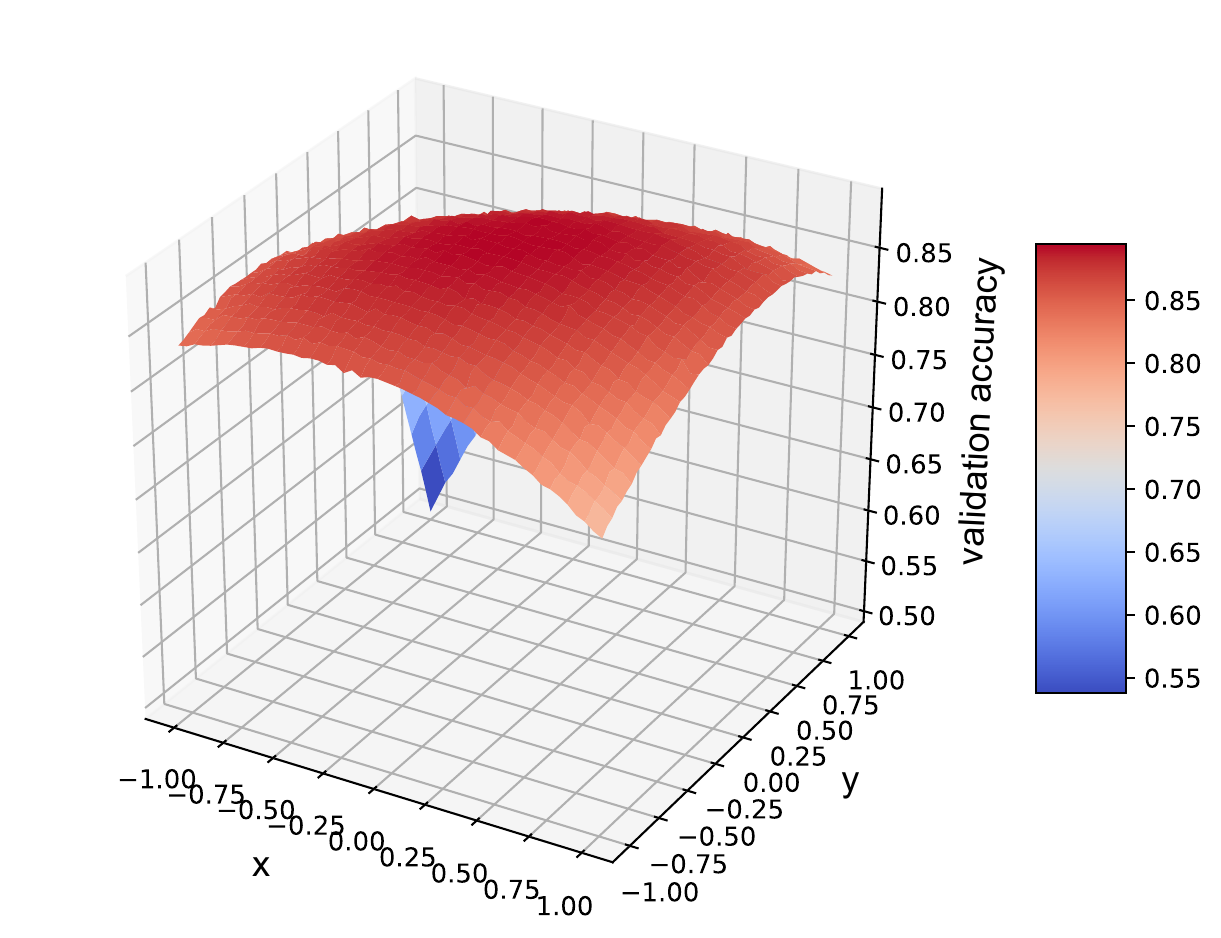}
		\includegraphics[width=0.2\textwidth]{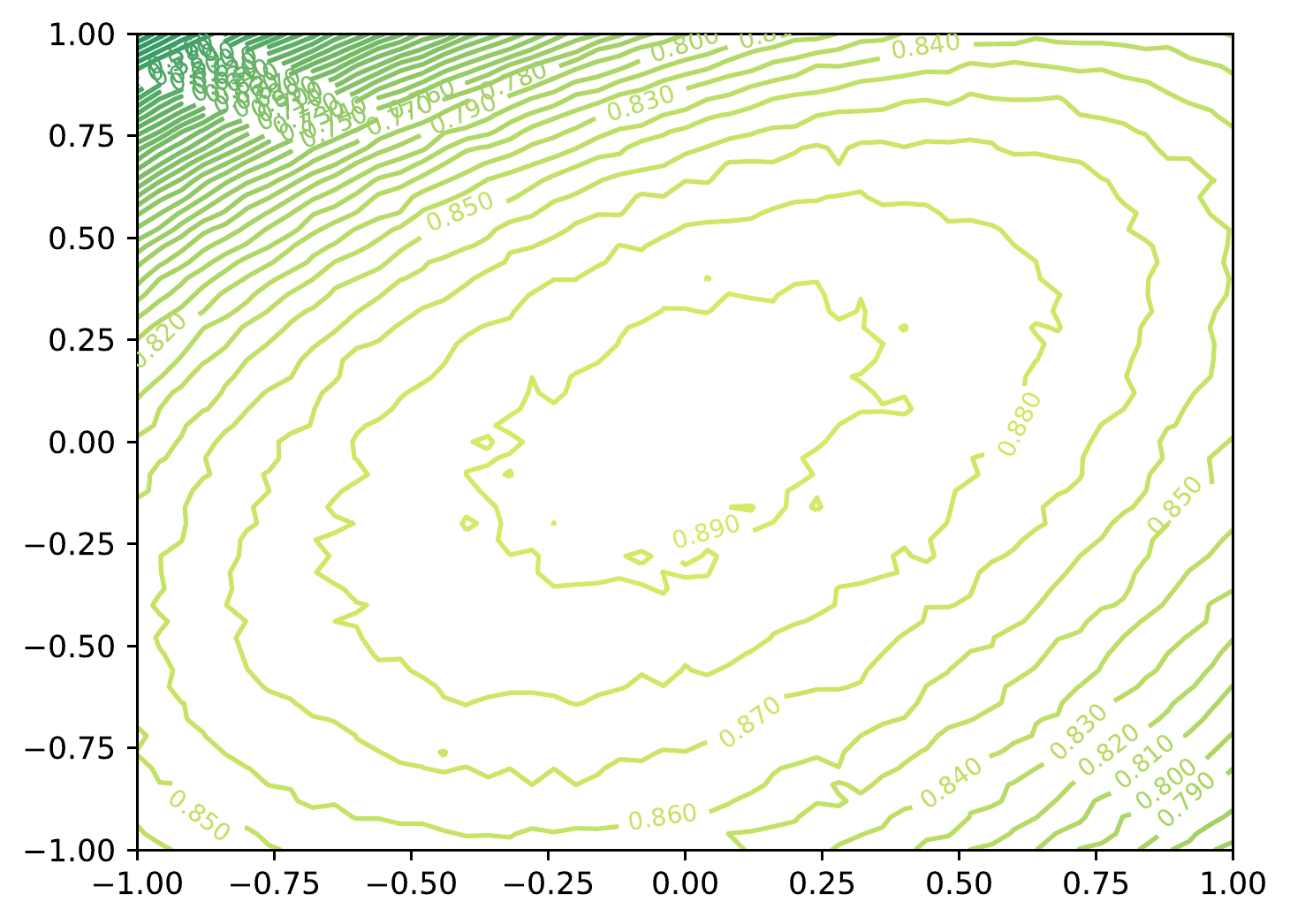}
	} 
	\subfigure[NoisyDARTS, OFS]{
		\includegraphics[width=0.24\textwidth]{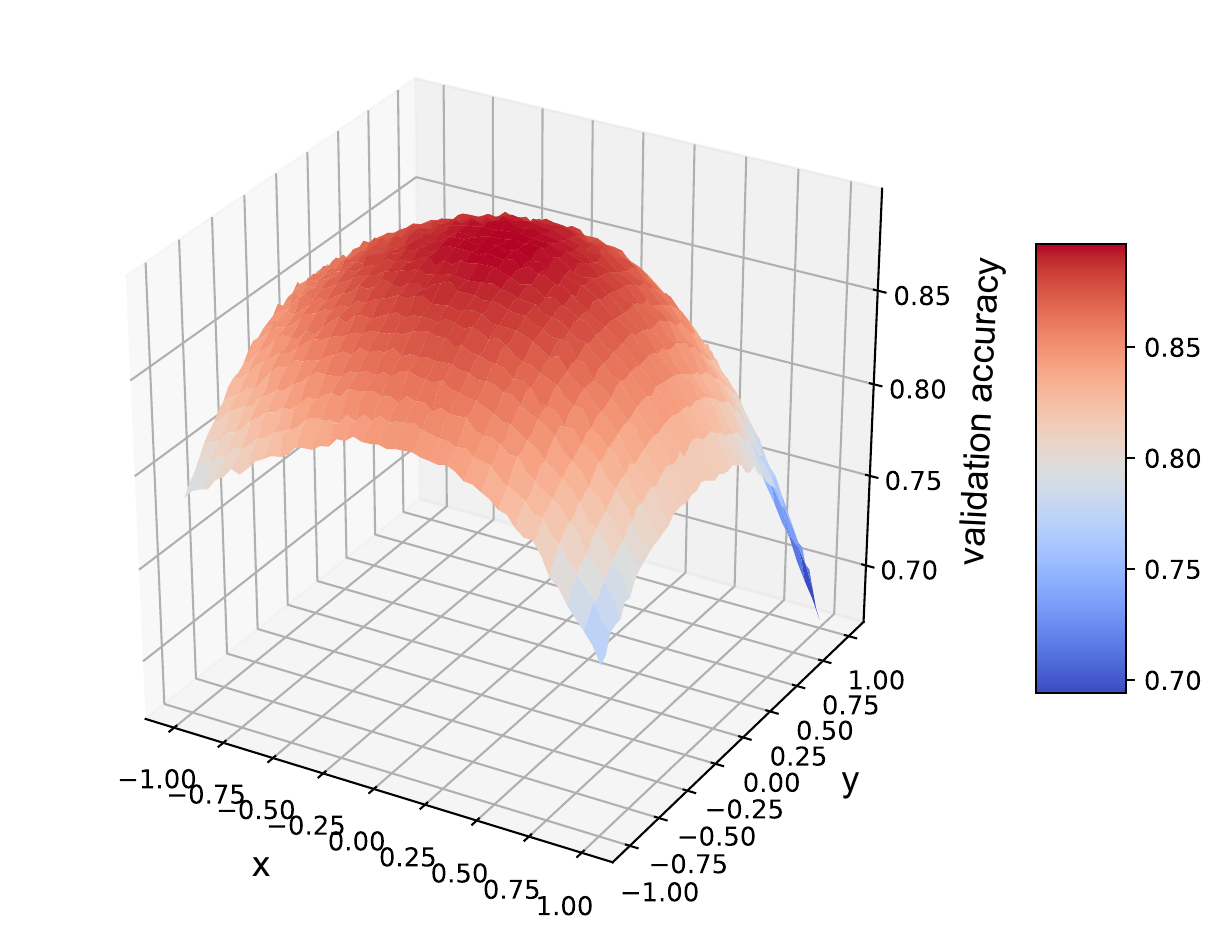}
		\includegraphics[width=0.2\textwidth]{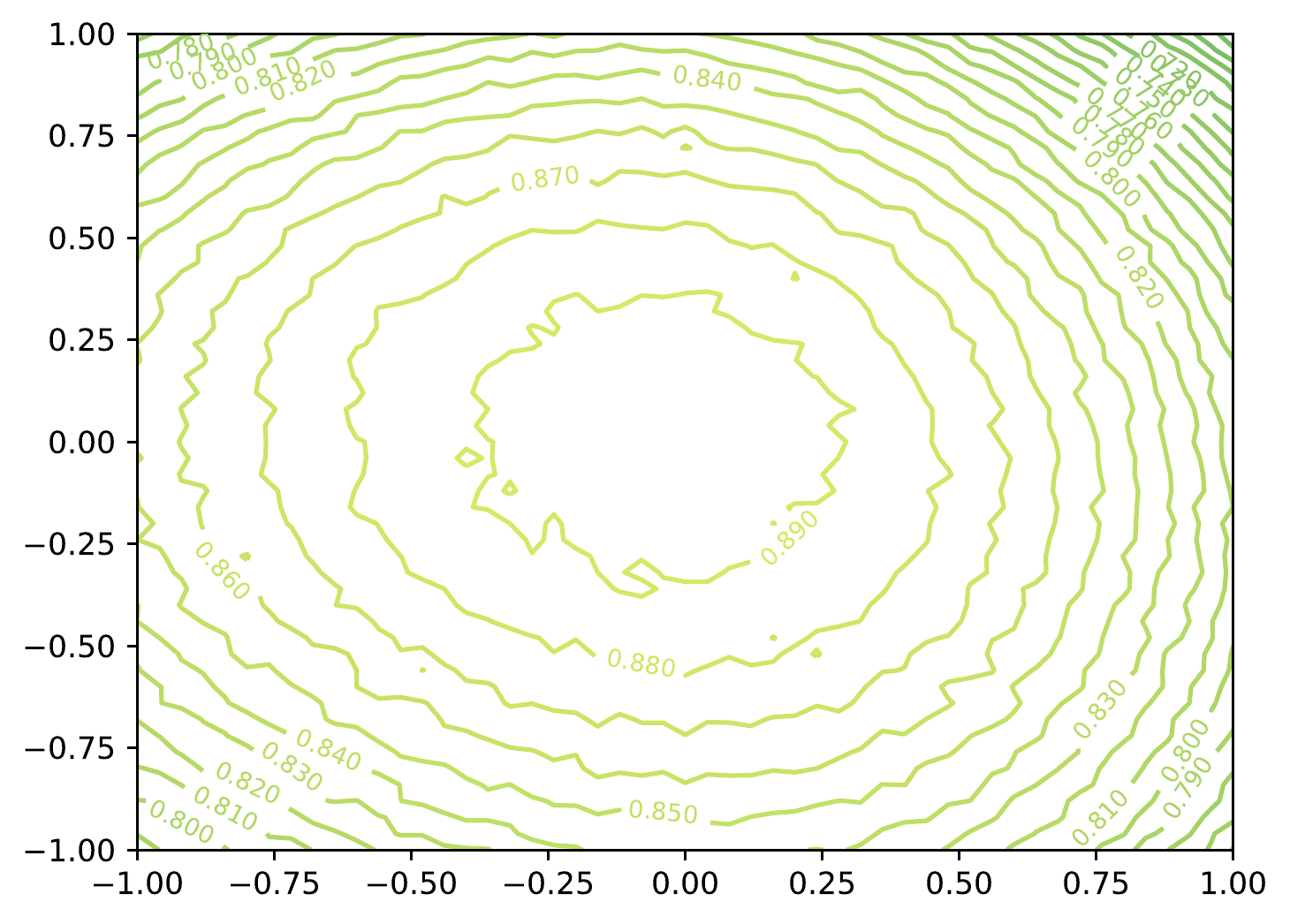}
	}
	\caption{The landscape of validation accuracy w.r.t the architectural weights on CIFAR-10 and the corresponding contours. Following  \cite{chen2020stabilizing}, axis $x$ and $y$ are orthogonal gradient direction of validation loss w.r.t. architectural parameters $\alpha$, axis $z$ refers to the validation accuracy. The related stand-alone model accuracies and Hessian eigenvalues are 96.96\%/0.3388, 97.21\%/0.1735, 97.42\%/0.4495 respectively. } 
	\label{fig:landscape-comparison}
\end{figure} 

\begin{figure}[ht]
	\centering
	\includegraphics[width=0.3\columnwidth,height=3.5cm]{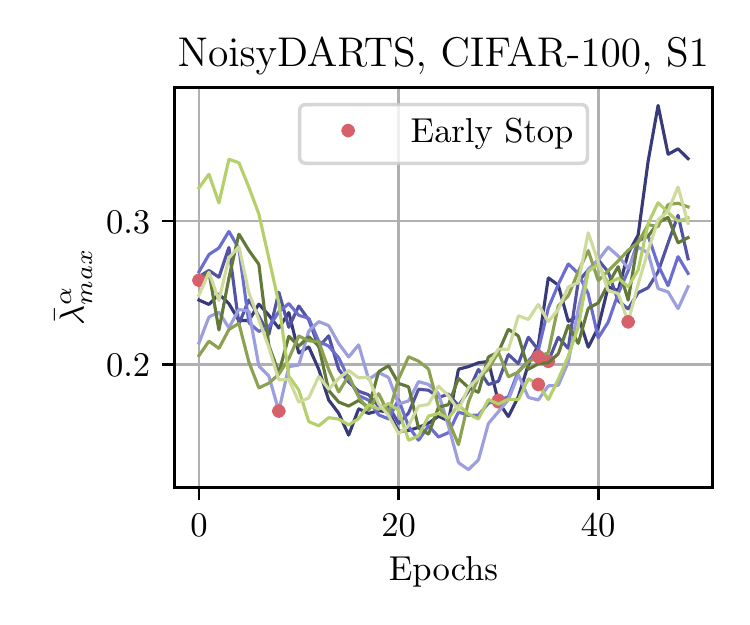}
	\includegraphics[width=0.3\columnwidth,height=3.5cm]{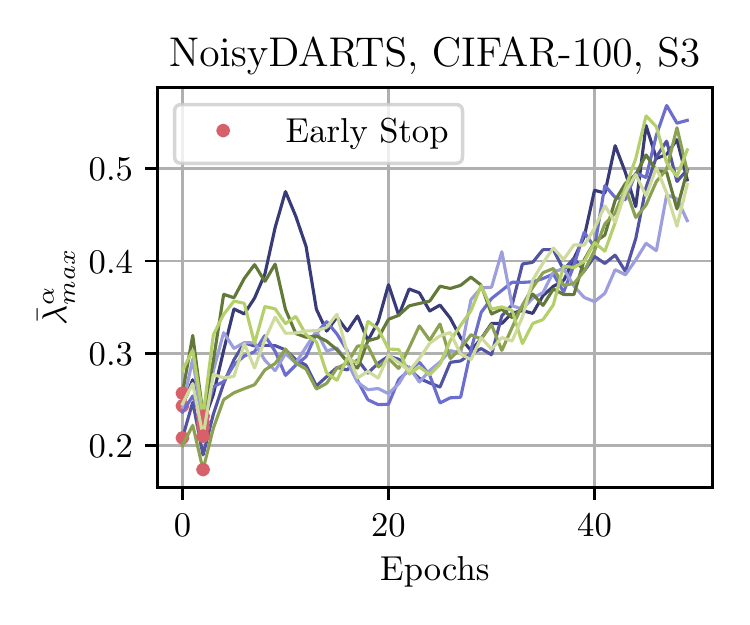}
	\includegraphics[width=0.3\columnwidth,height=3.5cm]{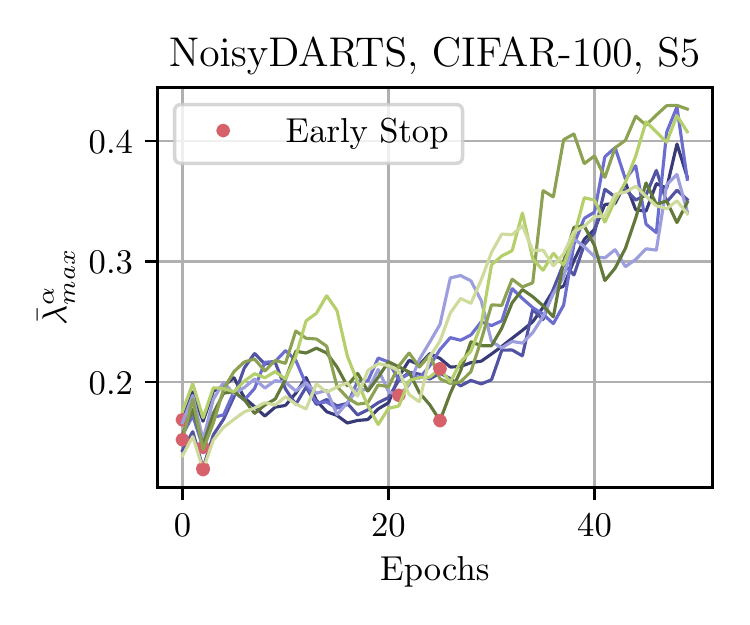}
			\vskip -0.2in
	\caption{Smoothed maximal Hessian eigenvalues $\bar{\lambda}_{max}^{\alpha}$ \cite{zela2020understanding} during the optimization on CIFAR-100 in reduced search spaces $S_1$, $S_3$, $S_5$ from \cite{zela2020understanding}. We observe the similar growing trend. Notwithstanding, we achieve a state-of-the-art 16.28\% test error rate in $S_5$.} 
	\label{fig:hessian_norm_comparison_cifar100}
\end{figure}

\section{Conclusion}

In this paper, we proposed a novel approach NoisyDARTS, to robustify differentiable architecture search. By injecting a proper amount of unbiased noise into candidate operations, we successfully let the  optimizer be perceptible about the disturbed gradient flow. As a result, the unfair advantage is largely attenuated, and the derived models generally enjoy improved performance. Experiments show that NoisyDARTS can work effectively and robustly, regardless of noise types. We  achieved state-of-the-art results on several  datasets and search spaces with low $CO_2$ emissions.

While most of the current approaches addressing the fatal collapse focus on designing various criteria to avoid stepping into the failure mode, our method stands out of the existing framework and no longer put hard limits as in \cite{chen2019progressive,liang2019darts}. We review the whole optimization process to find out what leads to the collapse and directly control the unfair gradient flow, which is more fundamental than a stationary failure point analysis. We hope this would bring a novel insight for the NAS community to shift attention away from criteria-based algorithms.

\clearpage
\bibliography{egbib.bib}

\newpage
\appendix

\section{Analysis of multiplicative noise}\label{supp:multi-noise}
We set the output of skip connection with  multiplicative noise is $x' = x \cdot \tilde{x} $, where $\tilde{x}$ is sampled from a certain distribution. Similar to Section~\ref{theory} (main text), the expectation of the gradient under multiplicative noise can be written as:
\begin{equation}{\label{eq:supp-gradient}}
\begin{aligned}
{\Bbb E} \left[\nabla_{skip} \right] &=
{\Bbb E} \left[\frac{\partial \cal L}{\partial{y}} \frac{\partial f(\alpha_{s})}{\partial{\alpha_{s}}} \left(x' \right) \right]  \\
&\approx \frac{\partial \cal L}{\partial{y^\star}}  \frac{\partial f(\alpha_{s})}{\partial{\alpha_{s}}} 
\left(x \cdot {\Bbb E}\left[\tilde{x} \right] \right).
\end{aligned}
\end{equation}
Again notice that taking  $\frac{\partial \cal L}{\partial{y^\star}} $ out of the expectation in Equation~\ref{eq:supp-gradient} requires Equation~\ref{eq:approximate} (main text) be satisfied. 
To keep the gradient unbiased, $\tilde{x}$ should be close to 1. Thus, we use Gaussian distribution $\tilde{x} \sim {\cal N}(1, \sigma^2)$.

\section{Algorithm}\label{app:alg}
\begin{algorithm}[H]
	\caption{NoisyDARTS-OFS  (default and recommended)}
	\label{alg:noisy-darts-ofs}
	\begin{algorithmic}[1]
		\STATE {\bfseries Input:} Architecture parameters $\alpha_{i,j}$, network weights $w$ , noise's standard variance $\sigma$, $Epoch_{max}$.
		\WHILE{\emph{not reach $Epoch_{max}$}}
		\STATE 		 Inject random Gaussian noise $\tilde{x}$ into the skip connections' output.
		\STATE Update weights $w$ by  $\nabla_{w}{\cal L}_{train}(w, \alpha)$
		\STATE Update architecture parameters $\alpha$  by  $\nabla_{\alpha }{\cal L}_{val}(w, \alpha)$
		\ENDWHILE
		\STATE Derive the final architecture according to learned $\alpha$.
	\end{algorithmic}
	\vskip -0.05in
\end{algorithm}

We give the NFA version of NoisyDARTS in Algorithm~\ref{alg:noisy-darts-nfa}.
\begin{algorithm}[H]
			\caption{NoisyDARTS-NFA}
			\label{alg:noisy-darts-nfa}
			\begin{algorithmic}[1]
				\STATE {\bfseries Input:} Architecture parameters $\alpha_{i,j}$, network weights $w$ , noise's standard variance $\sigma$, $Epoch_{max}$.
				\WHILE{\emph{not reach $Epoch_{max}$}}
				\STATE 		 Inject random Gaussian noise $\tilde{x}$ into all candidate operations' output.
				\STATE Update weights $w$ by  $\nabla_{w}{\cal L}_{train}(w, \alpha)$
				\STATE Update architecture parameters $\alpha$  by  $\nabla_{\alpha }{\cal L}_{val}(w, \alpha)$
				\ENDWHILE
				\STATE Derive the final architecture according to learned $\alpha$.
			\end{algorithmic}
		\end{algorithm}

\section{More Experiments and Details}\label{supp:exp}

\subsection{More  Ablation Studies}\label{sec:more_ablation}

\textbf{Gaussian noise  vs. uniform noise\quad} According to the analysis of Section~\ref{theory} in the main text, unbiased Gaussian noise is an appropriate choice that satisfies Equation~\ref*{eq:gradient}. In the same vein, unbiased uniform noise should be equally useful. We compare both types of noise in terms of effectiveness in Table~\ref{table:ablation-noise-type}. Both have improved performance while Gaussian is slightly better. This can be loosely explained. As the output feature $x$ from each skip connection tends to be Gaussian, i.e. $x \sim \mathcal{N}(\mu_1, \sigma_1^2)$ (see Figure~\ref{fig:skip-feature-hist}), a Gaussian noise $\tilde{x} \sim \mathcal N(0, \sigma_2^2)$ is preferred since the additive result shares the similar statistics, i.e., $x + \tilde{x} \sim N(\mu_1, \sigma_1^2 + \sigma_2^2)$.

\begin{table}[ht]
	\centering
	\small
	\begin{tabular}{*{4}{|c}|}
		\hline
		Noise Type &  $\mu$ & $\sigma$  & Avg. Top-1 (\%)\\
		\hline
		w/o Noise & - &- & 97.00$\pm$0.14 \\
		Gaussian & 0.0 &  0.1 &  97.21$\pm$0.21 \\ 
		Uniform & 0.0 &  0.1 & 97.12$\pm$0.15  \\ 
		\hline
		Gaussian &  0.0  & 0.2 &  97.35$\pm$0.23 \\ 
		Uniform &  0.0 & 0.2  & 97.15$\pm$0.23 \\ 
		\hline
	\end{tabular}
	\caption{Experiments on different types of noise. Each search is run 8 times}
	\label{table:ablation-noise-type}
\end{table}

\begin{figure*}[ht]
	\centering
	\includegraphics[width=\textwidth]{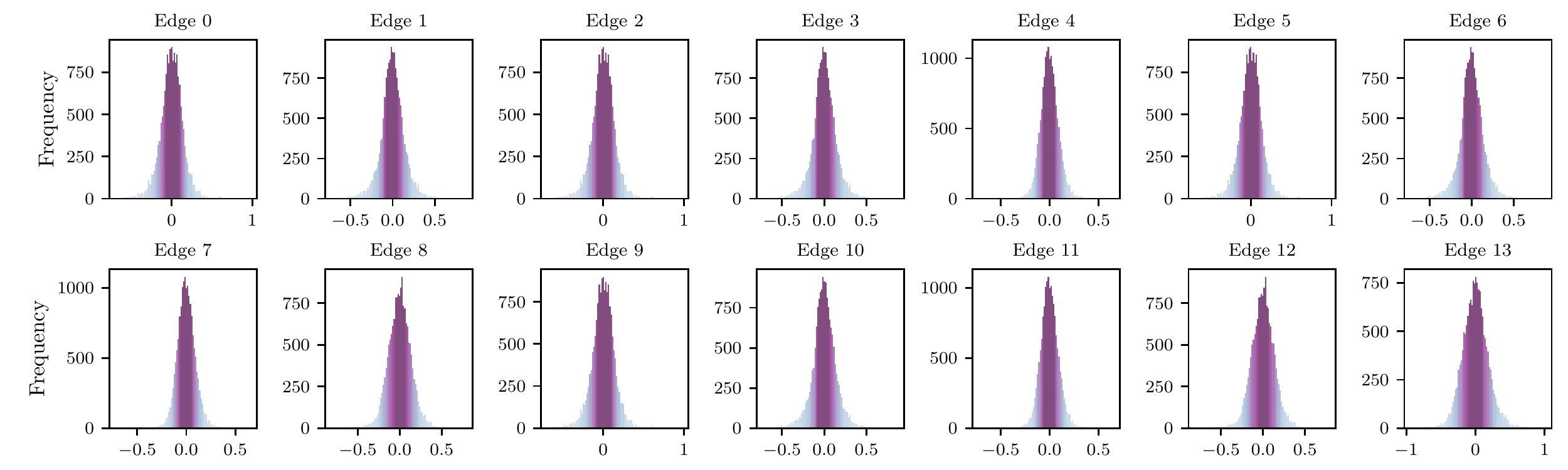}
	\caption{The Gaussian-like distribution of output features on all \emph{skip} edges in the original DARTS.}
	\label{fig:skip-feature-hist}
\end{figure*}

\paragraph{Additive noise vs. multiplicative noise}
Apart from additive noise, we also blend the noise ($\mu=1$) by multiplying it with the output $x$ of skip connections, which is approximately effective as additive noise, see Table~\ref{table:ablation-noise-mix}. 
In general, we also notice that searching with the biased noise also outperforms DARTS. This could be empirically interpreted as that resolving the aggregation of skip connections is more critical, while a slight deviation during the optimization matters less.	

\begin{table}[ht]
	\centering
	\small
	\begin{tabular}{*{4}{|c}|}
		\hline
		Noise Mixture &  $\mu$ & $\sigma$  & Top-1 (\%)\\
		\hline
		w/o Noise & - &- & 97.00$\pm$0.14 \\
		Additive &  0.0 & 0.1  & 97.21$\pm$0.21  \\ 
		Multiplicative & 1.0 & 0.1  & 97.15$\pm$0.23 \\ 
		\hline
		Additive  & 0.0 & 0.2 & 97.35$\pm$0.23 \\
		Multiplicative & 1.0 & 0.2 & 97.22$\pm$0.23  \\ 
		\hline
	\end{tabular}
	\caption{Experiments on different mixing operations. Each search is run 8 times}
	\label{table:ablation-noise-mix}
\end{table}

\begin{table}
	\centering
	\small
	\setlength{\tabcolsep}{2.5pt}
	\begin{tabular}{|l*{3}{|r}H|*{3}{|r}H|}
		\hline
		Type &  $\mu$ & $\sigma$ & Acc (\%)  & Best (\%) & $\mu$ & $\sigma$ & Acc (\%) & \\
		\hline	
		Gaussian & -1.0 &  0.1 &  96.92$\pm$0.36 & 97.49 & -1.0 &  0.2  &  97.14$\pm$0.15 & 97.35\\ 
		Gaussian & -0.5 &  0.1 &  97.13$\pm$0.20 & 97.36 & -0.5 &  0.2  & 97.07$\pm$0.12 & 97.33\\		
		Gaussian & 0.0 &  0.1  & \textbf{97.21$\pm$0.21} & \textbf{97.53} & 0.0 &  0.2 & \textbf{97.35$\pm$0.23} & \textbf{97.63}\\ 
		Gaussian & 0.5 &  0.1  & 97.02$\pm$0.21  & 97.28 & 0.5 &  0.2 & 97.16$\pm$0.15& 97.49\\ 
		Gaussian & 1.0 &  0.1  &  96.89$\pm$0.26 & 97.21 &1.0 &  0.2 & 96.82$\pm$0.57 & 97.35 \\ 
		
		\hline
	\end{tabular}
	\caption{Ablation on additive Gaussian noise on CIFAR-10 (each search is run 8 times)}\smallskip
	\label{table:ablation-exps-cifar}
\end{table}

\paragraph{Remove Skip Connection from the search space}
Skip connections are a necessary component but they are troublesome for DARTS. We remove this operation from the NAS-Bench-201 search space to study how well DARTS performs. Table~\ref{table:noskip-bench} hints that DARTS can find relatively competitive architectures (no longer suffering performance collapse), but not as good as those found by state-of-the-art methods in Table~\ref{tab:bench201} in the main text, for instance, it has a CIFAR-10 test accuracy 88.98\% vs. NoisyDARTS's 93.49\%. We suggest that skip connections  play an indispensable role in neural architecture search and have to be carefully dealt with as we did in NoisyDARTS.

\begin{table}[ht]
	\scriptsize
	\centering
		\begin{tabular}{*{7}{|c}|}
		\hline
		\multirow{2}{2.5em}{Setting} & \multicolumn{2}{c|}{CIFAR-10}  & \multicolumn{2}{c|}{CIFAR-100}   & \multicolumn{2}{c|}{ImageNet16-120}  \\
		\cline{2-7}
		 & valid & test & valid & test & valid & test \\
		\hline
		DARTS w/ skip & 39.77$\pm$0.00 & 54.30$\pm$0.00 & 15.03$\pm$0.00 & 15.61$\pm$0.00 & 16.43$\pm$0.00 & 16.32$\pm$0.00 \\
		DARTS w/o skip & 85.67$\pm$1.30	& 88.98$\pm$0.85	& 63.17$\pm$1.00	& 62.82$\pm$1.74	& 33.74$\pm$2.69	& 33.29$\pm$2.66	\\
		NoisyDARTS &  \textbf{90.26$\pm$0.22} & \textbf{93.49$\pm$0.25} & \textbf{71.36$\pm$0.21} & \textbf{71.55$\pm$0.51} & \textbf{42.47$\pm$0.00} & \textbf{42.34$\pm$0.06} \\
		\hline
		\end{tabular}
	\caption{Removing skip connection from search space  on NAS-Bench-201.}
	\label{table:noskip-bench}
\end{table}

\begin{table}
	\setlength{\tabcolsep}{1pt}
	\small
	\centering
	\begin{tabular}{*{1}{|l|}H*{8}{l|}}
		\hline
		Backbones & $\times +$  & Params &Acc    & AP & AP$_{50}$ & AP$_{75}$ & AP$_S$ & AP$_M$ & AP$_L$ \\
		& (M) & (M) & (\%) &(\%) & (\%)& (\%)&(\%) &(\%) &(\%) 
		\\
		\hline
		MobileNetV2 & 300 & 3.4& 72.0 & 28.3 & 46.7 & 29.3 & 14.8 & 30.7 & 38.1\\
		SingPath NAS & 365 & 4.3 & 75.0 & 30.7 & 49.8 & 32.2 & 15.4 &33.9 & 41.6\\
		MobileNetV3  & 219 & 5.4 & 75.2& 29.9 & 49.3 & 30.8 & 14.9 & 33.3 & 41.1\\
		MnasNet-A2  & 340& 4.8 & 75.6 & 30.5 & 50.2 & 32.0 & 16.6 & 34.1 & 41.1\\
		SCARLET-A & 365 & 6.7 & 76.9 & 31.4 & 51.2 & 33.0 &16.3& 35.1 &41.8 \\
		MixNet-M  & 360 & 5.0 & 77.0 & 31.3& 51.7 & 32.4& 17.0 & 35.0 & 41.9   \\
		FairNAS-A & 392 & 5.9 & 77.5 & 32.4 & 52.4 & 33.9 & 17.2 & 36.3 & 43.2\\
		\textbf{NoisyDARTS-A} & 449 & 5.5& \textbf{77.9} & \textbf{33.1}& \textbf{53.4} & \textbf{34.8}& \textbf{18.5} & \textbf{36.6} & \textbf{44.4}  \\
		\hline
	\end{tabular}
	\caption{COCO Object detection of various drop-in backbones.}\smallskip 
	\label{table:fairnas-coco-retina}
\end{table}
\paragraph{Noise For All (NFA) vs. Only For Skip connection (OFS)}
Applying noise to skip connections is not an ad-hoc decision. In theory, we can interfere with the optimization by injecting the noise to any operation. The underlying philosophy is that only those operations that can work robustly against noises will win the race without unfair advantage.  We compare the settings of NFA, ES (noise for all but excluding skip), OFS in Table~\ref{table:allnoise-bench}. \emph{This proves noise injection to skip connection is critical for a better searching performance.} Note that this approach also obtains much better result than DARTS. However, it requires a bit of trial-and-error to control the $\sigma$ when there are many candidate operations. Therefore, if otherwise specified, we use OFS as the default choice throughout the paper.

\begin{table}[ht]
	\setlength{\tabcolsep}{1pt}
	\centering
	\small
	\begin{tabular}{|*{7}{c|}}
		\hline
		Method  &\multicolumn{2}{c|}{CIFAR-10}  & \multicolumn{2}{c|}{CIFAR-100}  & \multicolumn{2}{c|}{ImageNet-16} \\
		\cline{2-7}
		& val & test & val & test & val & test \\
		\hline
		DARTS-V1 & 39.77$\pm$0.00& 54.30$\pm$0.00 & 15.03$\pm$0.00 & 15.61$\pm$0.00 &  16.43$\pm$0.00& 16.32$\pm$0.00 \\
		
		ES, $\sigma$=0.2 & 39.77$\pm$0.00  &54.30$\pm$0.00  &15.03$\pm$0.00  &15.61$\pm$0.00  &16.43$\pm$0.00  &16.32$\pm$0.00 \\
		ES, $\sigma$=0.4 & 49.27$\pm$16.46 &  59.84$\pm$9.59 & 22.87$\pm$13.59 &23.39$\pm$13.48 & 17.24$\pm$1.40 & 17.01$\pm$1.20\\
		NFA & \underline{88.17}$\pm$2.02 & \underline{91.60$\pm$1.74} & \underline{67.71$\pm$2.35} & \underline{68.26$\pm$1.59} & 41.91$\pm$2.00 & \underline{41.57$\pm$2.59}  \\
		OFS &\textbf{90.26$\pm$0.22} & \textbf{93.49$\pm$0.25} & \textbf{71.36$\pm$0.21} & \textbf{71.55$\pm$0.51} & \textbf{42.47$\pm$0.00} & \textbf{42.34$\pm$0.06} \\
		\hline
	\end{tabular}
	\caption{Comparison of NoisyDARTS (NFA, ES, OFS) on NAS-Bench-201.}
	\label{table:allnoise-bench}
\end{table}

\textbf{Searching GCN Architectures on ModelNet10.}\label{sec:gcn}
\begin{table}
	\setlength{\tabcolsep}{1.5pt}
	\small
	\centering
	\begin{tabular}{|l|c|c|}
		\hline
		Methods & Params (M) & OA (\%) \\
		\hline
		SGAS (Cri. 1 avg.) & 8.78 & 92.69$\pm$0.20\\ 
		SGAS (Cri. 1 best) & 8.63 & 92.87\\
		NoisyDARTS ($\sigma=0.3$ avg.) & 8.68 & 92.85$\pm$0.36 \\
		NoisyDARTS ($\sigma=0.3$ best) & 8.33 & 93.11 \\
		NoisyDARTS ($\sigma=0.4$ avg.) & 8.68 & 92.70$\pm$0.43 \\
		NoisyDARTS ($\sigma=0.4$ best) & 8.93 & 93.23 \\
		\hline
	\end{tabular}
	\caption{3D  classification on ModelNet40. OA: overall accuracy}\smallskip 
	\label{table:sota-modelnet40}
\end{table}

\subsection{Training Setting on Transferred Results on CIFAR-10} \label{sec:cifar10-training}
For transferred learning, we train the ImageNet-pretrained NoisyDARTS-A on CIFAR-10 for 200 epochs with a batch size of 256 and a learning rate of 0.05. We set the weight decay to be 0.0, a dropout rate of 0.1 and a drop connect rate of 0.1. In addition, we also use AutoAugment  as  \cite{tan2020mixconv}.

\subsection{Training Results on CIFAR-100}\label{sec:cifar100}

We show NoisyDARTS models searched in the DARTS search space and trained on CIFAR-100 in Table~\ref{tab:comparison-cifar100}. We set the initial channel as 36 and the number of layers as 20. 

\begin{table}[ht]
\setlength{\tabcolsep}{1pt}
	\centering
	\small
			\begin{tabular}{|l|lHll|} 	
				\hline		
			 Models   &  Params  &  FLOPs &  Error  &  Cost \\
				 & \scriptsize{(M)}  & \scriptsize{(M)}  & \scriptsize{(\%)} & \scriptsize{GPU Days}   \\
				\hline
				ResNet \cite{he2016deep}   &  1.7 &     &   22.10$^\diamond$  &  -  \\
				AmoebaNet \cite{real2019regularized} & 3.1 & &  18.93$^\diamond$ & 3150  \\
				PNAS \cite{liu2018progressive} & 3.2 & & 19.53$^\diamond$ & 150 \\
				ENAS \cite{pham2018efficient}  &  4.6  &    &  19.43$^\diamond$  & 0.45    \\
				DARTS \cite{liu2018darts}  & - & & 20.58$\pm$0.44$^\star$ &  0.4  \\ 
				GDAS \cite{dong2019searching}  &  3.4  &    &  18.38  & 0.2 \\

				P-DARTS \cite{chen2019progressive}  &  3.6  &   &  17.49$^\ddagger$  &  0.3 \\
				R-DARTS \cite{zela2020understanding}  & - &  - &  18.01$\pm$0.26 & 1.6 \\
				NoisyDARTS & 4.7 & 667 & \textbf{16.28}   & 0.4 \\
				\hline
			\end{tabular}
			\caption{Comparison of searched models on CIFAR-100. $^\diamond$: Reported by \cite{dong2019searching}, $^\star$: Reported by \cite{zela2020understanding}, $^\ddagger$:Rerun their code. }
	\label{tab:comparison-cifar100}
\end{table}

\subsection{Training Settings on COCO Object Detection}\label{sec:coco}

We use the MMDetection tool box since it provides a good implementation for various detection algorithms \cite{chen2019mmdetection}. Following the same training setting as \cite{lin2017focal}, all models in Table~\ref{table:fairnas-coco-retina} are trained and evaluated on the COCO dataset for 12 epochs. The learning rate is initialized as 0.01 and decayed by 0.1$\times$ at epoch 8 and 11.

\subsection{Training Settings on ImageNet}\label{sec:imagenet_training}
We split the original training set into two datasets with equal capacity to act as our training and validation dataset. The original validation set is treated as the test set. We use the SGD optimizer with a batch size of 768. The learning rate for the network weights is initialized as 0.045 and it decays to 0 within 30 epochs following the cosine decay strategy.  Besides, we utilize Adam optimizer ($\beta_1=0.5, \beta_2=0.999$) and a constant learning rate of 0.001. 
\subsection{Detailed Settings on NAS-Bench-201}\label{app:bench}
For NAS-Bench-201 experiments, we adapt the code from \cite{dong2020nasbench}. We only use the first-order DARTS optimization. We track the running statistics for batch normalization to be the same as DARTS \cite{liu2018darts}. Each setting is run 3 times to obtain the average. We use a noise of $\sigma=0.8$ regarding this particular search space.

\subsection{Relations to Other Work}\label{sec:comp-reg}


\textbf{Comparison with PNI.}  PNI \cite{He_2019_CVPR} uses parametric noise  to boost adversarial training.  In contrast, we  inject the fixed noise at the output of candidate operations and  smooth the loss landscape of the bi-level search to avoid collapse. Moreover, parametric noise leads to collapse on NAS-Bench-201 (see learnable $\sigma$ in Table~\ref{tab:bench201}).

\subsection{Transferred Results}
\paragraph{Transferred results on object detection.}
We further evaluate the transferability of our searched models on the COCO objection task \cite{lin2014coco}. Particularly, we utilize a drop-in replacement for the backbone based on  Retina \cite{lin2017focal}.  As shown in Table~\ref{table:fairnas-coco-retina} (supplementary), our model obtains the best transferability than other models under the mobile settings. Detailed setting is provided in Section~\ref{sec:coco} (supplementary).

\paragraph{Transferring ImageNet models to CIFAR-10.}We transferred our model NoisyDARTS-A  searched on ImageNet to CIFAR-10.  Specifically, the transferred model NoisyDARTS-A-t achieved $98.28\%$ top-1 accuracy with only 447M FLOPS, as shown in Table~\ref{tab:comparison-cifar10-imagenet}. Training details are listed in Section~\ref{sec:cifar10-training} (Supplementary).

\subsection{Evolution of NoisyDARTS architectural parameters}
We plot the evolution of architectural parameters during the NoisyDARTS optimization in Figure~\ref{fig:alpha-evolution-best}. The injected noise is zero-mean Gaussian with $\sigma=0.2$. As normal cells are the main building blocks (18 out of 20) of the network, we see that the number of skip connections is much reduced. Compared with \cite{liang2019darts} and \cite{chen2019progressive}, we don't set any hard limits for it. We also don't compute expensive Hessian eigenspectrum \cite{zela2020understanding} as a type of regularization for skip connections. Neither do we use Scheduled DropPath \cite{zoph2017learning} or fixed drop-path during the searching. It confirms that by simply disturbing the gradient flow of skip connections, the unfair advantage is much weakened so that the optimization is fairer to deliver better performing models.

\begin{figure*}[ht]
	\centering
	\subfigure[Normal cell]{
		\includegraphics[width=0.85\textwidth]{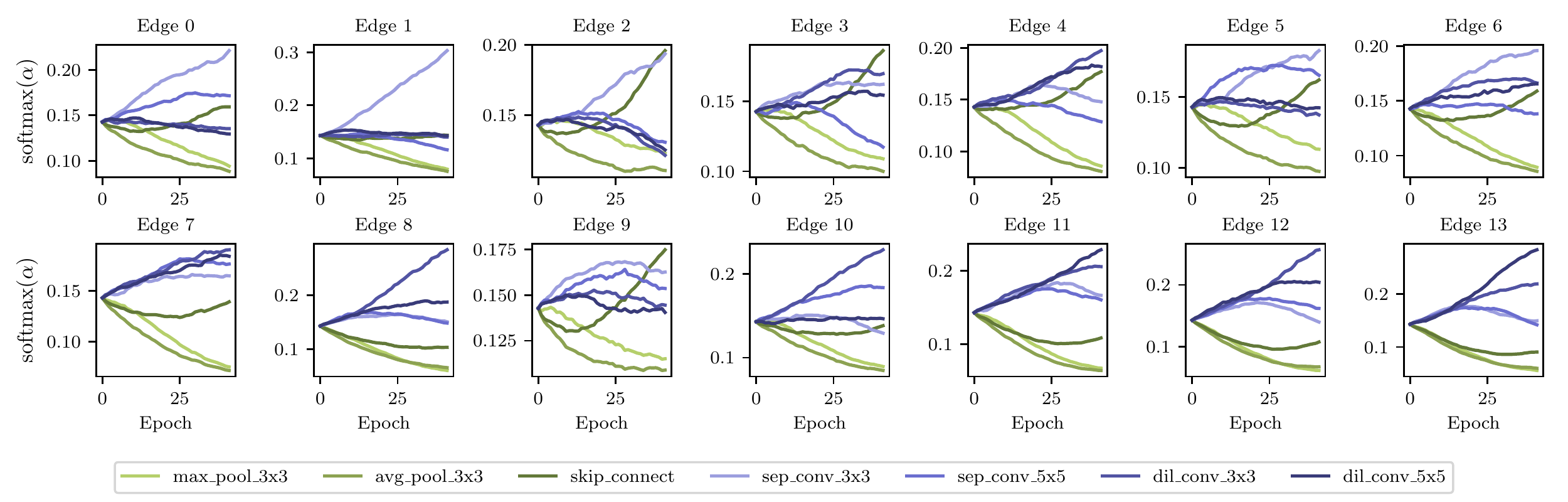}
	}
	\subfigure[Reduction cell]{
		\includegraphics[width=0.85\textwidth]{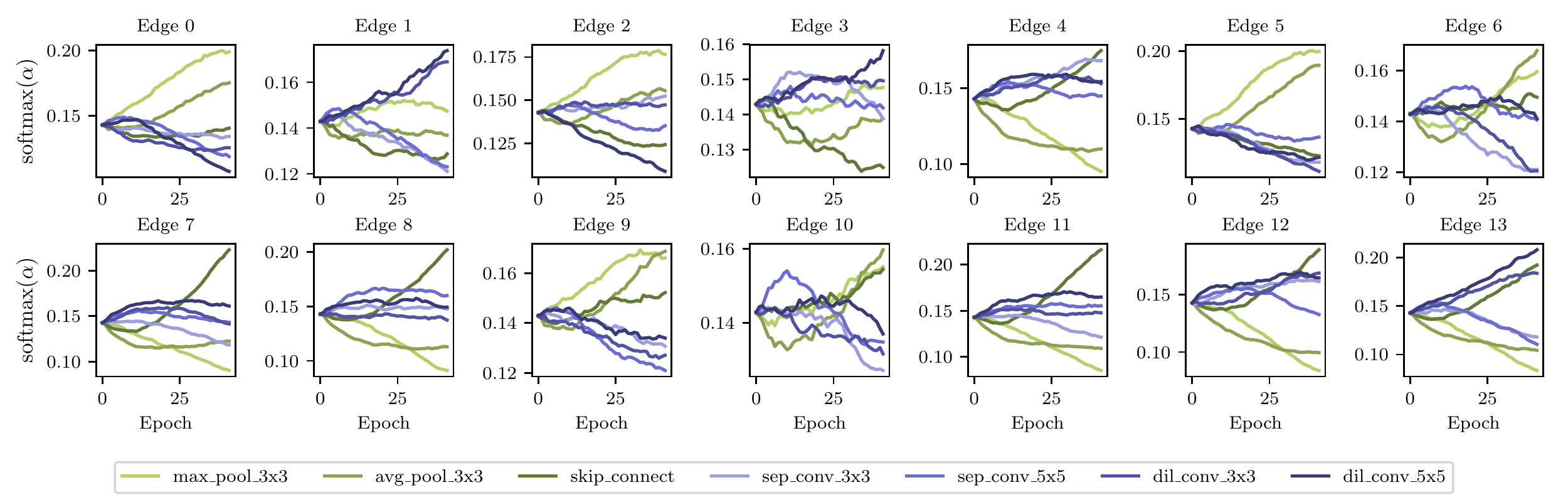}
	}
	\caption{Evolution of architectural weights during the NoisyDARTS searching phase on CIFAR-10. Skip connections in normal cells are largely suppressed.}
	\label{fig:alpha-evolution-best}
\end{figure*}


\begin{figure*}[h]
	\centering
	\includegraphics[width=\textwidth]{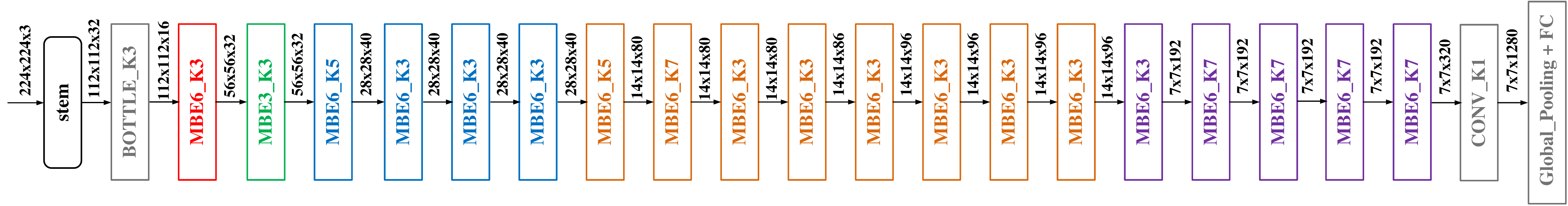}
	\caption{NoisyDARTS-A searched on ImageNet. Colors represent different stages.}
	\label{fig:imagenet-architecture}
\end{figure*}

\begin{figure}[ht]
	\centering		
	\includegraphics[width=\textwidth]{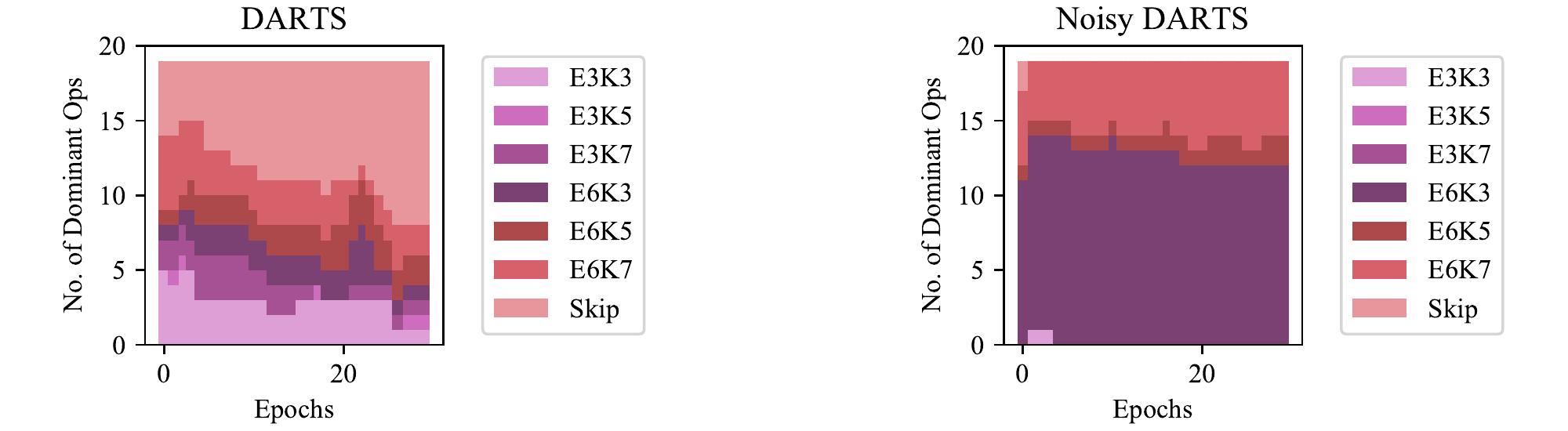}
	\caption{Stacked plot of dominant operations during searching on ImageNet. The inferred model of DARTS (left) obtains 66.4\% accuracy on ImageNet, while NoisyDARTS (right) obtains 76.1\%. }
	\label{fig:stacked-plot-dominant-imagent}
\end{figure}

\subsection{More discussions about Hessian Indicator in Reduced Search Space}\label{sec: more_dis_hessian}
Specifically, when training these models from supposed early-stop points, we only obtain a lower average performance 97.02$\pm$0.21. How about other search spaces? We further evaluate the Hessian eigenvalue trajectories of our method in the reduced search space, which are shown in Figure~\ref{fig:noisy-reduces-ss-eigen}.

\begin{figure}[ht]
	\centering
	\subfigure{
		\includegraphics[width=0.45\textwidth]{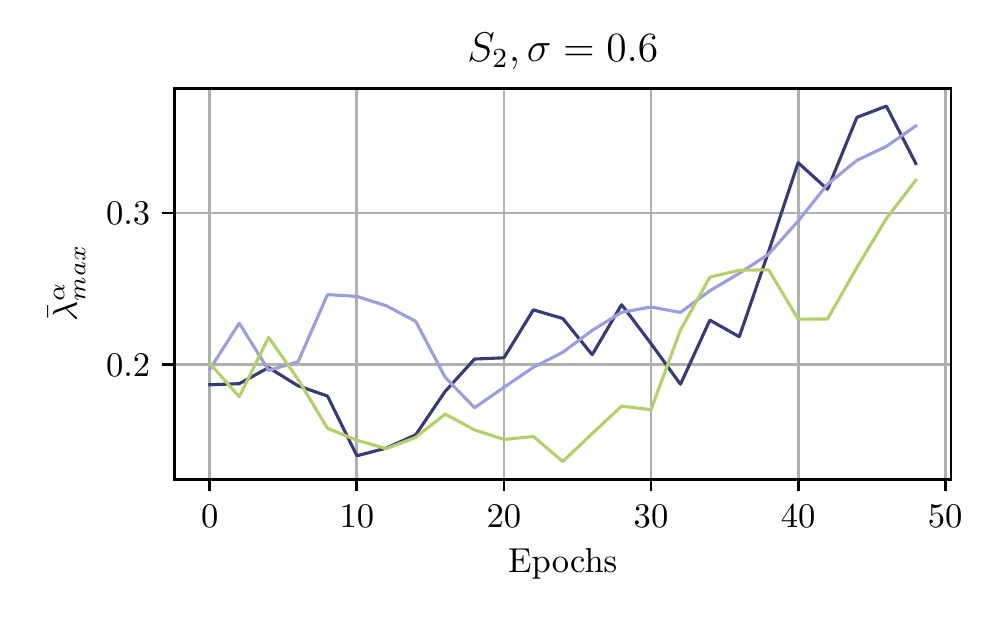}
	}
	\subfigure{
		\includegraphics[width=0.45\textwidth]{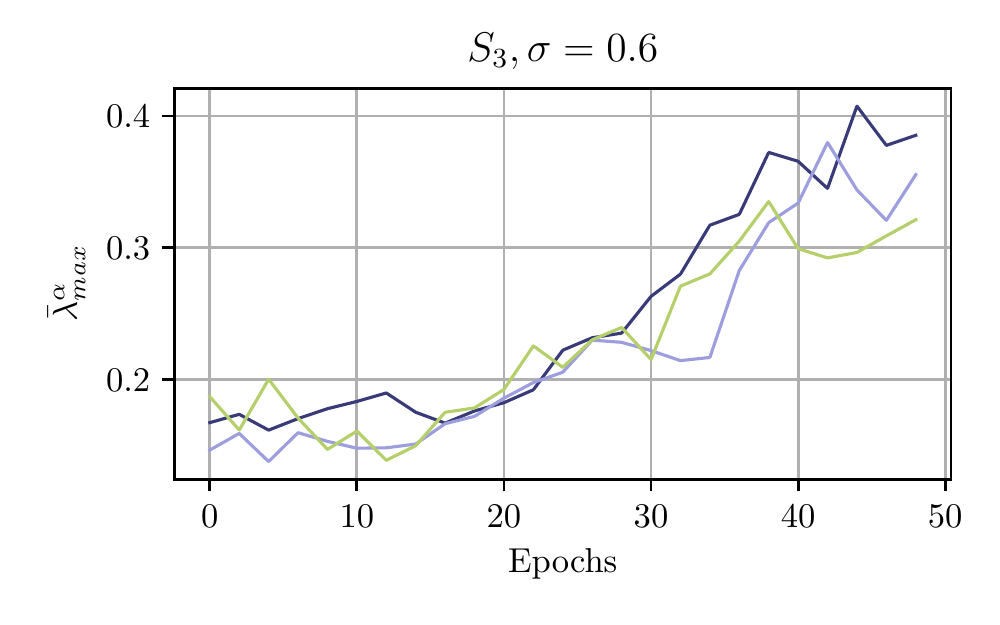}
	}
	\caption{Evolution of maximal Hessian eigenvalue when searching with NoisyDARTS on two reduced search spaces $S_2$ and $S_3$ proposed by \cite{zela2020understanding}. Compared with RDARTS, the eigenvalues still have a trend of increasing. Notice that better models can be found $3\times$ faster than RDARTS (they run four times to get the best model while we produce better ones at each single run).}
	\label{fig:noisy-reduces-ss-eigen}
\end{figure}

When we search with injected Gaussian noise, we still observe an obvious growth of eigenvalues in both of two spaces. However, when being trained from scratch, the models derived from the last epoch (without early-stopping or any regularization tricks) perform much better than their proposed adaptive eigenvalue regularization method DARTS-ADA \cite{zela2020understanding}. Compared with their best effort L2 regularization \cite{zela2020understanding} and another method SDARTS \cite{chen2020stabilizing} based on implicit regularization of Hessian norm, we also have better performance in $S_2$ and comparable performance in $S_3$ (see Table~\ref{table:rdarts-reduced-ss} in the main text). Notice we only use \textbf{3x fewer} searching cost. This reassures our observation that the Hessian norm \cite{zela2020understanding} may not be an ideal indicator of performance collapse, because it rejects good models by mistake, as illustrated in Figure~\ref{fig:hessian-set}.

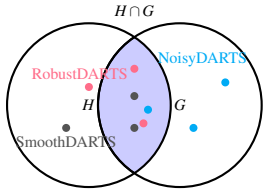
\begin{figure}[!ht]
	\centering
	\begin{tikzpicture}[auto,node distance=0.8cm,
	semithick,scale=0.6, every node/.style={scale=0.6}]
	\tikzset{filled/.style={fill=circle area, draw=circle edge, thick},
		outline/.style={draw=circle edge, thick}}
	\begin{scope}
	\clip \firstcircle;
	\fill[filled] \secondcircle;
	\end{scope}
	\draw[outline] \firstcircle node {$H$};
	\draw[outline] \secondcircle node {$G$};
	\filldraw [xapink] (1,.8) circle (2pt);
	\filldraw [xapink] (0,0.4) circle (2pt);
	\filldraw [xapink] (1.2,-0.4) circle (2pt);
	\filldraw[xagray] (1,0.2) circle (2pt);
	\filldraw[xagray] (-0.5,-0.5) circle (2pt);
	\filldraw[xagray] (1,-0.5) circle (2pt);
	\filldraw[xablue] (2.3,-0.5) circle (2pt);
	\filldraw[xablue] (1.3,-0.1) circle (2pt);
	\filldraw[xablue] (3,0.5) circle (2pt);
	\node[xablue] at (2.5,1) {NoisyDARTS};
	\node[xapink] at (-0.2,0.7) {RobustDARTS};
	\node[xagray] at (-0.5,-0.8) {SmoothDARTS};
	\node[anchor=south] at (current bounding box.north) {$H \cap G$};
	\end{tikzpicture}
	\vskip -0.1in
	\caption{Exemplary illustration on the relation of the set of models. Set $H$ means models found with low Hessian norms. Set $G$ are the models with better test accuracy. RobustDARTS's Hessian norm criterion \cite{zela2020understanding} tends to reject a part of good models, e.g. blue models found by NoisyDARTS that are not in $H \cap G$. }
	\label{fig:hessian-set}
\end{figure}



\subsection{More Details on Reduced RobustDARTS Experiments}\label{app:geno}

Like on CIFAR-10, we repeatedly find the Hessian eigenvalues are both growing when searching with NoisyDARTS on CIFAR-100 and SVHN datasets (see Figure~\ref{fig:noisy-reduces-ss-eigen-cifar100}), but models derived from these searching runs still outperform or are comparable to those from regularized methods like RDARTS \cite{zela2020understanding} and SDARTS \cite{chen2020stabilizing} (see Table~\ref{table:rdarts-reduced-ss} in the main text). These results again confirm that the eigenvalues are not necessarily a good indicator for finding better-performing models.

\begin{table}[h]
	\begin{center}
		\begin{footnotesize}
			\begin{tabular}{*{5}{|c}|*{3}{|c}|*{3}{|c}|}
				\hline
				\multirow{2}{2.5em}{Space} & \multirow{2}{1em}{$\sigma$} & \multicolumn{3}{c||}{Test acc. (\%)} & \multicolumn{3}{c||}{Params (M)}  & \multicolumn{3}{c|}{$\lambda_{max}^{\alpha}$}   \\
				\cline{3-11}
				& &  \multicolumn{3}{c||}{seed} &  \multicolumn{3}{c||}{seed} &  \multicolumn{3}{c|}{seed}  \\
				\cline{3-11}
				 & &  1 &  2 &  3 &  1 &  2 &  3 &  1 &  2 &  3  \\
				\hline
				\multirow{3}{1em}{$S_2$} & 0.6 & 97.43  & 97.32  & \textbf{97.46}  & 3.59  & 3.26  & 3.26 &0.260 &0.349 &0.309 \\
				& 0.8   & 97.30  & 97.39  & 97.37  & 3.62  & 3.62  & 3.62 & 0.133 &0.270& 0.429 \\
				& 1.0   & 97.35  & 97.34  & 97.25 &4.34  & 3.98  & 3.98 & 0.119 &0.171 &0.295\\
				\hline
				\multirow{3}{1em}{$S_3$} & 0.6   & 97.32  & \textbf{97.47}  & 97.28  & 3.62  & 3.98  & 3.62 & 0.345 &0.418 &0.290 \\
				& 0.8  & 97.32  & 97.24  & 97.27 & 3.98 & 3.62 & 3.26 & 0.393&0.327 &0.336 \\
				& 1.0   & 97.27  & 97.41  & 97.34 & 4.34 & 3.98 & 3.98 & 0.289 &0.341 &0.379\\
				\hline
			\end{tabular}
		\end{footnotesize}
	\end{center}
	\caption{Test accuracy and the maximum Hessian eigenvalue (in the final searching epoch) of NoisyDARTS models searched with different $\sigma$ in the reduced search spaces of RobustDARTS on CIFAR-10. Notice here we train models in $S_2$ with the same settings as in $S_3$. It's interesting to see that $\lambda_{max}^{\alpha}$ = 0.418 in $S_3$ is the best model with $97.47\%$ top-1 accuracy. However, such similar value in \cite{zela2020understanding}  indicates a failure ($94.70\%$) under the same setting}\smallskip
	\label{table:rdarts-reduced-ss-more-sigma}
\end{table}

\begin{figure*}[ht]
	\centering
	\subfigure{
		\includegraphics[width=0.45\textwidth]{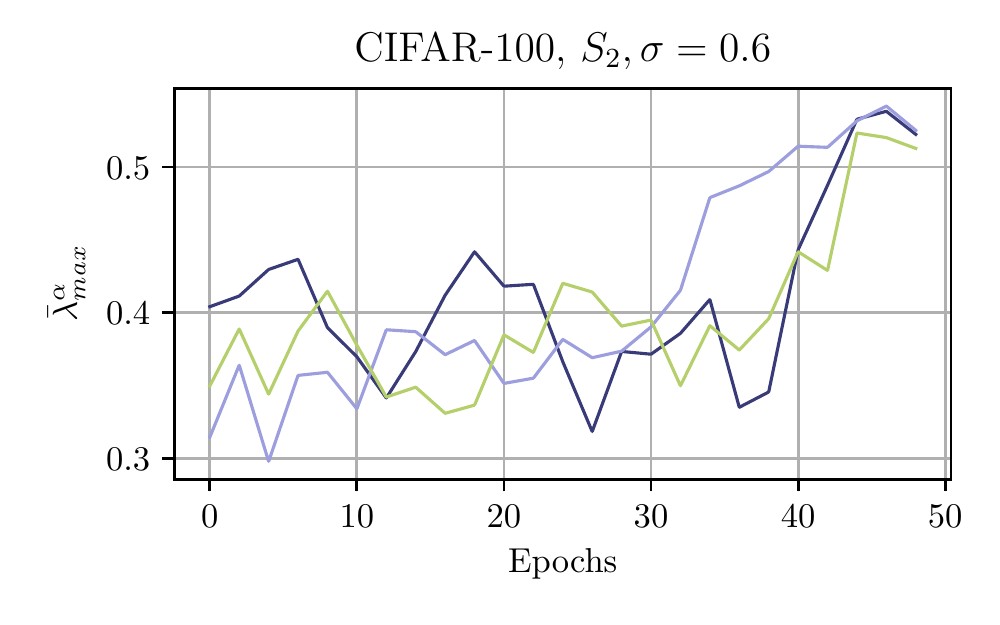}
	}
	\subfigure{
		\includegraphics[width=0.45\textwidth]{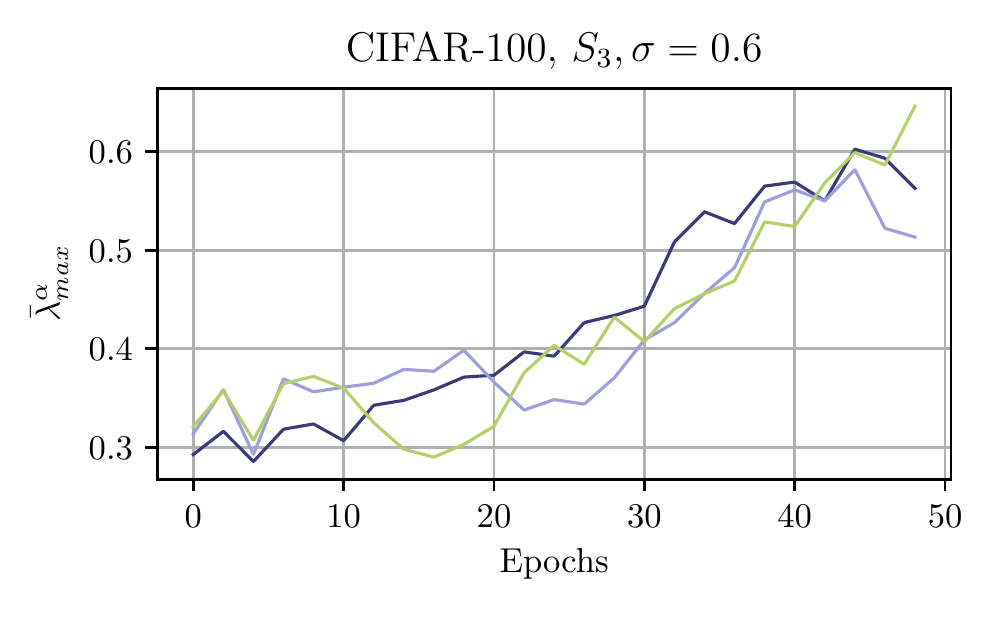}
	}
	\subfigure{
		\includegraphics[width=0.45\textwidth]{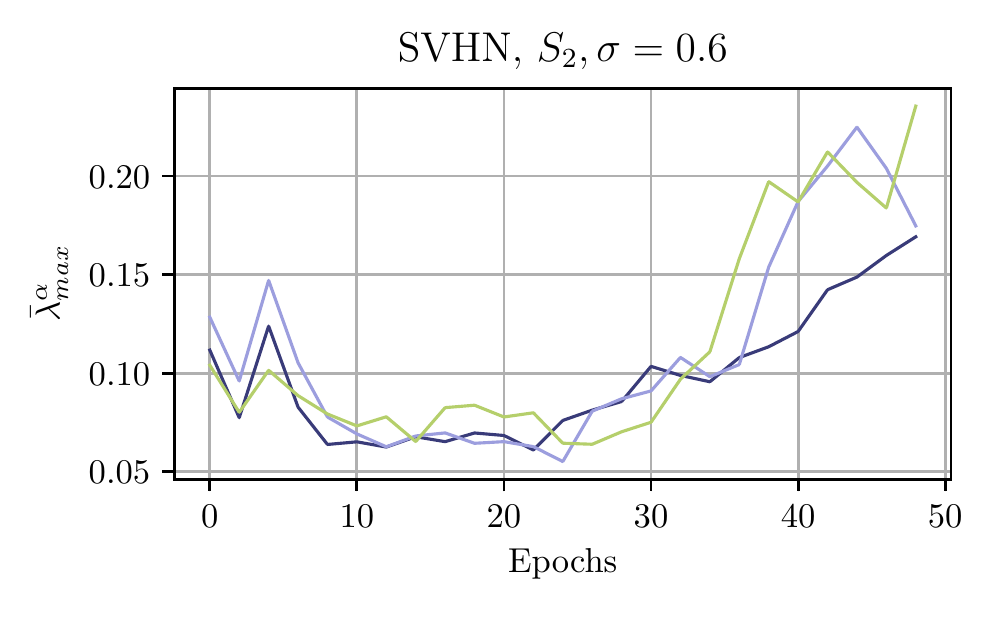}
	}
	\subfigure{
		\includegraphics[width=0.45\textwidth]{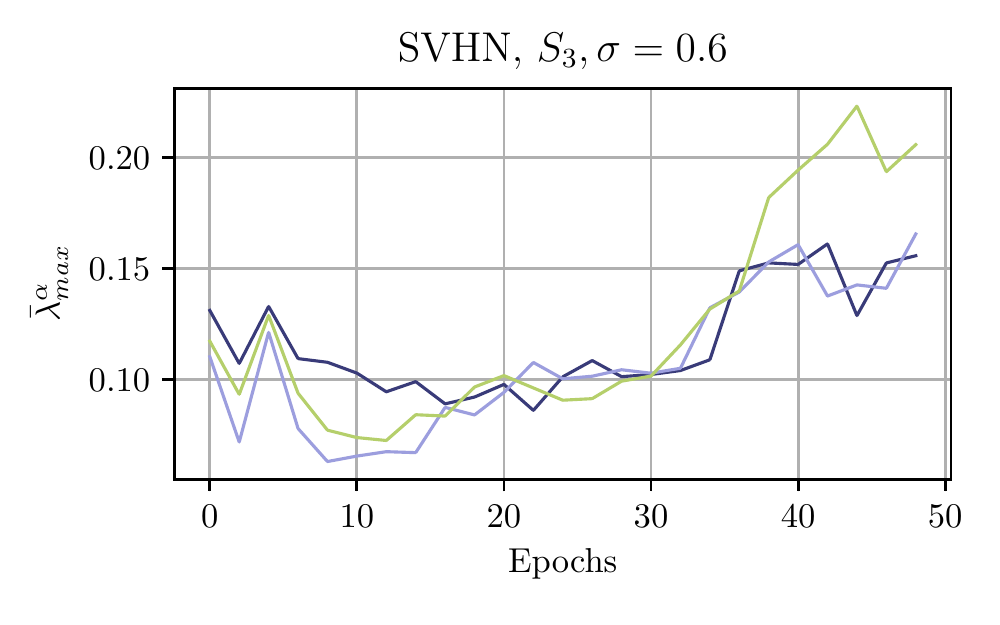}
	}
	\caption{Evolution of maximal Hessian eigenvalue when searching with NoisyDARTS on CIFAR-100 and SVHN, in two reduced search spaces $S_2$ and $S_3$ proposed by \cite{zela2020understanding}.}
	\label{fig:noisy-reduces-ss-eigen-cifar100}
\end{figure*}

\section{NoisyDARTS architectures}\label{app:fig-archs}

\subsection{Models searched on CIFAR-10 in the DARTS search space}

We plot all the best models in different configurations of searching from Figure~\ref{fig:the-searched-cells-a} to Figure~\ref{fig:the-searched-cells-j}.

\begin{figure}[t]
	\centering
	\subfigure[Normal cell]{
		\includegraphics[width=0.48\textwidth,scale=0.8]{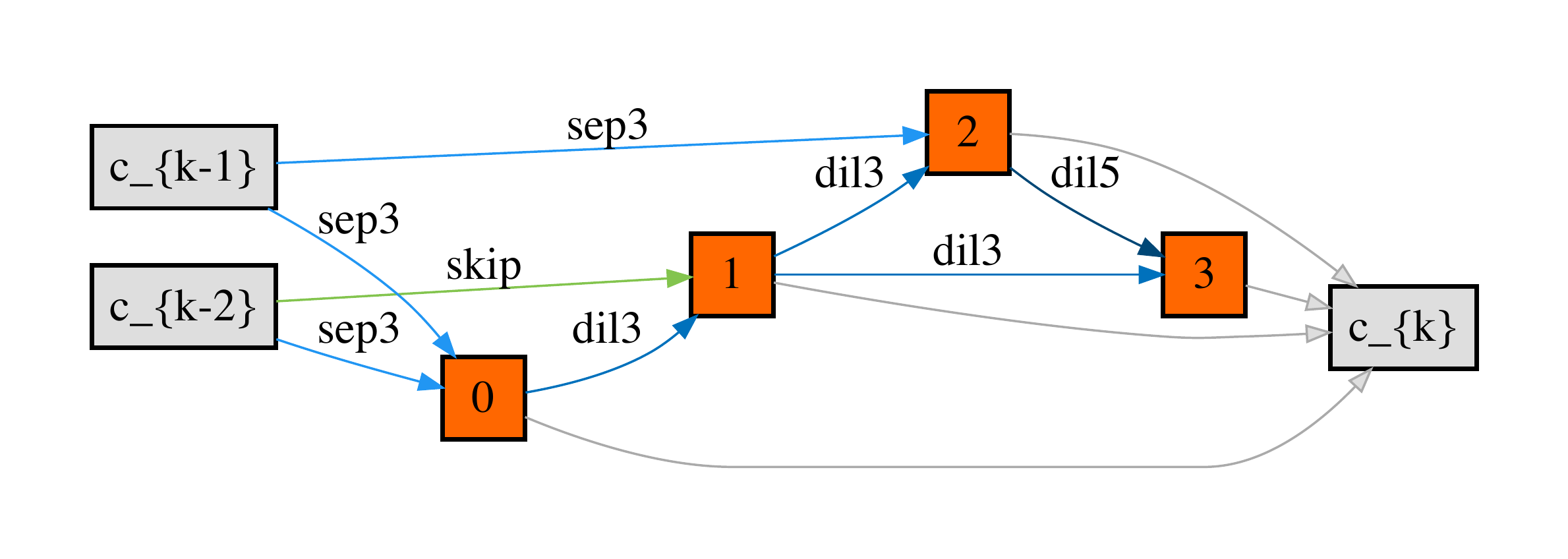}
	}
	\subfigure[Reduction cell]{
		\includegraphics[width=0.48\textwidth,scale=0.8]{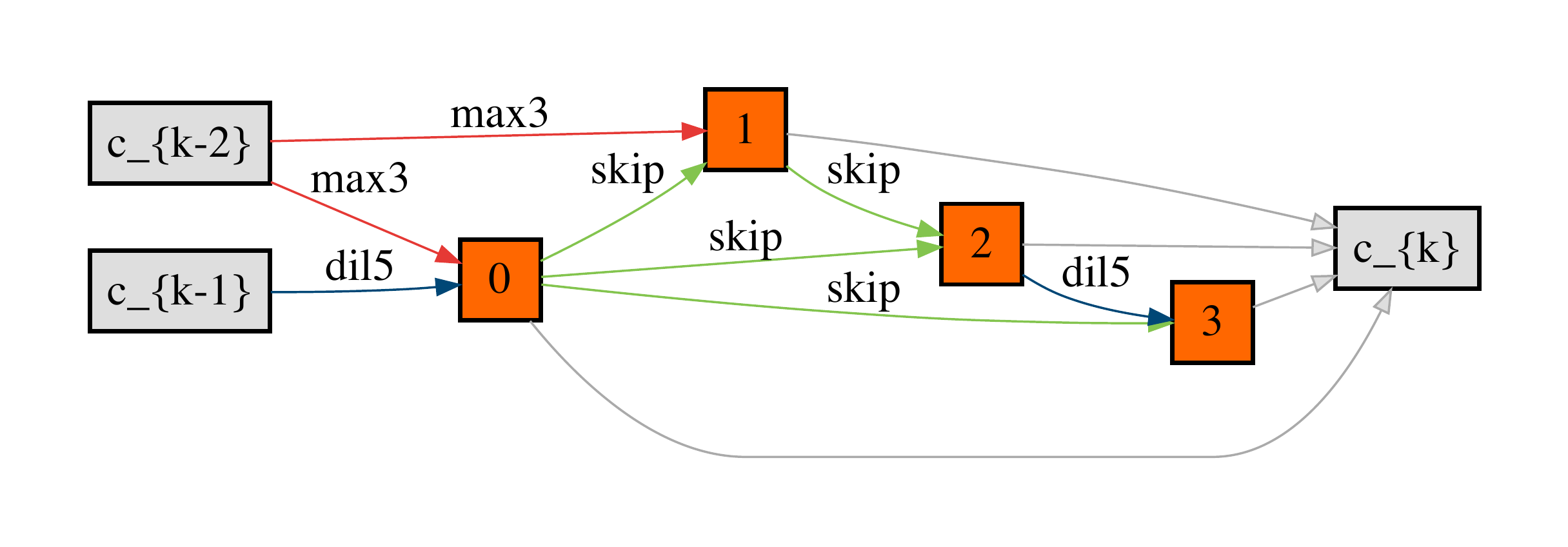}
	}
	\caption{NoisyDARTS-a cells searched on CIFAR-10.}
	\label{fig:the-searched-cells-a}
\end{figure} 



\begin{figure}[ht]
	\centering
	\subfigure[Normal cell]{
		\includegraphics[width=0.45\textwidth,scale=0.8]{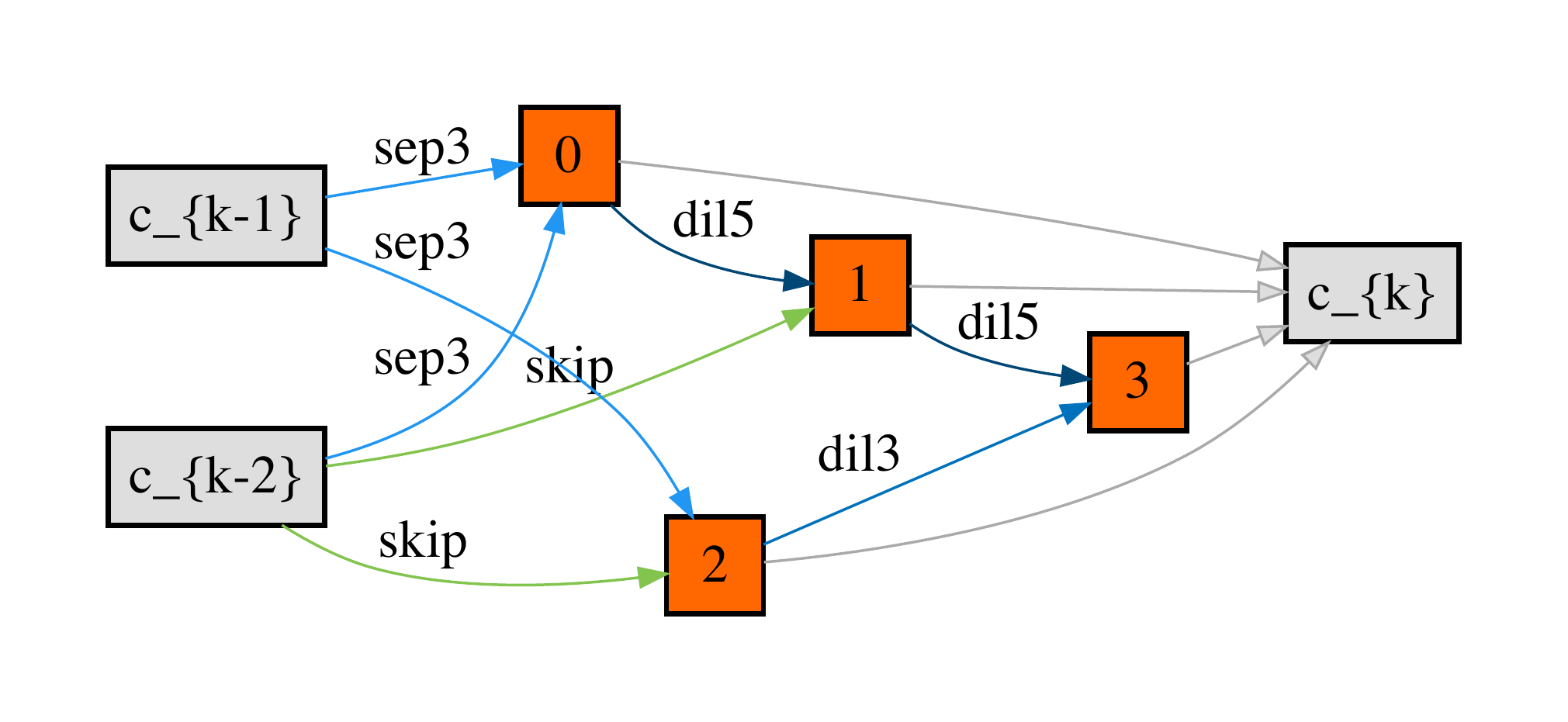}
	}
	\subfigure[Reduction cell]{
		\includegraphics[width=0.45\textwidth,scale=0.8]{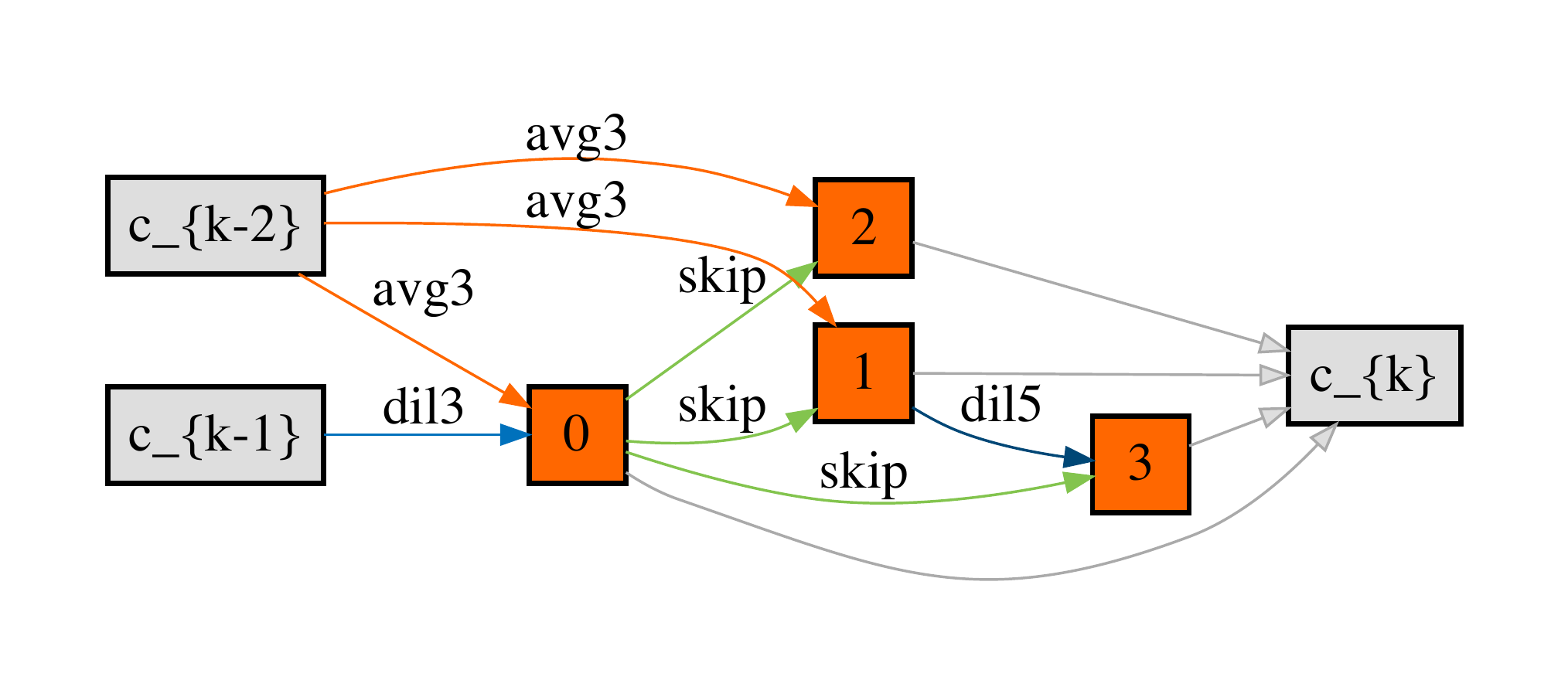}
	}
	\caption{NoisyDARTS-b cells searched on CIFAR-10 with additive Gaussian noise, $\mu=0$, $\sigma=0.1$.}
	\label{fig:the-searched-cells-b}
\end{figure}

\begin{figure}[ht]
	\centering
	\subfigure[Normal cell]{
		\includegraphics[width=0.45\textwidth,scale=0.8]{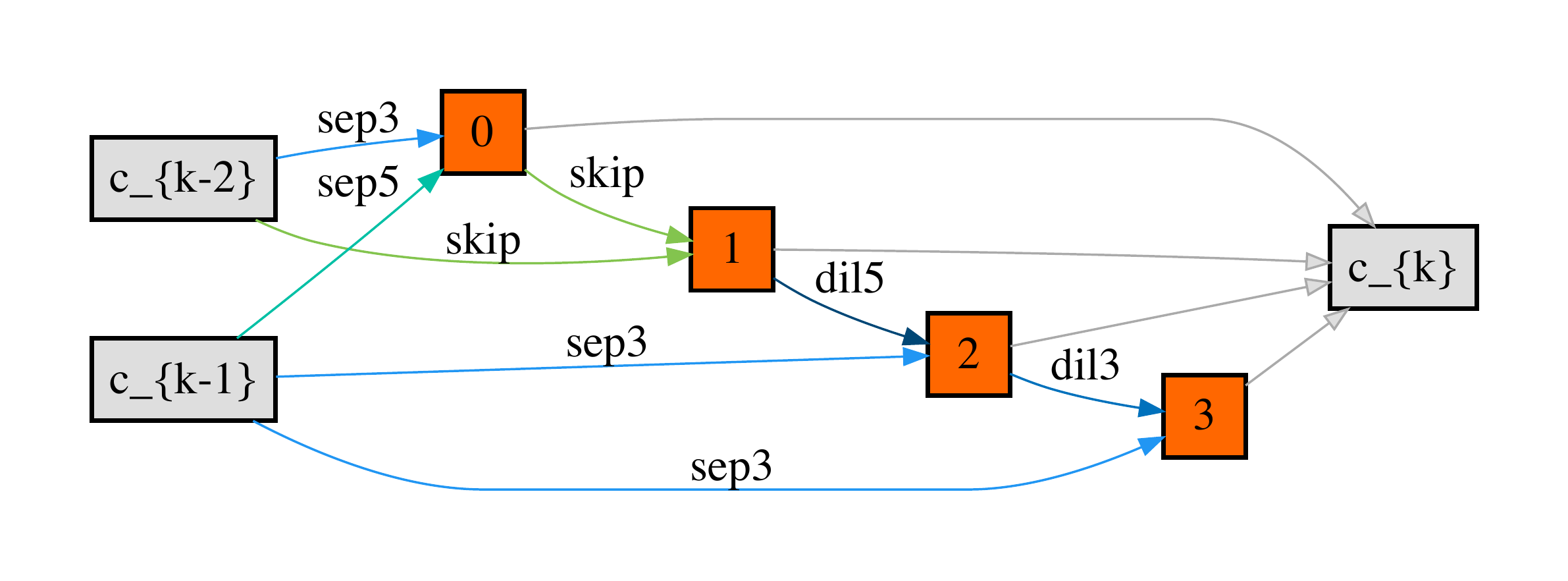}
	}
	\subfigure[Reduction cell]{
		\includegraphics[width=0.45\textwidth,scale=0.8]{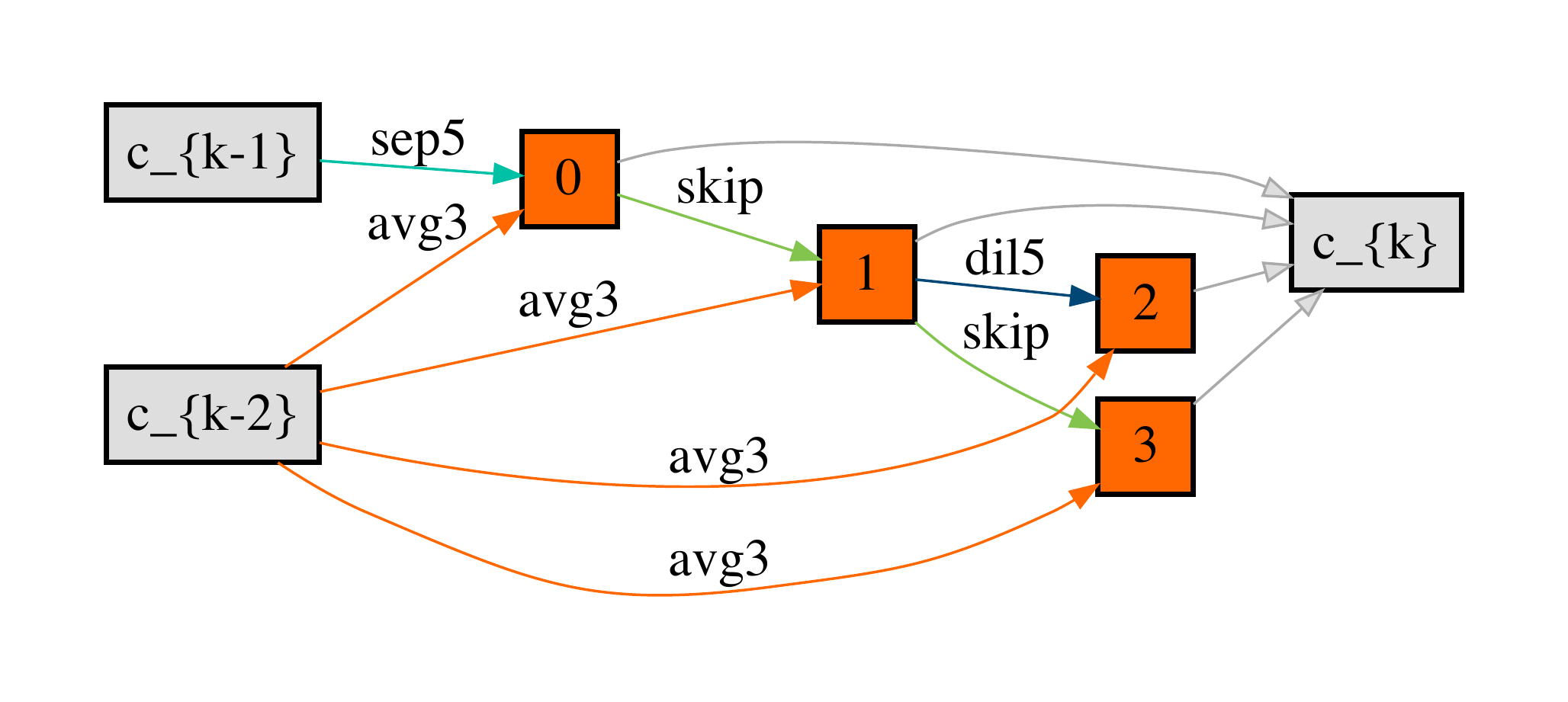}
	}
	\caption{NoisyDARTS-c cells searched on CIFAR-10 with additive uniform noise, $\mu=0$, $\sigma=0.2$.}
	\label{fig:the-searched-cells-c}
\end{figure}

\begin{figure}[ht]
	\centering
	\subfigure[Normal cell]{
		\includegraphics[width=0.45\textwidth,scale=0.8]{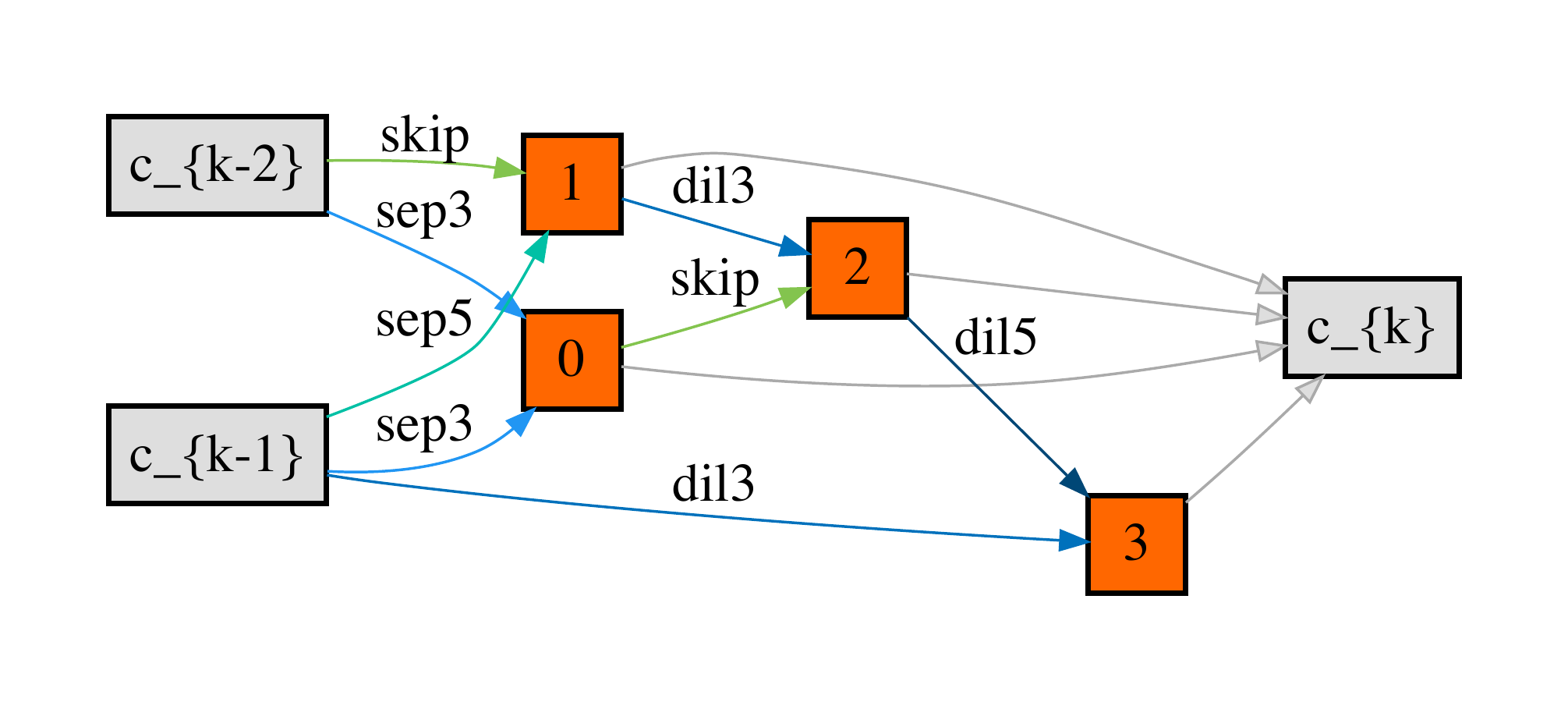}
	}
	\subfigure[Reduction cell]{
		\includegraphics[width=0.45\textwidth,scale=0.8]{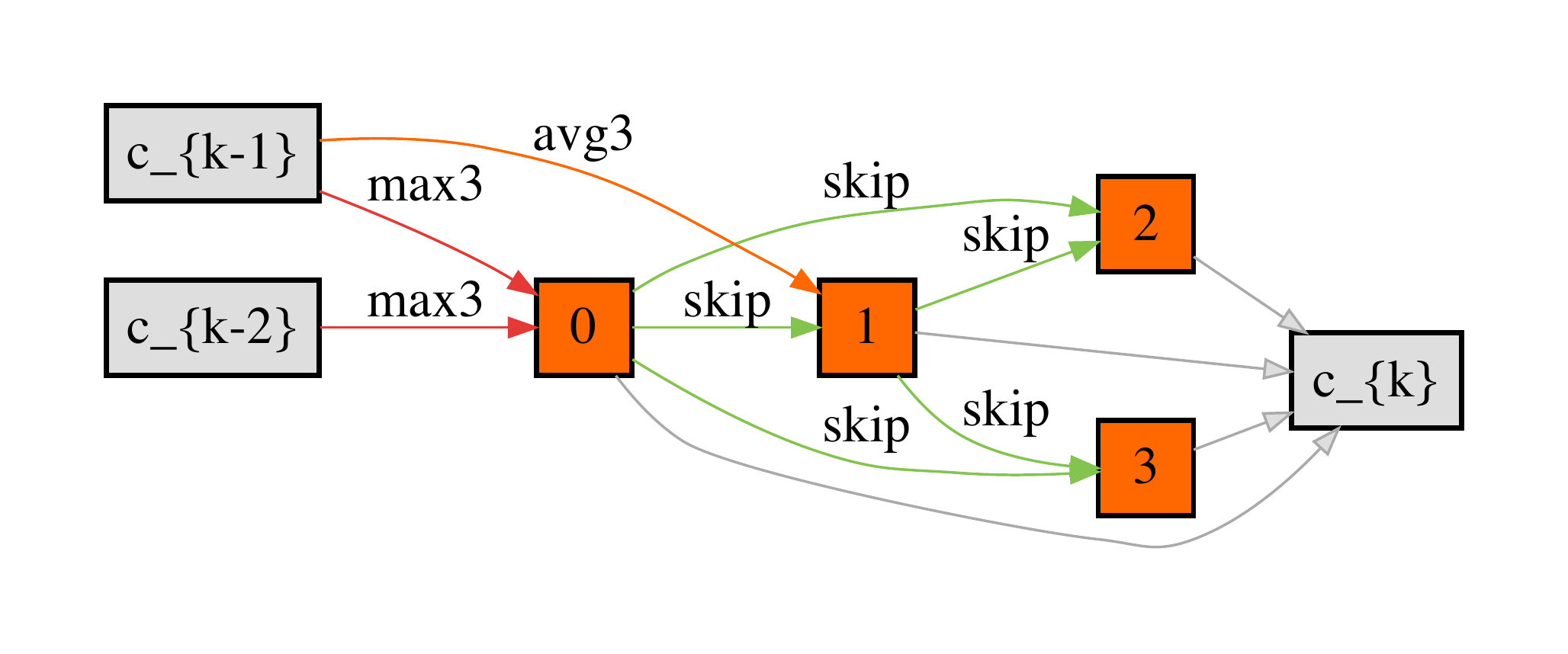}
	}
	\caption{NoisyDARTS-d cells searched on CIFAR-10 with additive uniform noise, $\mu=0$, $\sigma=0.1$.}
	\label{fig:the-searched-cells-d}
\end{figure}

\begin{figure}[ht]
	\centering
	\subfigure[Normal cell]{
		\includegraphics[width=0.45\textwidth,scale=0.8]{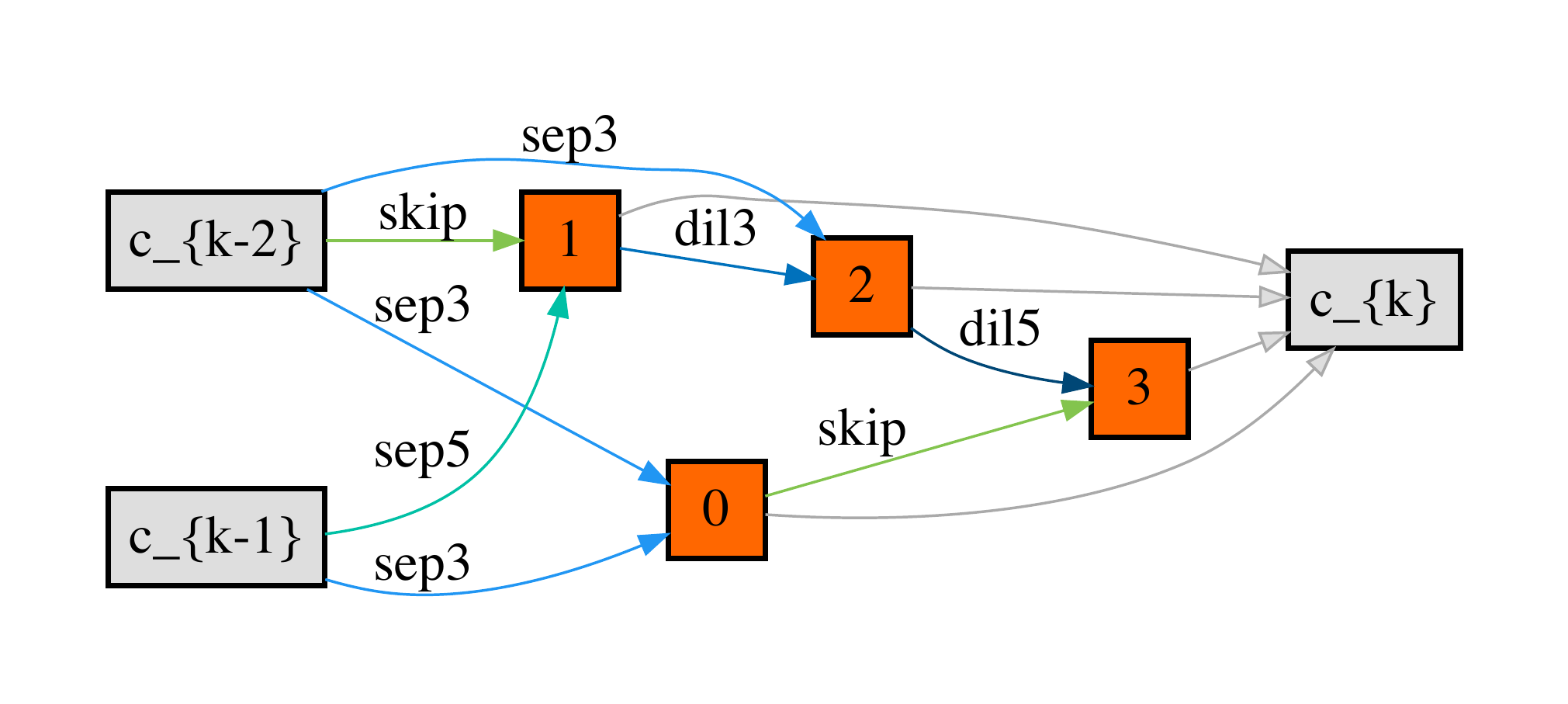}
	}
	\subfigure[Reduction cell]{
		\includegraphics[width=0.45\textwidth,scale=0.8]{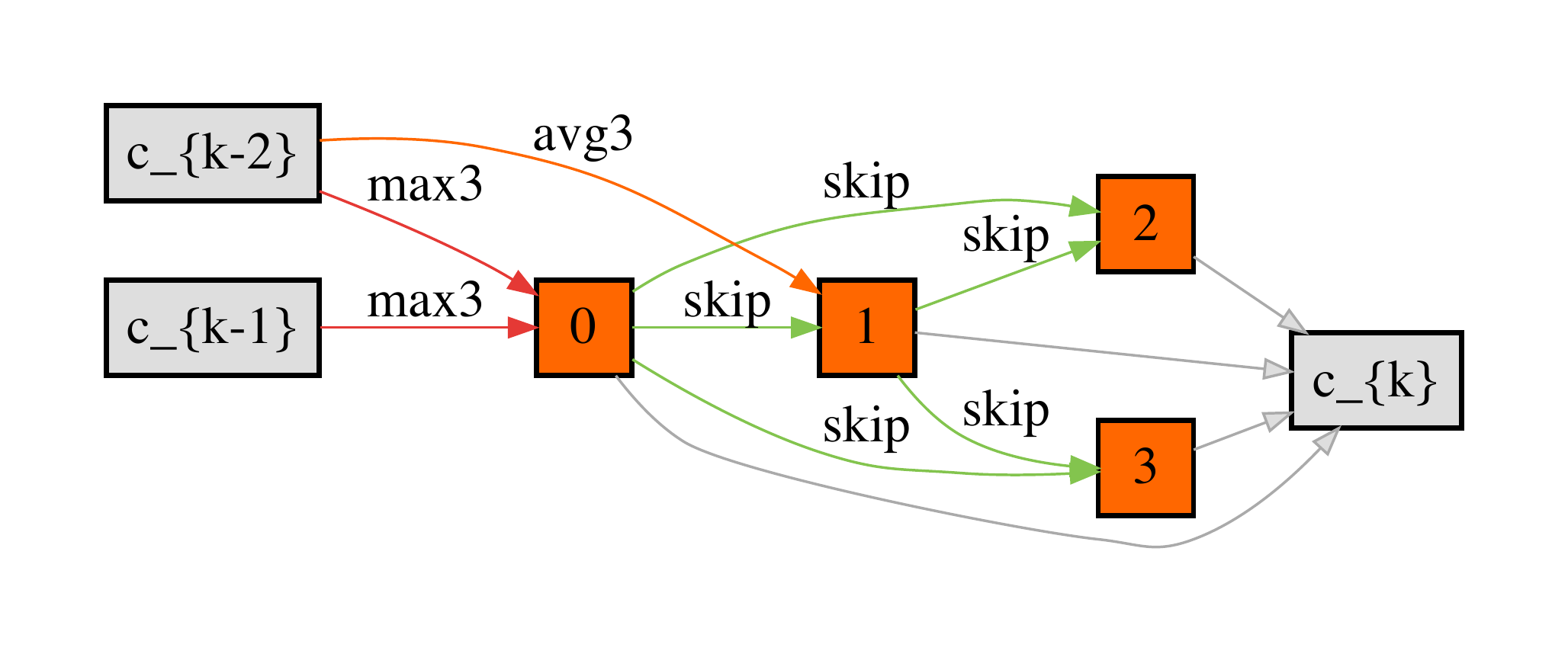}
	}
	\caption{NoisyDARTS-e cells searched on CIFAR-10 with multiplicative Gaussian noise, $\mu=0$, $\sigma=0.2$.}
	\label{fig:the-searched-cells-e}
\end{figure}

\begin{figure}[ht]
	\centering
	\subfigure[Normal cell]{
		\includegraphics[width=0.45\textwidth,scale=0.8]{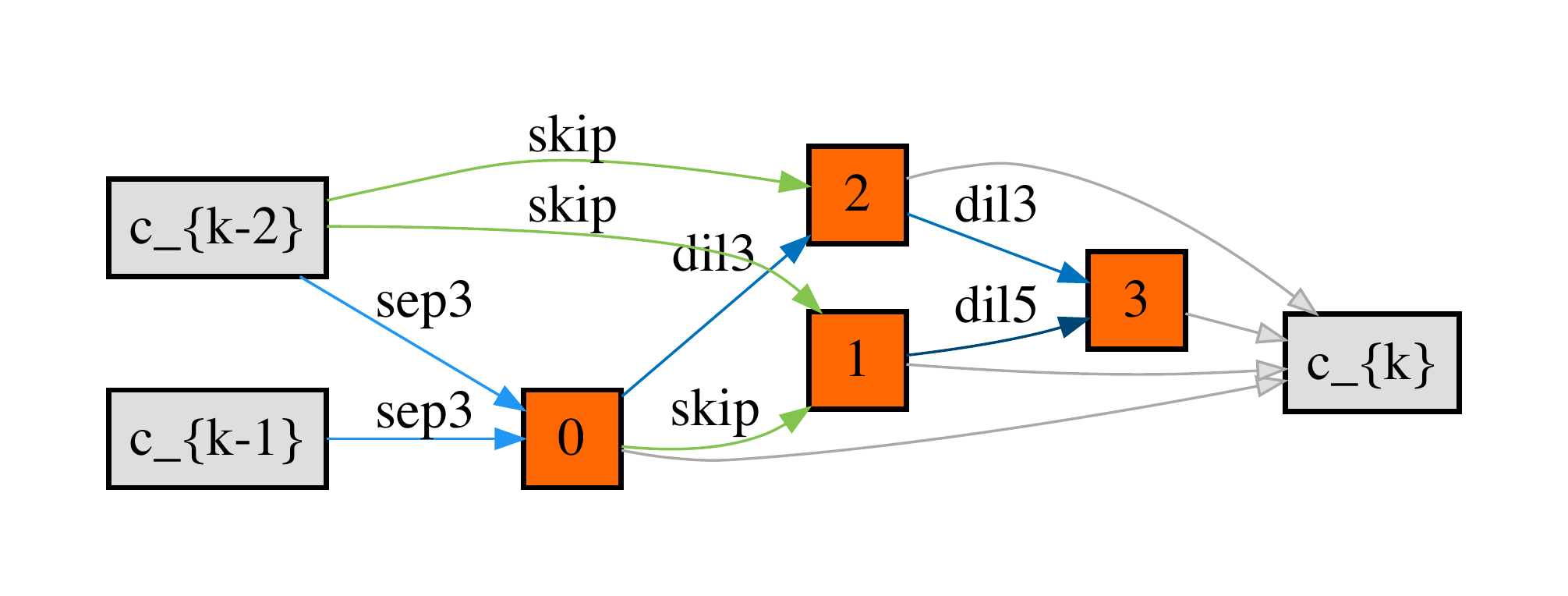}
	}
	\subfigure[Reduction cell]{
		\includegraphics[width=0.45\textwidth,scale=0.8]{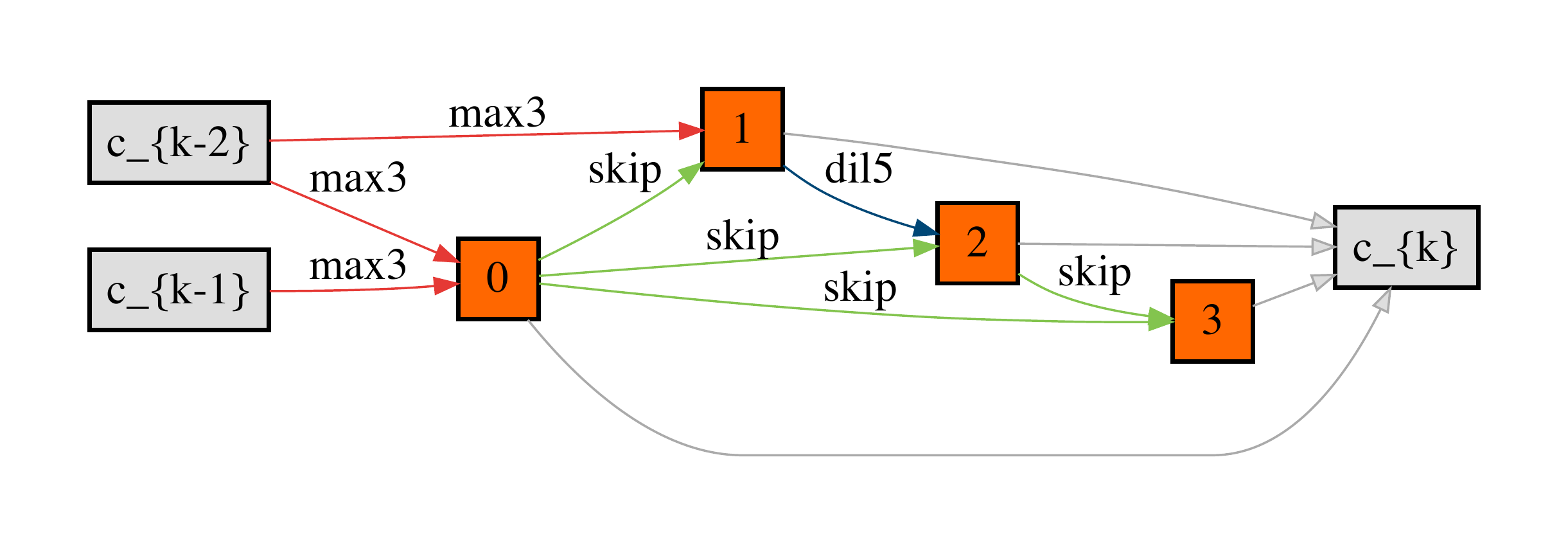}
	}
	\caption{NoisyDARTS-f cells searched on CIFAR-10 with multiplicative Gaussian noise, $\mu=0$, $\sigma=0.1$.}
	\label{fig:the-searched-cells-f}
\end{figure}

\begin{figure}[ht]
	\centering
	\subfigure[Normal cell]{
		\includegraphics[width=0.45\textwidth,scale=0.8]{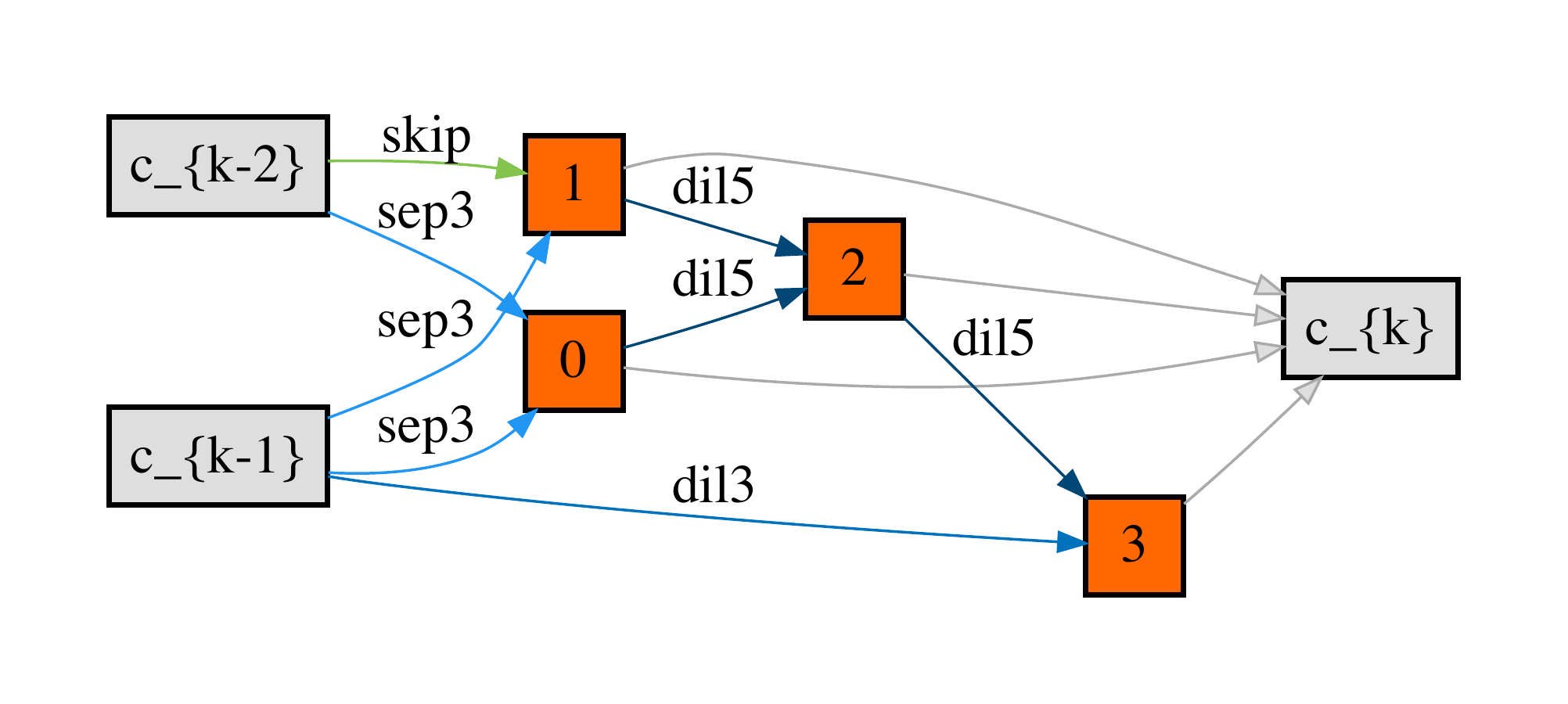}
	}
	\subfigure[Reduction cell]{
		\includegraphics[width=0.45\textwidth,scale=0.8]{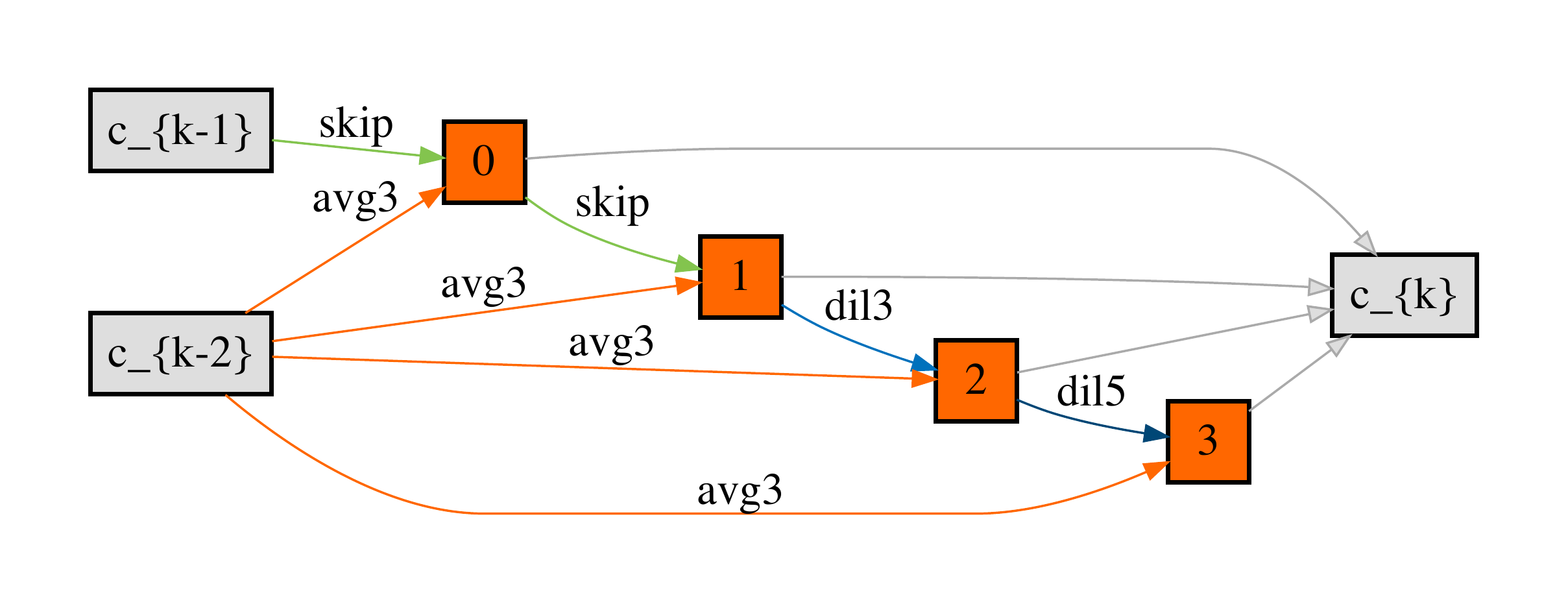}
	}
	\caption{NoisyDARTS-g cells searched on CIFAR-10 with additive Gaussian noise, $\mu=0.5$, $\sigma=0.2$.}
	\label{fig:the-searched-cells-g}
\end{figure}

\begin{figure}[ht]
	\centering
	\subfigure[Normal cell]{
		\includegraphics[width=0.45\textwidth,scale=0.8]{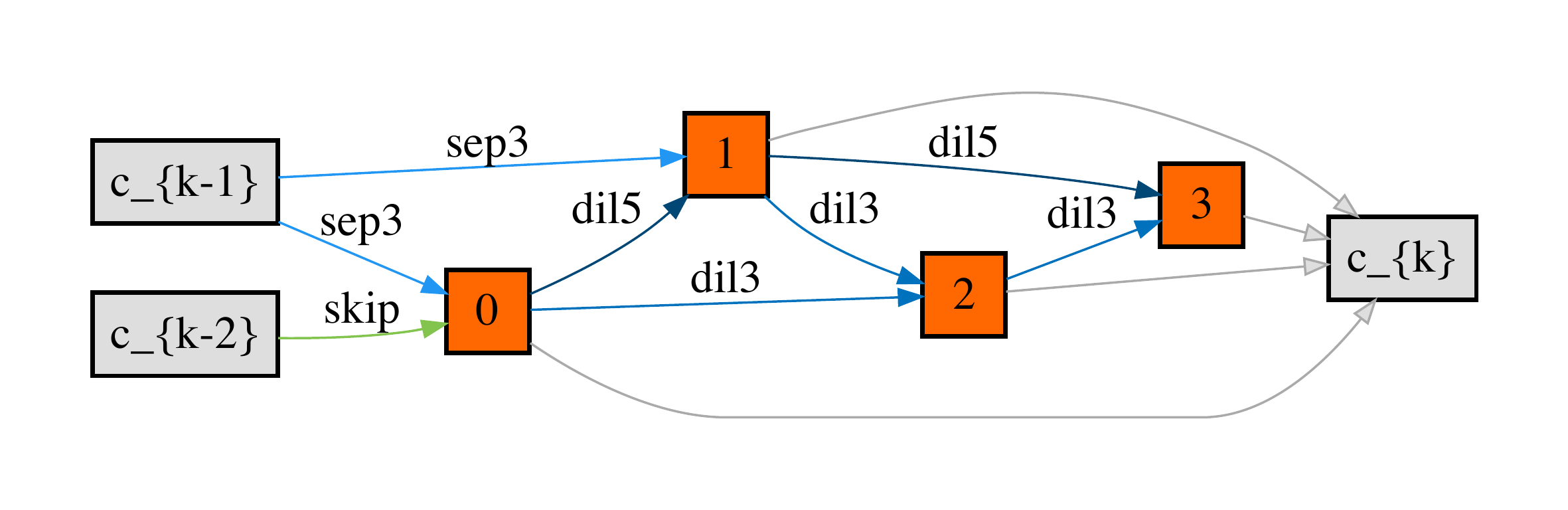}
	}
	\subfigure[Reduction cell]{
		\includegraphics[width=0.45\textwidth,scale=0.8]{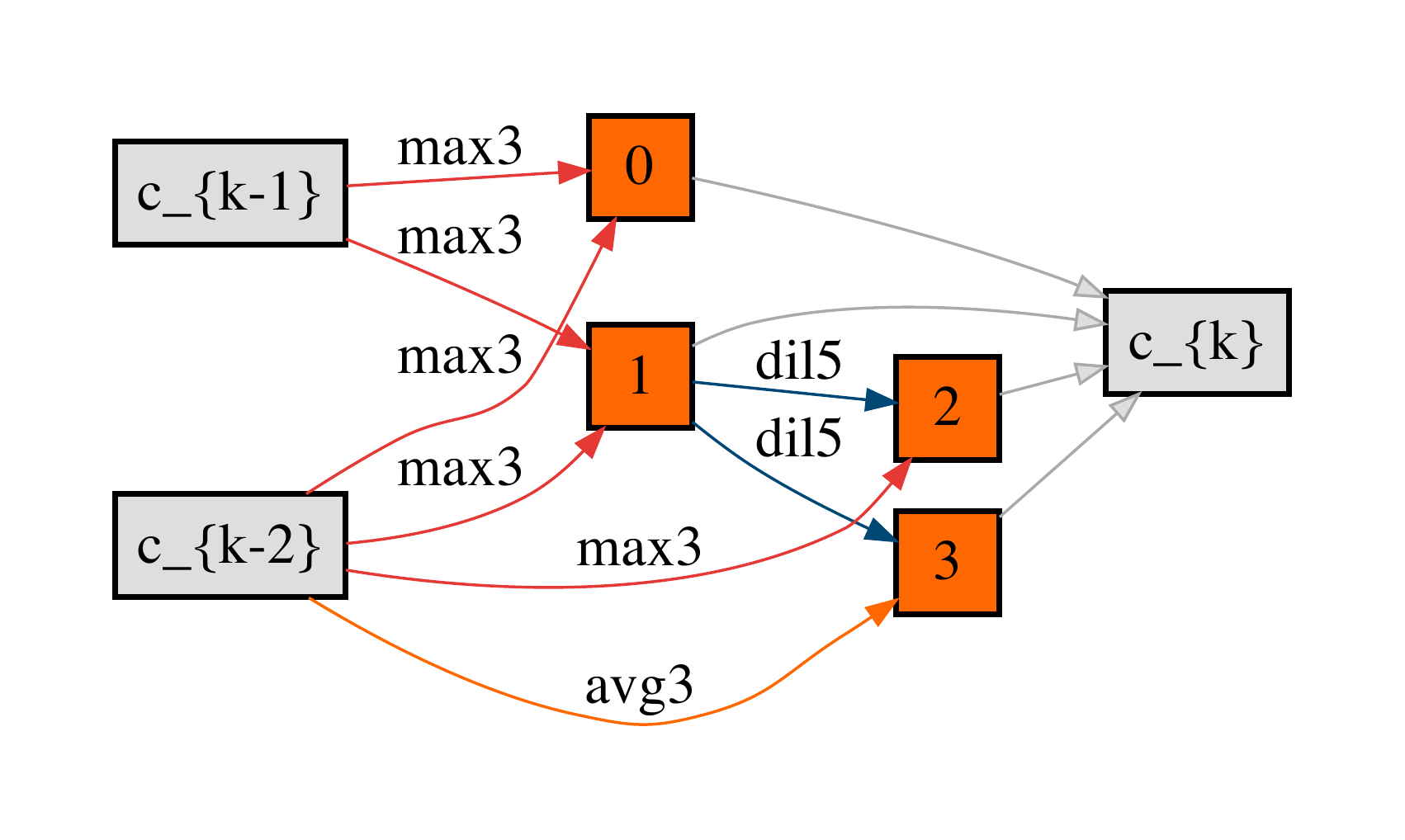}
	}
	\caption{NoisyDARTS-h cells searched on CIFAR-10 with additive Gaussian noise, $\mu=1.0$, $\sigma=0.2$.}
	\label{fig:the-searched-cells-h}
\end{figure}

\begin{figure}[ht]
	\centering
	\subfigure[Normal cell]{
		\includegraphics[width=0.45\textwidth,scale=0.8]{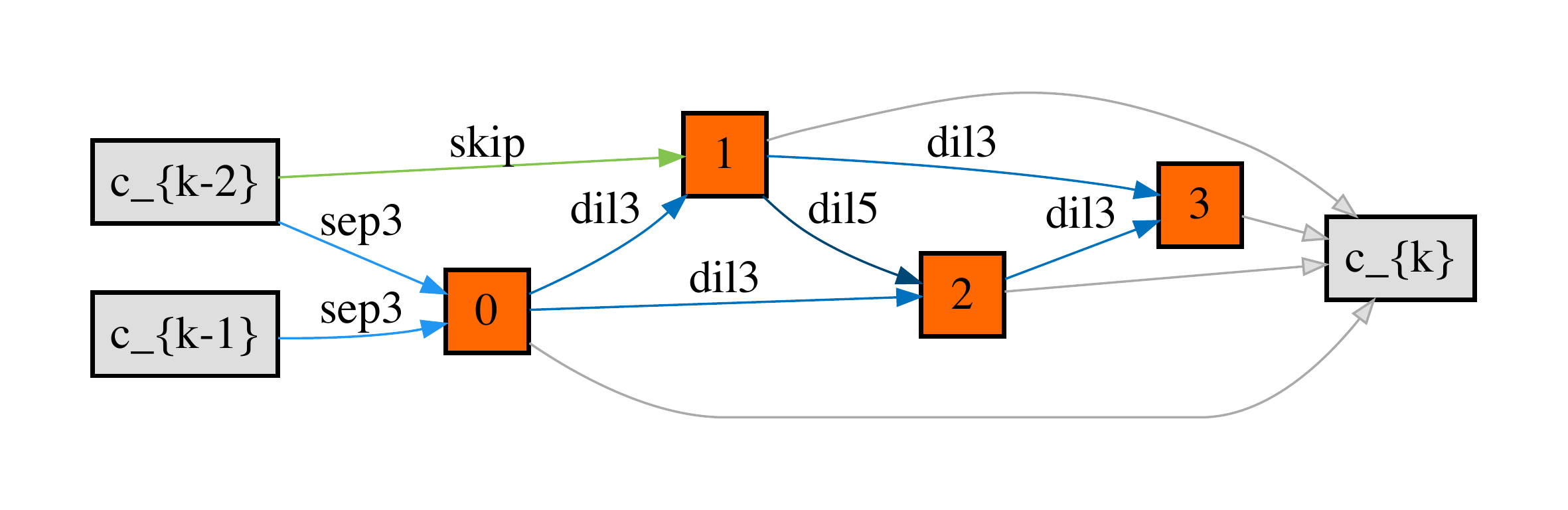}
	}
	\subfigure[Reduction cell]{
		\includegraphics[width=0.45\textwidth,scale=0.8]{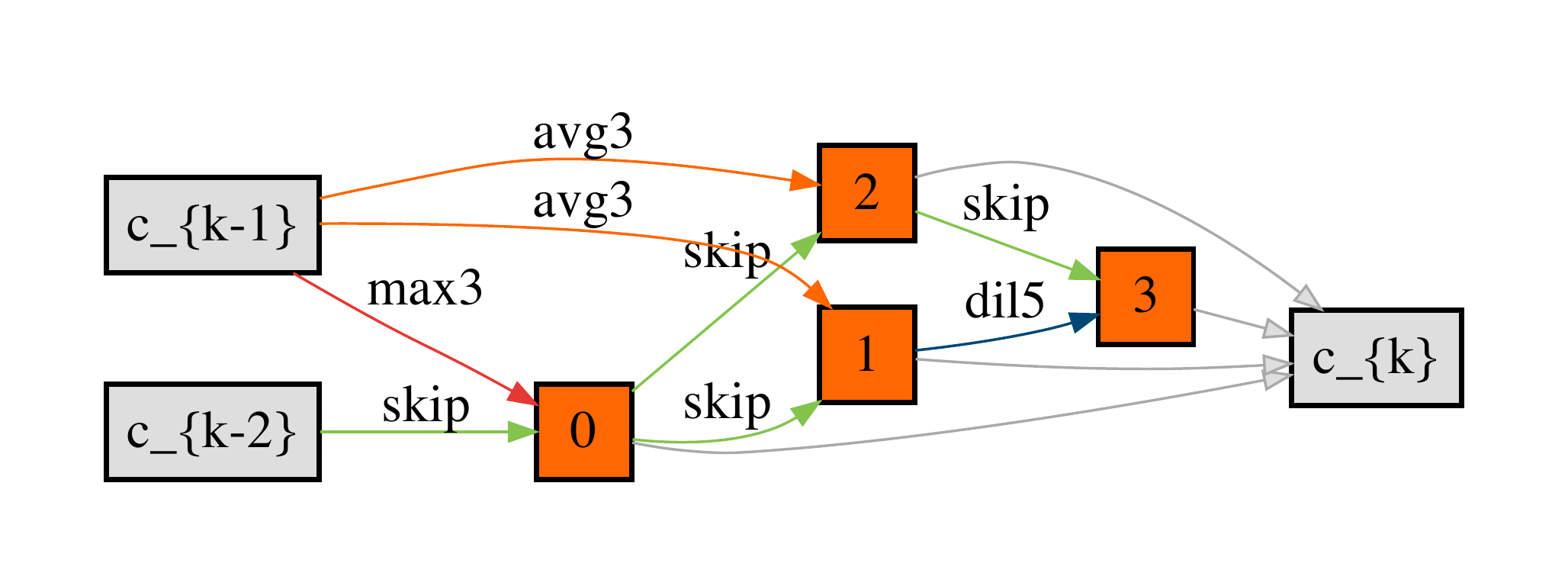}
	}
	\caption{NoisyDARTS-i cells searched on CIFAR-10 with additive Gaussian noise, $\mu=0.5$, $\sigma=0.1$.}
	\label{fig:the-searched-cells-i}
\end{figure}

\begin{figure}[ht]
	\centering
	\subfigure[Normal cell]{
		\includegraphics[width=0.45\textwidth,scale=0.8]{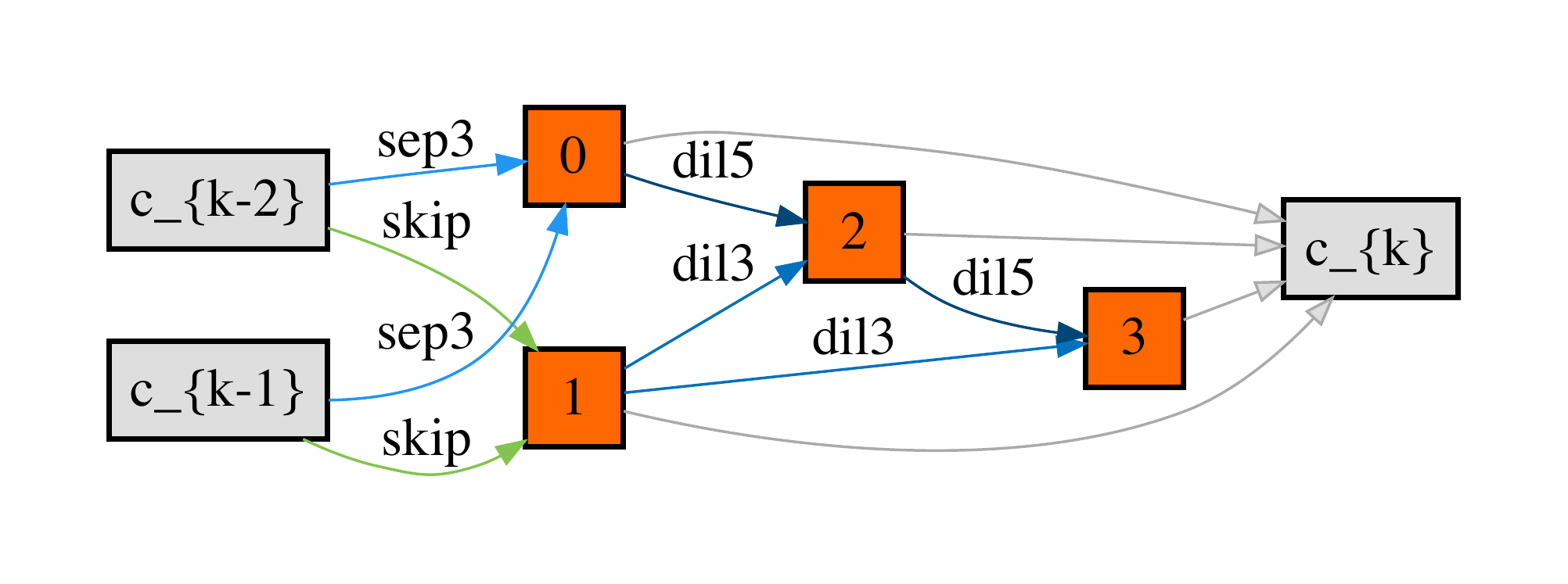}
	}
	\subfigure[Reduction cell]{
		\includegraphics[width=0.45\textwidth,scale=0.8]{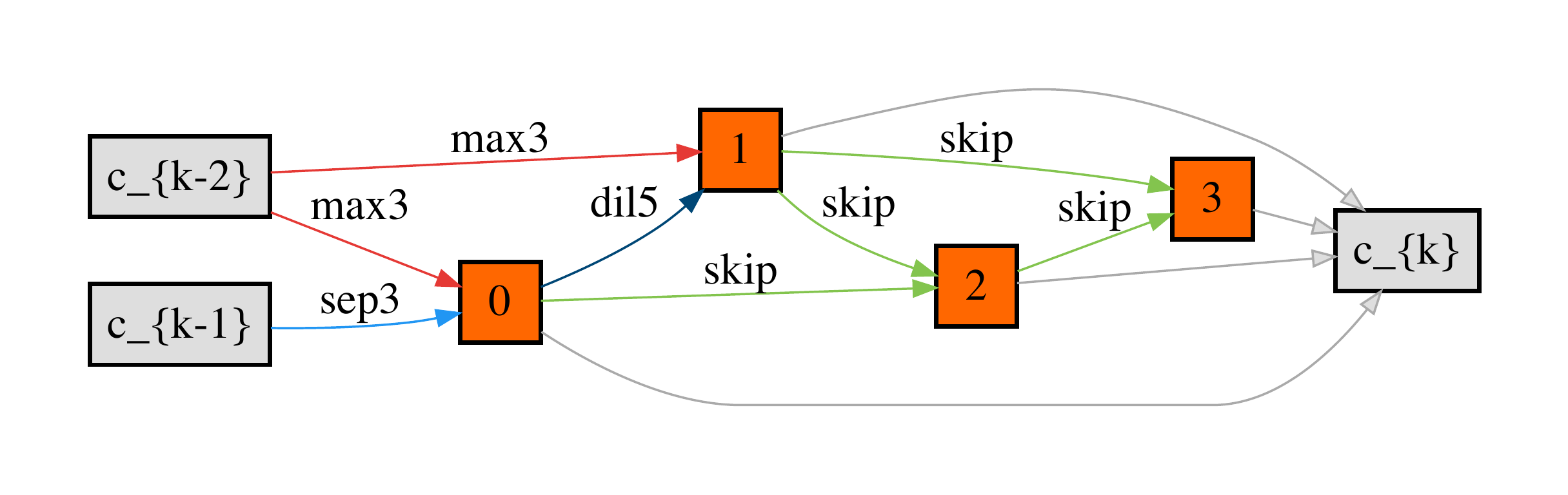}
	}
	\caption{NoisyDARTS-j cells searched on CIFAR-10 with additive Gaussian noise, $\mu=1.0$, $\sigma=0.1$.}
	\label{fig:the-searched-cells-j}
\end{figure}

\subsection{Models searched on CIFAR-10 in the reduced search spaces of RDARTS}

We plot them in Figure~\ref{fig:s2_factor_06_seed_0} and Figure~\ref{fig:s3_factor_06_seed_1}.

\begin{figure}[ht]
	\centering
	\subfigure[Normal cell]{
		\includegraphics[width=0.4\textwidth,scale=0.8]{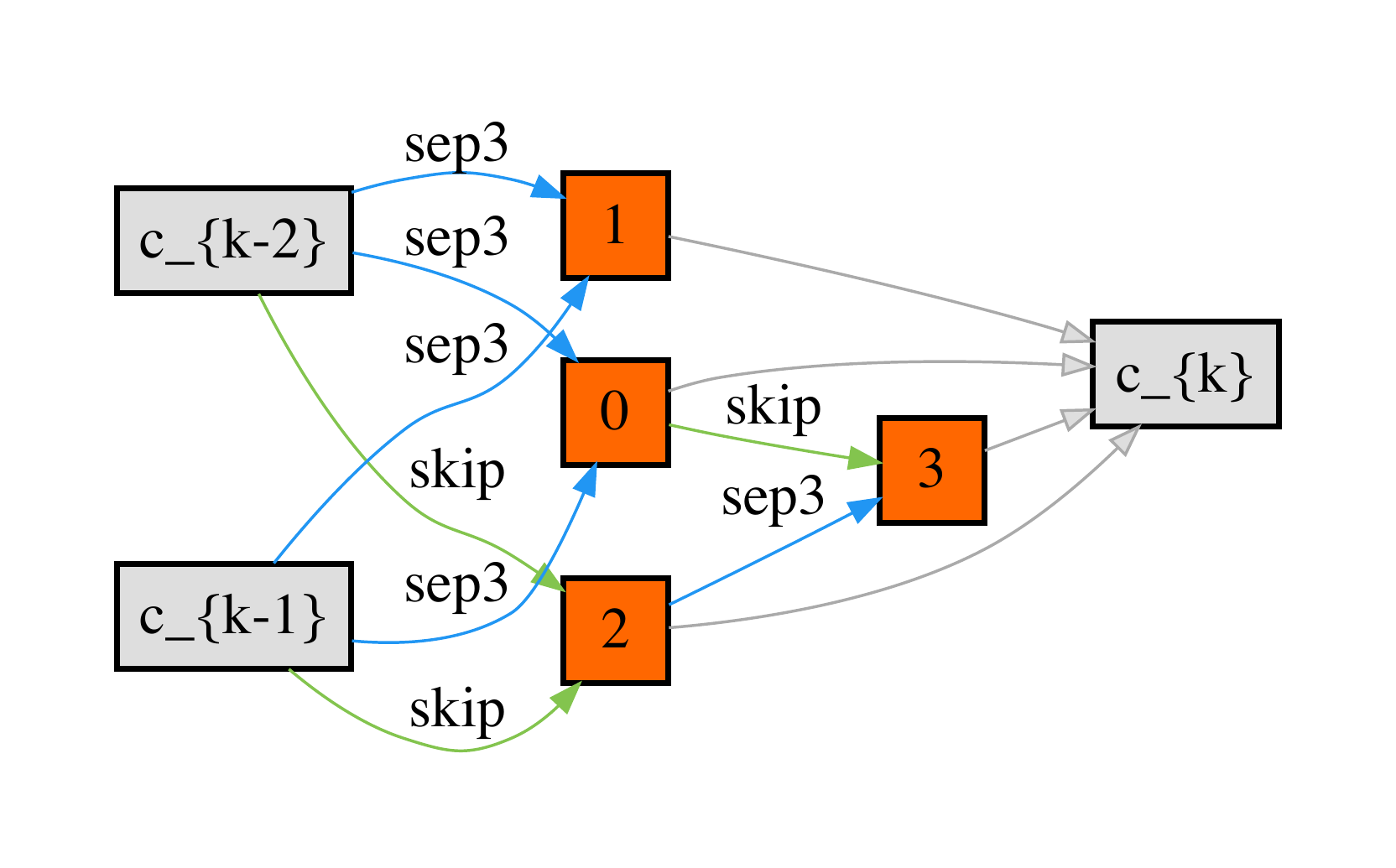}
	}
	\subfigure[Reduction cell]{
		\includegraphics[width=0.4\textwidth,scale=0.8]{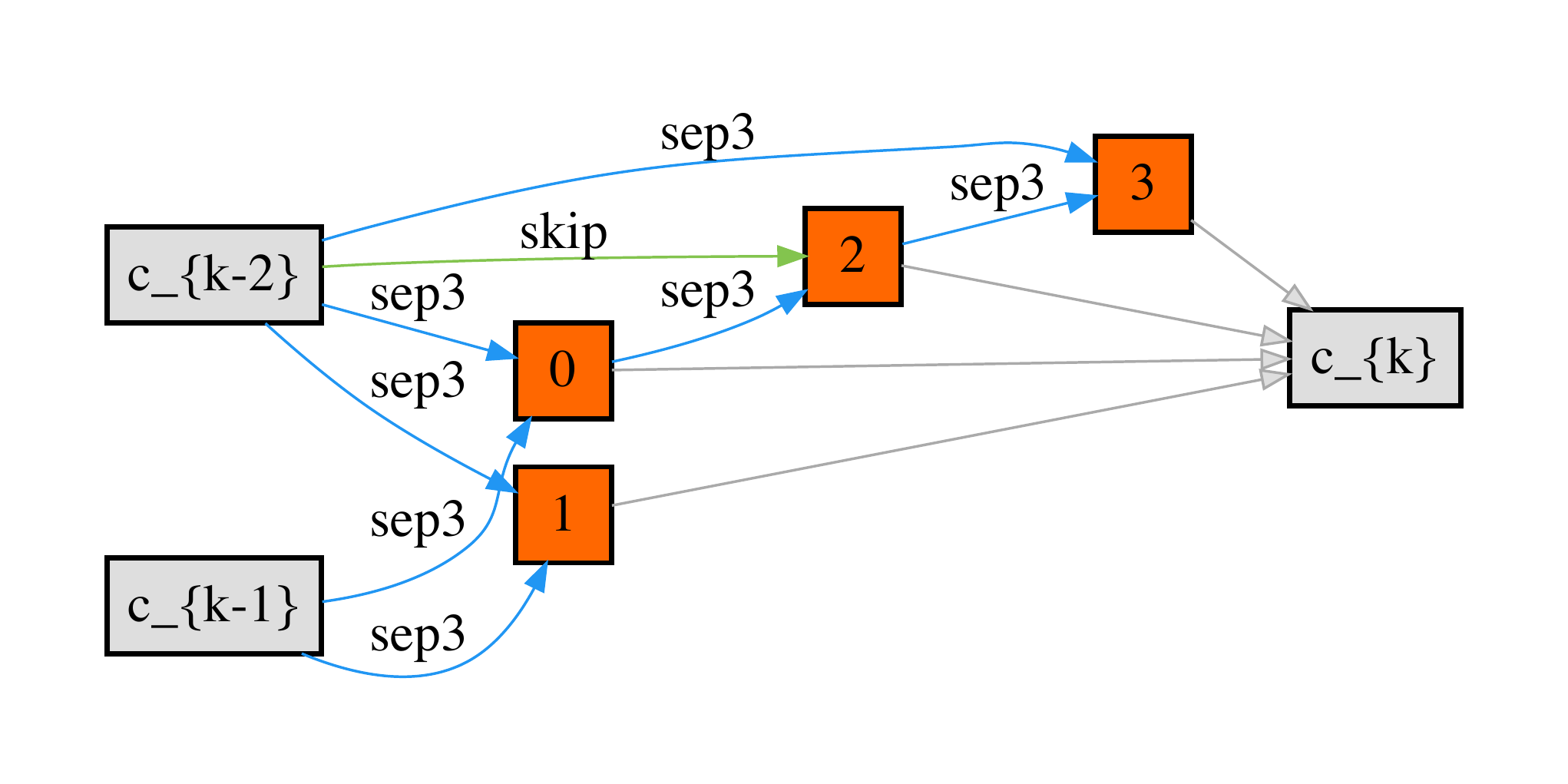}
	}
	\caption{NoisyDARTS cells searched on CIFAR-10 with additive Gaussian noise $\mu=0$, $\sigma=0.6$, in $S_2$ of RobustDARTS.}
	\label{fig:s2_factor_06_seed_0}
\end{figure}

\begin{figure}[ht]
	\centering
	\subfigure[Normal cell]{
		\includegraphics[width=0.35\textwidth,scale=0.8]{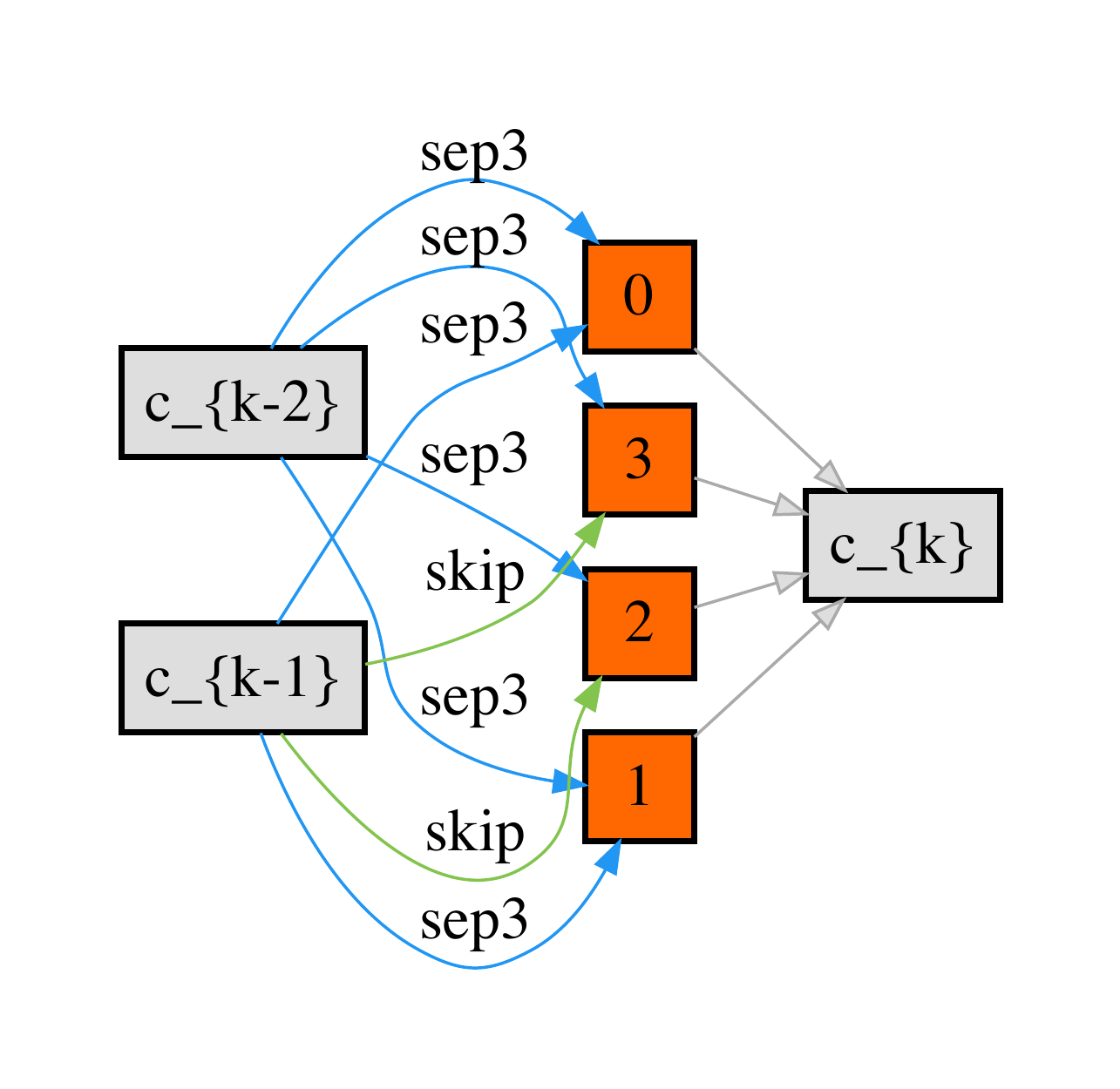}
	}
	\subfigure[Reduction cell]{
		\includegraphics[width=0.35\textwidth,scale=0.8]{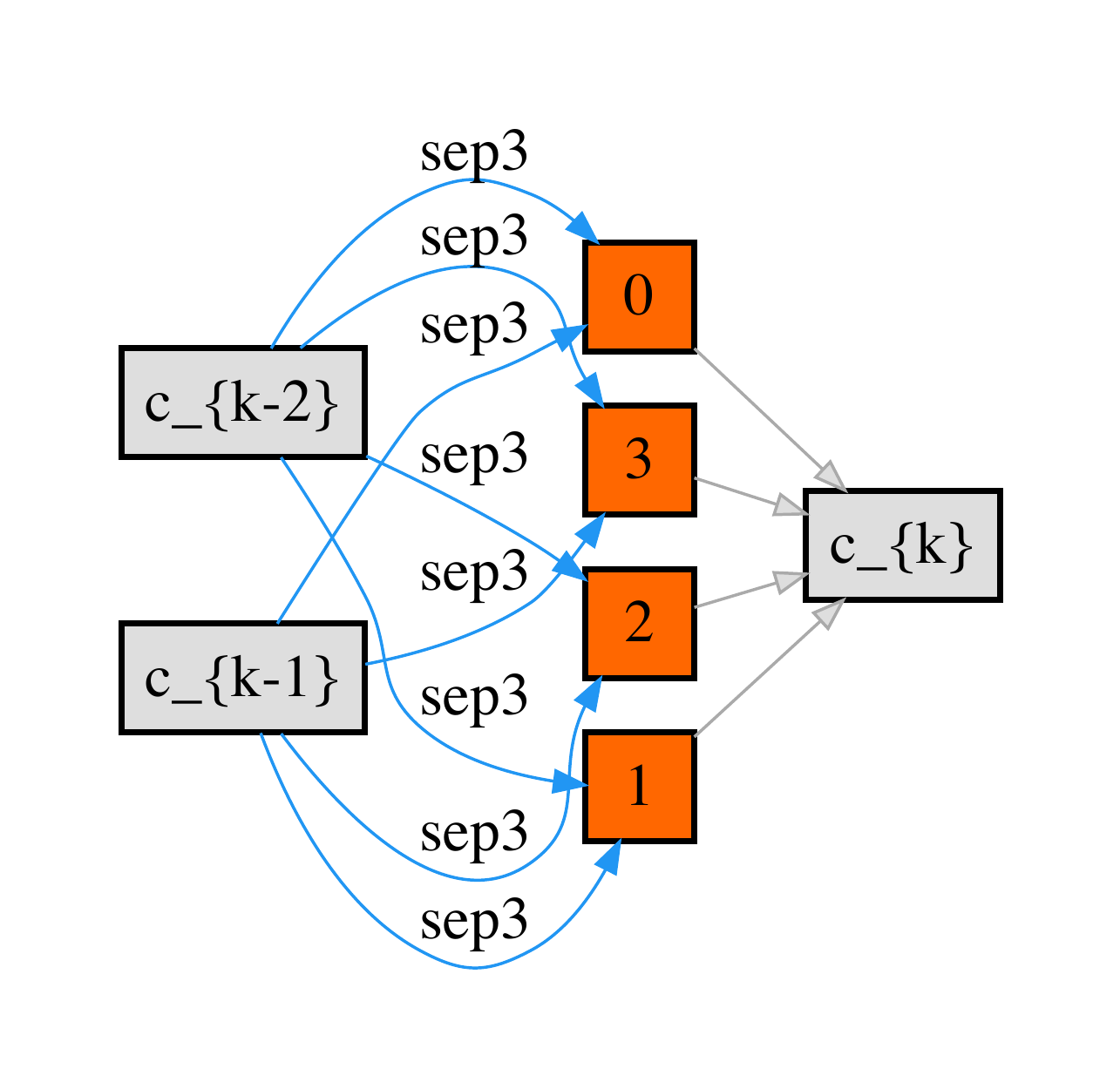}
	}
	\caption{NoisyDARTS cells searched on CIFAR-10 with additive Gaussian noise $\mu=0$, $\sigma=0.6$, in $S_3$ of RobustDARTS.}
	\label{fig:s3_factor_06_seed_1}
\end{figure}

\subsection{GCN Models searched on ModelNet10}
They are depicted in Figure~\ref{fig:gcn_models}.

\begin{figure}[ht]
	\centering
	\subfigure[$\sigma=0.2$]{
		\includegraphics[width=0.3\textwidth,scale=0.8]{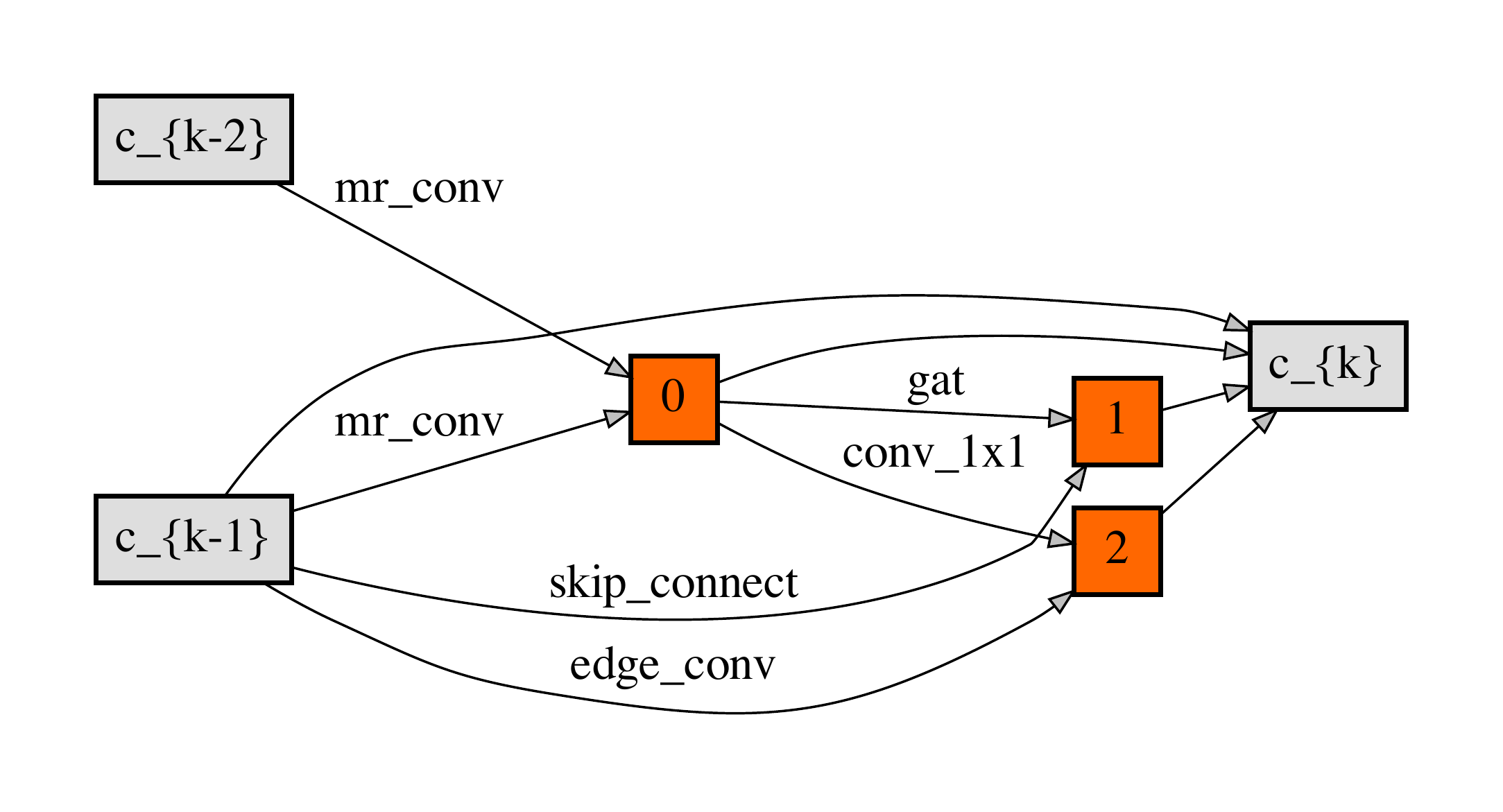}
	}
	\subfigure[$\sigma=0.3$]{
		\includegraphics[width=0.3\textwidth,scale=0.8]{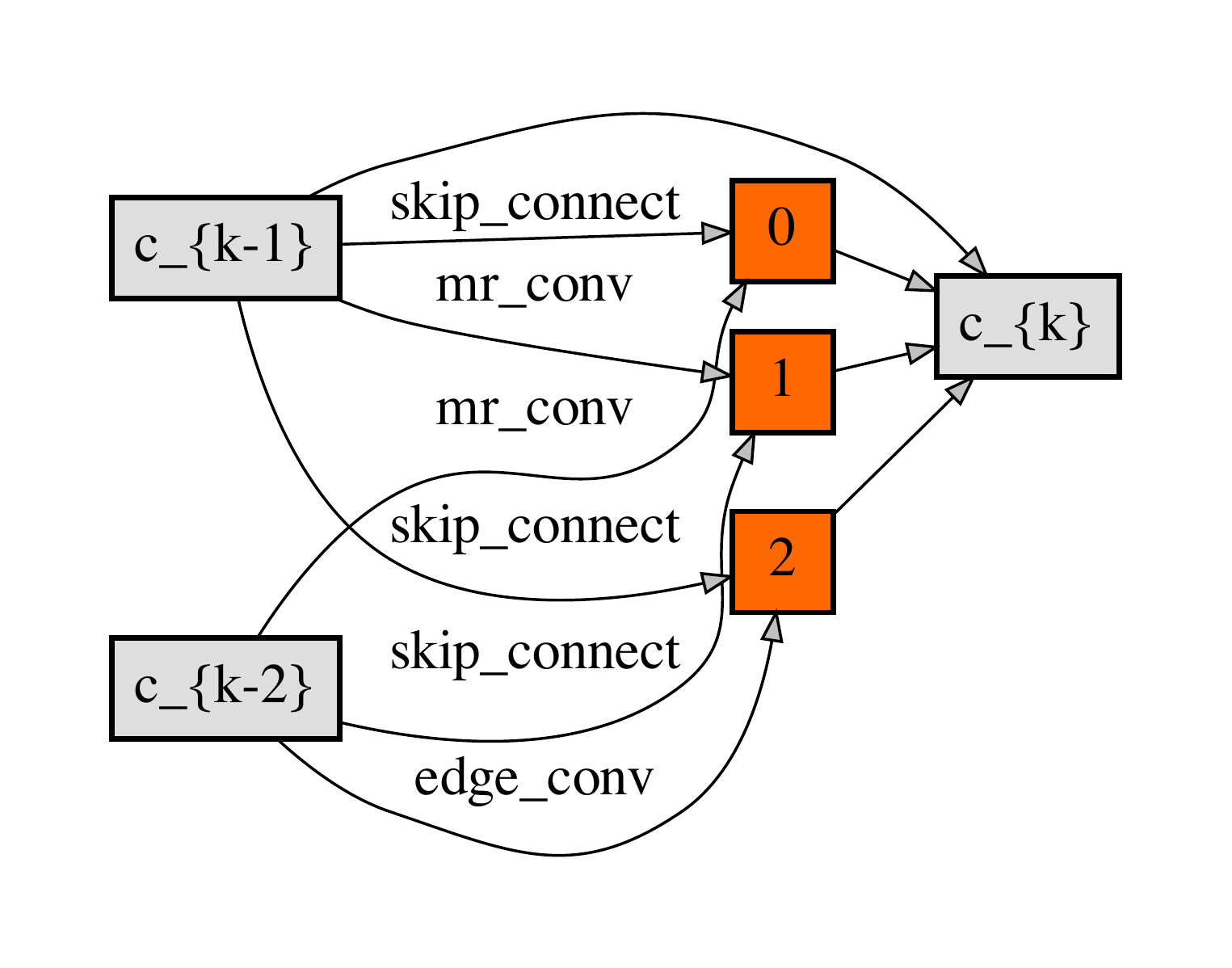}
	}
	\subfigure[$\sigma=0.4$]{
		\includegraphics[width=0.3\textwidth,scale=0.8]{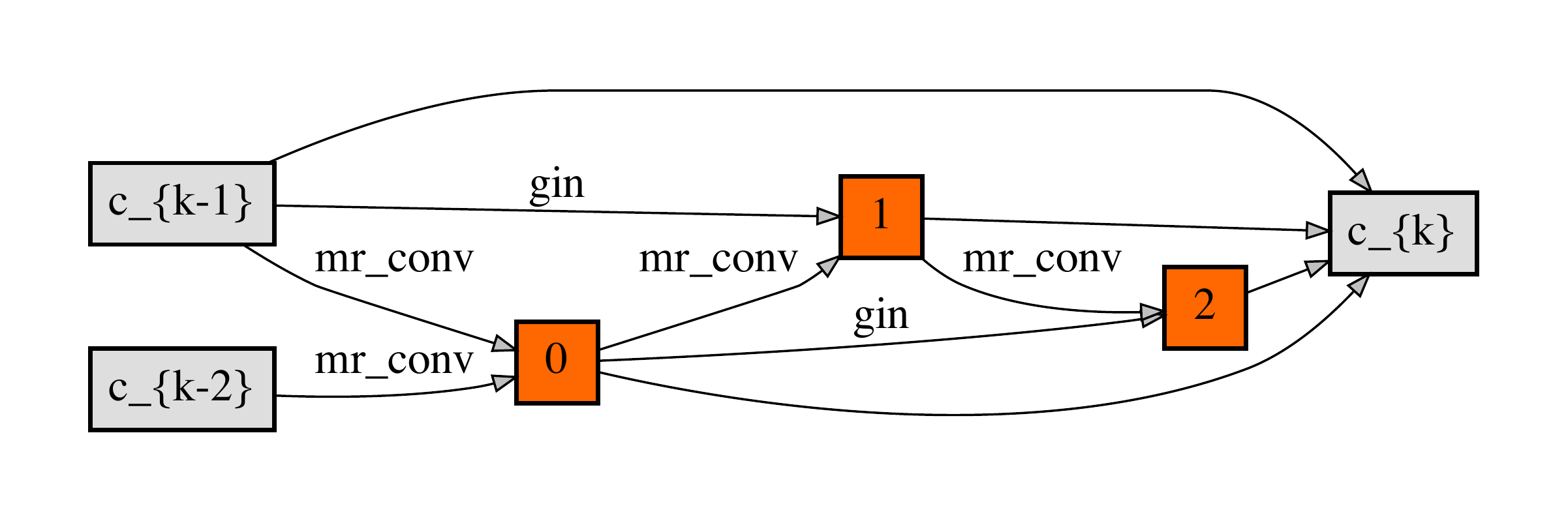}
	}
	\caption{NoisyDARTS GCN cells searched on ModelNet-10 with additive Gaussian noise $\mu=0$.}
	\label{fig:gcn_models}
\end{figure}

\end{document}